\documentclass[10pt,twocolumn,letterpaper]{article}

\usepackage[pagenumbers]{cvpr}

\usepackage{graphicx}
\usepackage{amsmath}
\usepackage{amssymb}
\usepackage{booktabs}
\usepackage{comment}
\usepackage{bm}
\usepackage{xcolor}
\usepackage{bbm}
\usepackage{makecell}
\usepackage{enumitem}
\usepackage[accsupp]{axessibility}

\newcommand{\ul}[1]{\underline{#1}}

\usepackage[linesnumbered,vlined]{algorithm2e}
\DontPrintSemicolon

\SetKwComment{Comment}{\color{green!50!black}\# }{}

\SetKwProg{Function}{function}{}{}

\newcommand{\ccom}[1]{\textcolor{green!50!black}{\# #1}}
\newcommand{\cres}[1]{\textcolor{blue!90!black}{\ttfamily\bfseries #1}}

\usepackage[pagebackref,breaklinks,colorlinks]{hyperref}

\usepackage[capitalize]{cleveref}
\crefname{section}{Sec.}{Secs.}
\Crefname{section}{Section}{Sections}
\Crefname{table}{Table}{Tables}
\crefname{table}{Tab.}{Tabs.}

\def\cvprPaperID{11779}
\def\confName{CVPR}
\def\confYear{2024}

\begin{document}

\title{Finding Lottery Tickets in Vision Models \\via Data-driven Spectral Foresight Pruning}

\author{
Leonardo Iurada \hspace{0.5cm} Marco Ciccone \hspace{0.5cm} Tatiana Tommasi\\
Politecnico di Torino, Italy \\
\texttt{\small \{leonardo.iurada, marco.ciccone, tatiana.tommasi\}@polito.it}}
\maketitle

\begin{abstract}

Recent advances in neural network pruning have shown how it is possible to reduce the computational costs and memory demands of deep learning models before training. We focus on this framework and propose a new pruning at initialization algorithm that leverages the Neural Tangent Kernel (NTK) theory to align the training dynamics of the sparse network with that of the dense one. Specifically, we show how the usually neglected data-dependent component in the NTK's spectrum can be taken into account by providing an analytical upper bound to the NTK's trace obtained by decomposing neural networks into individual paths. This leads to our Path eXclusion (PX), a foresight pruning method designed to preserve the parameters that mostly influence the NTK's trace. PX is able to find lottery tickets (\ie good paths) even at high sparsity levels and largely reduces the need for additional training. When applied to pre-trained models it extracts subnetworks directly usable for several downstream tasks, resulting in performance comparable to those of the dense counterpart but with substantial cost and computational savings. \emph{Code available at:} \url{https://github.com/iurada/px-ntk-pruning}

\end{abstract}
\section{Introduction}
\begin{figure}
\centering
\includegraphics[width=\linewidth]{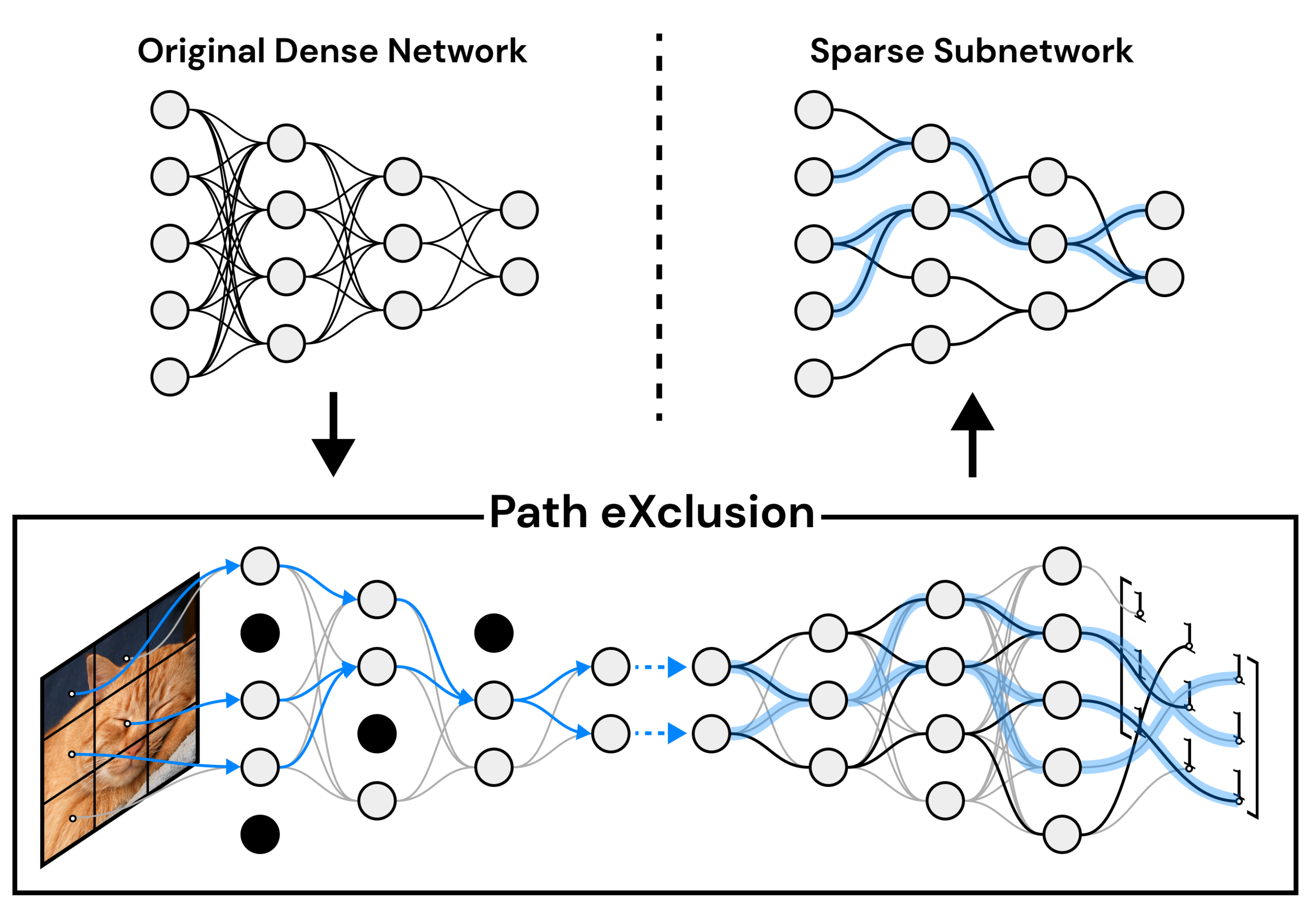}
\vspace{-5mm}
\caption{
Our Path eXclusion (PX) involves two copies of the original dense network. One copy (bottom left) estimates data-relevant paths, depicted by blue arrows, and injects the extracted information into the other network (blue shading). The other copy (bottom right) 
evaluates path relevance in terms of parameter connections in the network, illustrated by black connections. 
These estimations are then combined to score each parameter, finding a subnetwork by retaining only the most relevant paths based on data, architecture, and initialization. The identified sparse subnetwork closely mimics the training dynamics of %exhibited by 
the original dense network.
}
\label{fig:teaser} \vspace{-5mm}
\end{figure}

Almost daily we hear about new breakthroughs achieved by artificial intelligence.  Most of them are obtained by powerful foundational models \cite{brown2020language, radford2021learning, Kirillov_2023_ICCV} that however require prohibitively high computational resources and energy costs. These issues raise critical concerns in terms of financial and environmental sustainability \cite{strubell-etal-2019-energy, luccioni2023counting} and pose significant challenges for future applications requiring lightweight and efficient models embedded in always-on devices and the Internet of Things (IoT).

Given the over-parameterized nature of modern deep neural networks, one solution to alleviate their resource demands involves removing a significant number of less important neurons or connections. 
Several pruning approaches have been developed with the goal of lowering networks' complexity without sacrificing accuracy \cite{louizos2018learning,wen2016sparse,han2015learning,ICMLwang19,ICMLverma21b}, and they can further benefit from efficient implementations of sparse primitives~\cite{elsen2020fast, cusparse} and hardware designed to exploit sparsity\cite{toon2020}.
These methods are traditionally applied late in training or post-training with the goal of reducing inference time, but recent findings suggest that pruning can also be performed in advance\cite{frankle2018lottery}. 

Specifically, \emph{Pruning at Initialization} (PaI) searches for randomly initialized subnetworks that once trained can match the test accuracy of the original dense networks with a largely reduced learning cost. Prior works have proposed PaI strategies based on the impact of each parameter on the loss 
\cite{lee2018snip,alizadeh2022prospect} or on different saliency metrics that estimate the information flow in the network \cite{tanaka2020synflow}. 
Some recent publications have targeted the evaluation of the training dynamics based on the Neural Tangent Kernel Theory (NTK,\cite{NTK_Jacot2018}) to define the parameters' scores. Although showing promising results, they usually neglect \cite{gebhart2021unified} or loosely approximate \cite{ntksap2023wang} the data contribution to the NTK spectrum as they claim that data has a minimal impact on finding lottery tickets \ie good paths in the network \cite{frankle2018lottery}. Some of these approaches also suffer for layer collapse \ie the premature pruning of an entire layer that would make the network untrainable. As discussed in \cite{tanaka2020synflow}, its occurrence can be avoided under specific conditions. 

One question that remains open is whether pruning can be applied to pre-trained networks before their downstream transfer \cite{chen2021lottery, iofinova2022well}. This is a crucial and timely problem as pre-trained models continue to grow in size, and pruning could be used to reduce the cost of fine-tuning on downstream tasks. 
This still defines a proper PaI setting where the initial model is trained on huge corpora and the goal is not only to compress it but also to preserve its transferability capabilities in the obtained subnetworks.

\textbf{With our work we advance PaI research by proposing the following contributions:}
\begin{itemize}[leftmargin=*]
\item We present Path eXclusion (PX, see Figure \ref{fig:teaser}.) a PaI method that estimates the relevance of each network's parameter to the training dynamic through a newly defined bound on the trace of the NTK. 
\begin{itemize}[leftmargin=*]
\vspace{-1.5mm}\item The saliency function formulated from the bound guarantees that the network parameters have only positive scores. Together with the iterative nature of PX, this provides guarantees on avoiding layer collapse.
\item Differently from previous work, the relevance score of our PX depends jointly on the data, and on the native neurons and layer connections. 
\end{itemize}
\vspace{-3mm}\item We experimentally show that PX is not only robust to different architectures and tasks, but can be effectively used to search for subnetworks in large pre-trained models that maintain almost intact transferability.
\end{itemize}

\section{Related Works}
The question of how to significantly reduce the number of parameters of a neural network while maintaining its performance dates back to the 1980s\cite{lecun_optimal}. Several strategies include matrix and tensor factorization \cite{NIPS2015tensorizing}, generalized dropout 
\cite{Srinivas2016GeneralizedD}, and adding regularization terms in the learning objective \cite{louizos2018learning,wen2016sparse} to enforce sparse networks. Other approaches identify parameters with low magnitude after training and discard the corresponding connections \cite{lecun_optimal,hassibi1993surgeon,han2015learning,ICMLwang19,ICMLverma21b}. For all these methods, the main goal is to improve test efficiency while the computational cost of training remains the same as that of a dense network.

In the last years, the focus has moved towards efficient training with one milestone provided by the \emph{Lottery Ticket Hypothesis}~\cite{frankle2018lottery}. It demonstrated that within overly large networks it is possible to identify \emph{winning tickets}, \ie smaller subnetworks that once trained perform nearly as well as their dense counterpart.
The Iterative Magnitude Pruning (IMP) algorithm discovers these subnetworks through several rounds of alternated training and progressive pruning guided by the magnitude of the surviving parameters. Despite its effectiveness, the high computational costs of IMP 
led to the development of alternative cheaper methods for finding sparse networks. 
They are usually identified as \emph{Pruning-at-Initialization} (PaI), or \emph{foresight pruning} algorithms and can be organized into two main families. 

The \emph{data-agnostic} methods exploit either random or constant mini-batches to probe the network and score each parameter on the basis of its relevance to some network's property. Then, only a small fraction of the parameters with top scores is kept for training. SynFlow \cite{tanaka2020synflow} builds on the hypothesis that the synaptic saliency for the incoming parameters to a hidden neuron is equal to the sum of the synaptic saliency for the outgoing ones. Thus it evaluates the importance of each parameter on the basis of its relation to those in the previous and following layers. SynFlow-L2 \cite{gebhart2021unified} scores each parameter by considering its contribution to the network's training dynamics estimated via the Neural Tangent Kernel theory \cite{NTK_Jacot2018}. LogSynFlow \cite{cavagnero2023freerea} rescales the scores of SynFlow to account for the possible issue of exploding gradients.  
NTK-SAP \cite{ntksap2023wang} improves the previous methods by exploiting a more precise estimate of the training dynamics defined from the full spectrum of the Neural Tangent Kernel and then discards parameters that contribute the least to it.
All these approaches compute importance scores iteratively with multiple forward-backward passes over the network, while PHEW \cite{patil2021phew} introduces random walks biased towards higher parameter magnitudes and requires a single pruning round.

The \emph{data-driven} methods assert the relevance of the data and of the learning task in evaluating the importance of each network's parameter when pruning and avoiding large degradation in model performance. SNIP \cite{lee2018snip} defines a saliency score for the parameters based on how they contribute to changing the initial loss. ProsPR \cite{alizadeh2022prospect} combines an estimate of the effect of pruning on the loss and on the meta-gradients that define the optimization trajectory.
GraSP \cite{grasp2020wang} takes the gradient norm after pruning as a reference criterion, and drops the parameters that result in its least decrease. 
These methods are single-shot, while some variants of SNIP such as IterSNIP \cite{de2020itersnip} and FORCE \cite{de2020force} exploit iterative solutions to avoid layer collapse. 

Our work falls in between the two families described. We build on the NTK theory already used by the data-agnostic approaches and we show how information from the data can be used to guide the pruning process with significant advantages in training efficiency. Indeed, as already discussed in \cite{Yang22bmvc,frankle2021missing,HoeflerJMLR21} data independence can be considered as a limitation rather than a benefit. An intuitive reason is that data statistics have a crucial effect on some network components as batch normalization that contributes to the overall network behavior and parameter relevance \cite{tanaka2020synflow}. 
Moreover, by focusing on the network training dynamics rather than on the loss, as most of the data-driven approaches do, our method proves to be task-independent, with the obtained sparse network remaining effective even when transferred to new downstream tasks.
\section{Method}
In this section, we start by describing the standard framework adopted by Pruning-at-Initialization (PaI) methods and the intuition of our foresight pruning algorithm designed to calculate and preserve the trace of the Neural Tangent Kernel (NTK). Afterward, we provide a brief overview of the theory underpinning NTK and present how to express its trace by exploiting the notion of network paths. Finally, we introduce our Path eXclusion (PX) algorithm that drops those network weights that minimally change the trace of the NTK, so that the obtained sparse network retains only the most relevant paths of the original dense network.

\subsection{Problem Formulation}
Let us consider a neural network $\bm{f}: \mathbb{R}^d \rightarrow \mathbb{R}^K$ parametrized by $\bm{\theta} \in \mathbb{R}^m$ and a dataset of $N$ data points ${(\bm{X}, \bm{Y}) = \{(\bm{x}_i, \bm{y}_i)\}_{i=1}^N}$ with $\bm{x}_i \in \mathbb{R}^{d}$ and $\bm{y}_i \in \{1,...,K\}$. The problem of \emph{unstructured} neural network pruning can be formalized as finding a binary mask $\bm{M} \in \{0,1\}^m$ that optimizes the following objective
\begin{equation}\label{eq:pruning_problem}
\begin{split}
    \min_{\bm{M}} \frac{1}{N} \sum_{i=1}^N \mathcal{L}(\bm{f}(\bm{x}_i; \mathcal{A}(\bm{\theta}_0, \bm{M}) \odot \bm{M}), \bm{y}_i) \\
 \text{s.t.  } \bm{M} \in \{0,1\}^m, ~~~ \|\bm{M}\|_0 / m \leq 1 - q  ~,
\end{split}
\end{equation}
where $\mathcal{L}$ is a suitable loss function for the downstream task, $q$ is the desired sparsity of the resulting subnetwork and $\bm{\theta}_0$ are the initial parameters. $\mathcal{A}$ is an optimization algorithm (\eg SGD \cite{ruder2016overview}, Adam \cite{kingma2014adam}) that takes as input the mask $\bm{M}$ and the initialization $\bm{\theta}_0$ and returns the trained parameters at convergence $\bm{\theta}_{final}\odot \bm{M}$, where $\odot$ denotes the element-wise (Hadamard) product.

Due to the practical intractability of the described optimization problem, recent PaI algorithms focus on the notion of \emph{saliency}, which is used as a score to assess the significance of network parameters regarding some property $F$ of the network. After ranking the parameters' scores, only the top-$S$ mask elements are retained and the final mask is used to approximate a solution for Eq. (\ref{eq:pruning_problem}). Formally, the saliency takes the following form
\begin{equation}
    S(\bm{\theta}) = \frac{\partial F}{\partial \bm{\theta}} \odot \bm{\theta}~.
\end{equation}
For instance, in {SNIP} \cite{lee2018snip} the saliency is the loss function:  $F=\mathcal{L}(\bm{\theta} \odot \bm{M}; (\bm{X}, \bm{Y}))$. Thus, that method assigns to each parameter a score which reflects how the loss would change when removing that specific parameter from the network.

In this work, our goal is to devise a suitable saliency score that correctly reflects how much each weight contributes to the trace of the NTK. As described in detail in the next subsection, the NTK approximates the training dynamics of the network \cite{NTK_Lee2019}, so removing those weights that minimally change its trace will result in small variations in the NTK spectrum, producing a subnetwork with similar predictive potential to the original larger network. 

\subsection{Neural Tangent Kernel and Pruning}
We indicate the output of a ReLU-based neural network as $\bm{f}(\bm{X}, \bm{\theta}) \in \mathbb{R}^{NK}$. Under the \emph{gradient flow} regime (\ie continuous-time gradient descent, with learning rate $\alpha$) we can use a first-order Taylor expansion to approximate the network's output at a time step $t$ of the optimization process:
\vspace{-5mm}
\begin{equation}
    \bm{f}(\bm{X}, \bm{\theta}_{t+1}) = \bm{f}(\bm{X}, \bm{\theta}_{t}) - \alpha \bm{\Theta}_{t}(\bm{X}, \bm{X}) \nabla_{\bm{f}}\bm{\mathcal{L}}~.
\end{equation}
The matrix $\bm{\Theta}_{t}(\bm{X}, \bm{X}) = \nabla_{\bm{\theta}} \bm{f}(\bm{X}, \bm{\theta}_t) \nabla_{\bm{\theta}} \bm{f}(\bm{X}, \bm{\theta}_t)^T \in \mathbb{R}^{NK \times NK}$ is the \emph{Neural Tangent Kernel} at time step $t$ \cite{NTK_Jacot2018}. 
For infinitely wide networks, the exact training dynamics is described by the NTK which is a constant matrix that depends only on data, initialization, and architecture. It holds $\bm{\Theta}_{t}(\bm{X}, \bm{X}) = \bm{\Theta}_{0}(\bm{X}, \bm{X})$, thus we can drop the subscripts. 
Further works \cite{NTK_Lee2019} observed that the NTK can approximate the training dynamics of networks of any depth without necessarily being infinitely wide by rendering its theory usable in practice. Additionally, it has been shown that a faster convergence is correlated with the direction in the parameter space pointed by the eigenvector with the largest corresponding eigenvalue of the NTK \cite{arora19a}. 

It is apparent that the NTK and its spectrum encapsulate crucial information about their model and offer an appealing way to evaluate the alignment between two networks. 
Models sharing the same NTK exhibit similar training dynamics \cite{ntksap2023wang}, even with different parameter counts. 
Empirical results indicate that sparse subnetworks maintaining the NTK's largest eigenvalues of their dense counterpart converge more rapidly~\cite{grasp2020wang, ntksap2023wang} and better replicate the training dynamics of denser networks~\cite{kopitkov2020neural, su2019learning}.

While the strategy of using the NTK for network pruning seems promising, calculating the entire NTK spectrum is only feasible for very small neural networks with limited data \cite{arora19a}. For context, recent results on the NTK computation state a time complexity of $N^2K[FP]$ \cite{novak2022fastntk}, where $N$ is the size of the dataset, $K$ is the output size of the network, and $[FP]$ is the cost of a single forward pass. Indeed previous pruning methods that exploited the NTK theory either resorted to different approximations of the NTK spectrum \cite{patil2021phew, gebhart2021unified,ntksap2023wang} or indirectly tried to preserve it by maintaining the gradient flow in the network \cite{grasp2020wang}.

The next subsection explains how, by analyzing the activation paths within a neural network, it is possible to obtain an analytical decomposition of the trace of the NTK that is instrumental for PaI.

\subsection{Neural Tangent Kernel and Activation Paths}
Let $\mathcal{P}$ be the set of all paths connecting any input neuron to any output neuron of the network $\bm{f}$, where the edges of those paths are the weights\footnote{We use the terms ``parameters'' and ``weights'' interchangeably to refer to the network's parameters $\bm{\theta}$, as paths within a neural network are weighted by the value of each parameter.} of the network. 
Each specific path can be referred to by its index $p = 1, ..., P$ in the set $\mathcal{P}$. The presence of weight $\theta_i$ in path $p$ is denoted as $p_i = \mathbb{I}[\theta_i \in p]$.
We can now define the product of weights within a path $p$ as $v_p(\bm{\theta}) = \prod_{i=1}^m \theta_i^{p_i}$, where  ${p_i}$ is the exponent. Given an input example $\bm{x} \in \bm{X}$, the activation status of a path is $a_p(\bm{x}, \bm{\theta}) = \prod_{\{ i | \theta_i \in p\}} \mathbb{I}[z_i > 0]$, where $z_i$ is the activation of the neuron connected to the previous layer through $\theta_i$. 
Thus,  we can describe the $k$-th component of the output function of the network as:
\begin{equation}\label{eq:f_k_out}
    \bm{f}^k (\bm{x}, \bm{\theta}) = \sum_{s=1}^d \sum_{p \in \mathcal{P}_{s \rightarrow k}} \bm{v}_p(\bm{\theta}) a_p(\bm{x}, \bm{\theta}) \bm{x}_s~,
\end{equation}
where $\bm{x}_s$ indicates the $s$-th term of the $\bm{x}$ vector and $\mathcal{P}_{s \rightarrow k}$ is the set of all paths from the input $s$ to output neuron $k$ \cite{meng2018gsgd}.

By applying the chain rule it is possible to factorize the NTK as follows \cite{gebhart2021unified}:
\begin{align} \label{eq:NTK}\nonumber
    \bm{\Theta}(\bm{X}, \bm{X}) &= \nabla_{\bm{\theta}} \bm{f}(\bm{X}, \theta) \nabla_{\bm{\theta}} \bm{f}(\bm{X}, \theta)^T \\ \nonumber
    &= \frac{\partial \bm{f}(\bm{X}, \bm{\theta})}{\partial \bm{\theta}} \frac{\partial \bm{f}(\bm{X}, \bm{\theta})}{\partial \bm{\theta}}^T \\ \nonumber
    &= \frac{\partial \bm{f}(\bm{X}, \bm{\theta})}{\partial \bm{v}(\bm{\theta})} \frac{\partial \bm{v}(\bm{\theta})}{\partial \bm{\theta}} \frac{\partial \bm{v}(\bm{\theta})}{\partial \bm{\theta}}^T \frac{\partial \bm{f}(\bm{X}, \bm{\theta})}{\partial \bm{v}(\bm{\theta})}^T \\ \nonumber
    &= \bm{J}_{\bm{v}}^{\bm{f}}(\bm{X}) \bm{J}_{\bm{\theta}}^{\bm{v}} (\bm{J}_{\bm{\theta}}^{\bm{v}})^T (\bm{J}_{\bm{v}}^{\bm{f}}(\bm{X}))^T \\
    &= \bm{J}_{\bm{v}}^{\bm{f}}(\bm{X}) \bm{\Pi}_{\bm{\theta}} (\bm{J}_{\bm{v}}^{\bm{f}}(\bm{X}))^T~. 
\end{align}
Here $\bm{J}_{\bm{\theta}}^{\bm{v}} \in \mathbb{R}^{P \times m}$ compose the so-called \emph{Path Kernel} matrix $\bm{\Pi}_{\bm{\theta}} \in \mathbb{R}^{P \times P}$ which is symmetric positive semi-definite and depends solely on the initialization and the network's architecture. The eigenvectors of the Path Kernel can be described as a collection of paths where the eigenvector associated with the largest eigenvalue represents the set of paths that maximize the flow within the network \cite{gebhart2021unified}. On the other hand, the matrix $\bm{J}_{\bm{v}}^{\bm{f}}(\bm{X}) \in \mathbb{R}^{NK \times P}$, which we renamed, \textit{Path Activation Matrix}, represents the change in output with respect to path values and entirely captures the dependence of $\bm{f}$ on the inputs by reweighting the paths within the network based on the training data.

Considering the eigenvalues $\pi_i$, $\nu_i$ and $\lambda_i$ respectively of $\bm{\Pi}_{\bm{\theta}}$, $\bm{J}_{\bm{v}}^{\bm{f}}(\bm{X})$ and $\bm{\Theta}(\bm{X}, \bm{X})$, it was demonstrated that $Tr[\bm{\Theta}(\bm{X}, \bm{X})]=\sum_i^{NK} \lambda_i \leq \sum_i^{NK} \nu_i \pi_i$ \cite{gebhart2021unified}.
Previous works mentioning this upper bound~\cite{patil2021phew, gebhart2021unified}, end up focusing only on the Path Kernel of the pruned networks and maximize its trace $Tr(\bm{\Pi}_{\bm{\theta}}) = \sum_i \pi_i$ to preserve the largest NTK eigenvalue of the original network, which produces the highest ﬂow through the network and hence, similar training dynamics.
However, this might be misleading as the data-dependent term $\bm{J}_{\bm{v}}^{\bm{f}}(\bm{X})$ is neglected. 

In the following, we present a new upper bound for the NTK's trace that considers both the Path Kernel and the Path Activation Matrix, along with an exact calculation method, forming the core of our novel Path eXclusion approach for pruning.

\subsection{Foresight Pruning via Path eXclusion}

Starting from the decomposition of the NTK in Eq. (\ref{eq:NTK}) and from the definition  of the Frobenius norm $\| A \|_F = \sqrt{Tr(A A^T)}$ it is possible to write
\begin{align}
\label{eq:UpBo}
\nonumber
    Tr[\bm{\Theta}(\bm{X}, \bm{X})] &= Tr[\nabla_{\bm{\theta}} \bm{f}(\bm{X}, \bm{\theta}) \nabla_{\bm{\theta}} \bm{f}(\bm{X}, \bm{\theta})^T]\\
    \nonumber
    &= \|\nabla_{\bm{\theta}} \bm{f}(\bm{X}, \bm{\theta})\|_F^2 \\
    \nonumber
    &= \|\bm{J}_{\bm{v}}^{\bm{f}}(\bm{X}) \bm{J}_{\bm{\theta}}^{\bm{v}}\|_F^2 \\
    &\leq \|\bm{J}_{\bm{v}}^{\bm{f}}(\bm{X})\|_F^2 \cdot \| \bm{J}_{\bm{\theta}}^{\bm{v}}\|_F^2 ~,
\end{align}
where the last inequality arises from the submultiplicative property of the matrix norm. It is easy to show that 
\begin{align}
\label{eq:Jv}
\nonumber
    \|\bm{J}_{\bm{v}}^{\bm{f}}(\bm{X})\|_F^2 &= \sum_{n=1}^N \sum_{k=1}^{K} \sum_{p=1}^{P} \left(\sum_{s=1}^d \mathbb{I}[p \in \mathcal{P}_{s \rightarrow k}] a_p(\bm{x}_n, \bm{\theta}) \bm{x}_{n_s}\right)^2 \\
    &= \sum_{n=1}^N \sum_{k=1}^{K} \sum_{p=1}^{P} a_p(\bm{x}_n, \bm{\theta}) \bm{x}_{n_{s|s \in p}}^2 ~,
\end{align}
where $\bm{x}_{n_s}$ is the $s$-th component of the $n$-th sample vector $\bm{x}$. This term captures the dependence of the NTK's trace on the input data by choosing which paths are active and re-weighting by the input activations.
The second term of the upper bound relates to the Path Kernel and  
as already discussed in \cite{gebhart2021unified}, it holds
\begin{align}\label{eq:Jtheta}
 \|\bm{J}_{\bm{\theta}}^{\bm{v}}\|_F^2 &= 
    \sum_{p=1}^{P} \sum_{j=1}^{m} \left( \frac{\bm{v}_p(\bm{\theta})}{\bm{\theta}_j} \right)^2 ~.
\end{align}
Both Eq. (\ref{eq:Jv}) and (\ref{eq:Jtheta}) can be calculated efficiently by exploiting the implicit computation of the network's gradients via automatic differentiation. To do that we introduce two auxiliary networks $\bm{h}$ and $\bm{g}$ which have the same architecture as the original $\bm{f}$ and are described by the input data $\bm{x}$, their parameters $\bm{\theta}$, and the status $\bm{a}$ of their ReLU activations. Using  $\mathbbm{1}$ to indicate a vector of $1$'s, the $k$-th component of the output for each of these networks is defined as 
$$
\bm{h}^k(\mathbbm{1}, \bm{\theta}^2, \mathbbm{1}) = \sum_{p=1}^P \bm{v}_p(\bm{\theta}^2) = \sum_{p=1}^P \bm{v}_p^2(\bm{\theta})~,
$$
$$
\bm{g}^k(\bm{x}^2, \mathbbm{1}, \bm{a}) = \sum_{p=1}^P a_p(\bm{x}, \bm{\theta}) \bm{x}_{s|s\in p}^2 ~.
$$
Here $\bm{h}$ takes simplified data $\mathbbm{1}$ as input, with squared parameters, and a vector of activations that are all one. Instead $\bm{g}$ takes the squared data as input, the parameters are all one, and the activations status is an exact copy of that of $\bm{f}$.
Finally, we consider these two networks working jointly with the overall behavior described by 
\begin{align*}
    \mathcal{R}(\bm{x}, \bm{\theta}, \bm{a}) &= \sum_{n=1}^N \sum_{k=1}^K \bm{g}^k(\bm{x}_n^2, \mathbbm{1}, \bm{a}_n) \cdot \bm{h}^k(\mathbbm{1}, \bm{\theta}^2, \mathbbm{1})  ~.
\end{align*}
We can compute the gradient of this function via backpropagation obtaining 
\begin{align*}
    &\sum_{j=1}^m \frac{\partial \mathcal{R}(\bm{x}, \bm{\theta}, \bm{a})}{\partial \bm{\theta}_j^2} =\\
   \hspace{-3mm} &= \frac{\partial}{\partial \bm{\theta}_j^2}
    \sum_{n=1}^N \sum_{k=1}^K \sum_{j=1}^m \bm{g}^k(\bm{x}_n^2, \mathbbm{1}, \bm{a}_n) \cdot \bm{h}^k(\mathbbm{1}, \bm{\theta}^2, \mathbbm{1})\\
    &= \sum_{n=1}^N \sum_{k=1}^K \bm{g}^k(\bm{x}_n^2, \mathbbm{1}, \bm{a}_n) \cdot \sum_{j=1}^m \frac{\partial \bm{h}^k(\mathbbm{1}, \bm{\theta}^2, \mathbbm{1})}{\partial \bm{\theta}_j^2}\\
    &= \sum_{n=1}^N \sum_{k=1}^K \sum_{p=1}^P a_p(\bm{x}_n, \bm{\theta}) \bm{x}_{n_{s|s\in p}}^2 \cdot \sum_{p=1}^P \sum_{j=1}^m \frac{\bm{v}_p^2(\bm{\theta})}{\bm{\theta}_j^2}\\
    &= \|\bm{J}_{\bm{v}}^{\bm{f}}(\bm{X})\|_F^2 \cdot \| \bm{J}_{\bm{\theta}}^{\bm{v}}\|_F^2 ~.
\end{align*}
Thus we are able to explicitly compute the upper bound in Eq. (\ref{eq:UpBo}) and the value of each of the $m$ components of the gradient $\partial \mathcal{R}(\bm{x}, \bm{\theta}, \bm{a})/ \partial \bm{\theta}_j^2$ is our saliency score indicating the importance of each parameter $\bm{\theta}_j$ in composing the trace of the NTK. To summarize, the final PX saliency score is:
\begin{equation}\label{eq:saliency_PX}
    S_{\text{PX}}(\bm{x}, \bm{\theta}, \bm{a}) = \frac{\partial \mathcal{R}(\bm{x}, \bm{\theta}, \bm{a})}{\partial \bm{\theta}^2} \odot \bm{\theta}^2.
\end{equation}
As we perform global masking, we can observe from Eq. (\ref{eq:Jv}), (\ref{eq:Jtheta}) and (\ref{eq:saliency_PX}) that our saliency function yields only positive scores which means that the saliency among layers is conserved. Combined with the iterative application of our pruning procedure, we satisfy the hypotheses of the Theorem of Maximal Critical Compression \cite{tanaka2020synflow}, which allows us to avoid \textit{layer collapse}, namely pruning all neurons within one layer and preventing the information flow.

The full PX algorithm pseudocode is provided in the supplementary material.

\section{Experiments}

In this section, we describe the results of our experimental analysis that thoroughly compares our PX with several baseline methods. In terms of datasets, tasks, and architectures we align with the literature and adopt well-established setups that are briefly summarized in the following \cite{tanaka2020synflow, lee2018snip, frankle2021missing, frankle2018lottery, chen2021lottery}. Moreover, we investigate whether PaI can be applied to pre-trained models without damaging their downstream transferability. Our empirical evaluation provides a positive answer to this new research question.

\smallskip
\noindent\textbf{Datasets \& Tasks.} For the classification experiments, we use CIFAR-10, CIFAR-100 \cite{krizhevsky2009learning}, Tiny-ImageNet and ImageNet \cite{deng2009imagenet}. For the segmentation experiments we follow \cite{chen2021lottery} and use the training and validation splits of Pascal VOC 2012 \cite{everingham2015pascal} for model learning and evaluation.

\smallskip
\noindent\textbf{Architectures.}
As done by \cite{ntksap2023wang, tanaka2020synflow}, for the classification experiments we use ResNet-20 \cite{he2016deep} on CIFAR-10, VGG-16 \cite{simonyan2014very} on CIFAR-100. ResNet-18 on Tiny-ImageNet and ResNet50 on the ImageNet dataset. By following \cite{chen2021lottery}, on the segmentation task we use DeepLabV3+ \cite{chen2018encoder} with ResNet-50.

\smallskip
\noindent\textbf{Initialization.} As in \cite{frankle2021missing} we initialize each model using Kaiming normal initialization \cite{he2015delving}. Furthermore, we assess how pre-trained parameters affect the foresight pruning procedure. For this analysis we adopt a setting analogous to that in \cite{chen2021lottery} that originally considered only iterative unstructured magnitude pruning. Specifically, we use a ResNet-50 pre-trained on ImageNet as well as two self-supervised models obtained with MoCov2 \cite{chen2020improved} and CLIP \cite{radford2021learning}.

\smallskip
\noindent\textbf{Implementation details.} Regarding the training procedure we follow \cite{frankle2021missing} and \cite{chen2021lottery} when assessing respectively our PX with respect to the PaI state-of-the-art methods and the pre-training transferability.
We evaluate each algorithm on trivial (36.00\%, 59.04\%, 73.80\%), mild (83.22\%, 89.30\%, 93.12\%) and extreme (95.60\%, 97.17\%, 98.20\%) sparsity ratios as \cite{ntksap2023wang}. We use 100 rounds for iterative PaI methods adopting an exponential schedule as \cite{tanaka2020synflow, frankle2021missing}. Full implementation details can be found in the supplementary.

\subsection{Classification with Random Initialization}
\begin{figure*}
\centering
\includegraphics[width=0.33\linewidth]{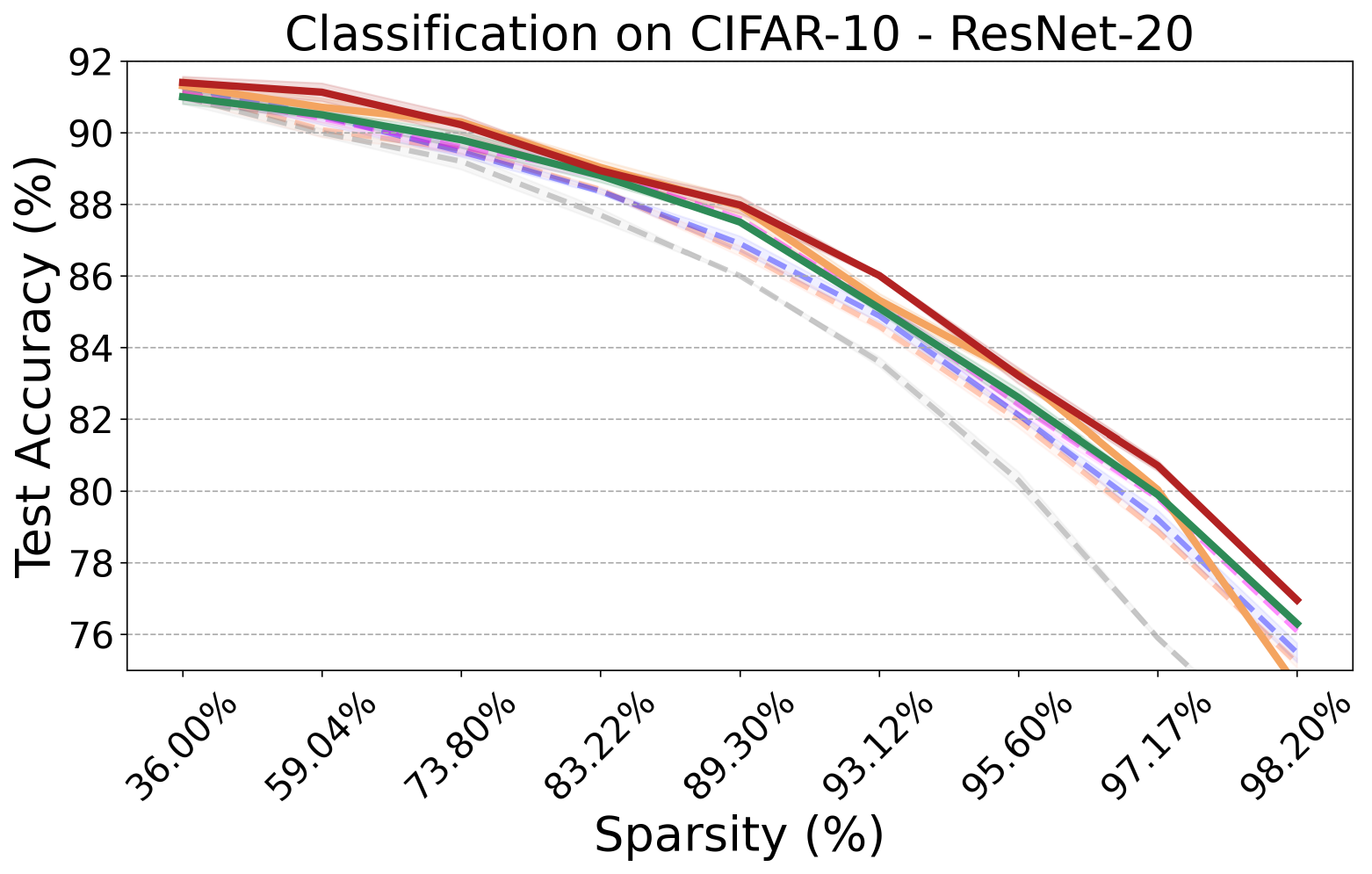}
\includegraphics[width=0.33\linewidth]{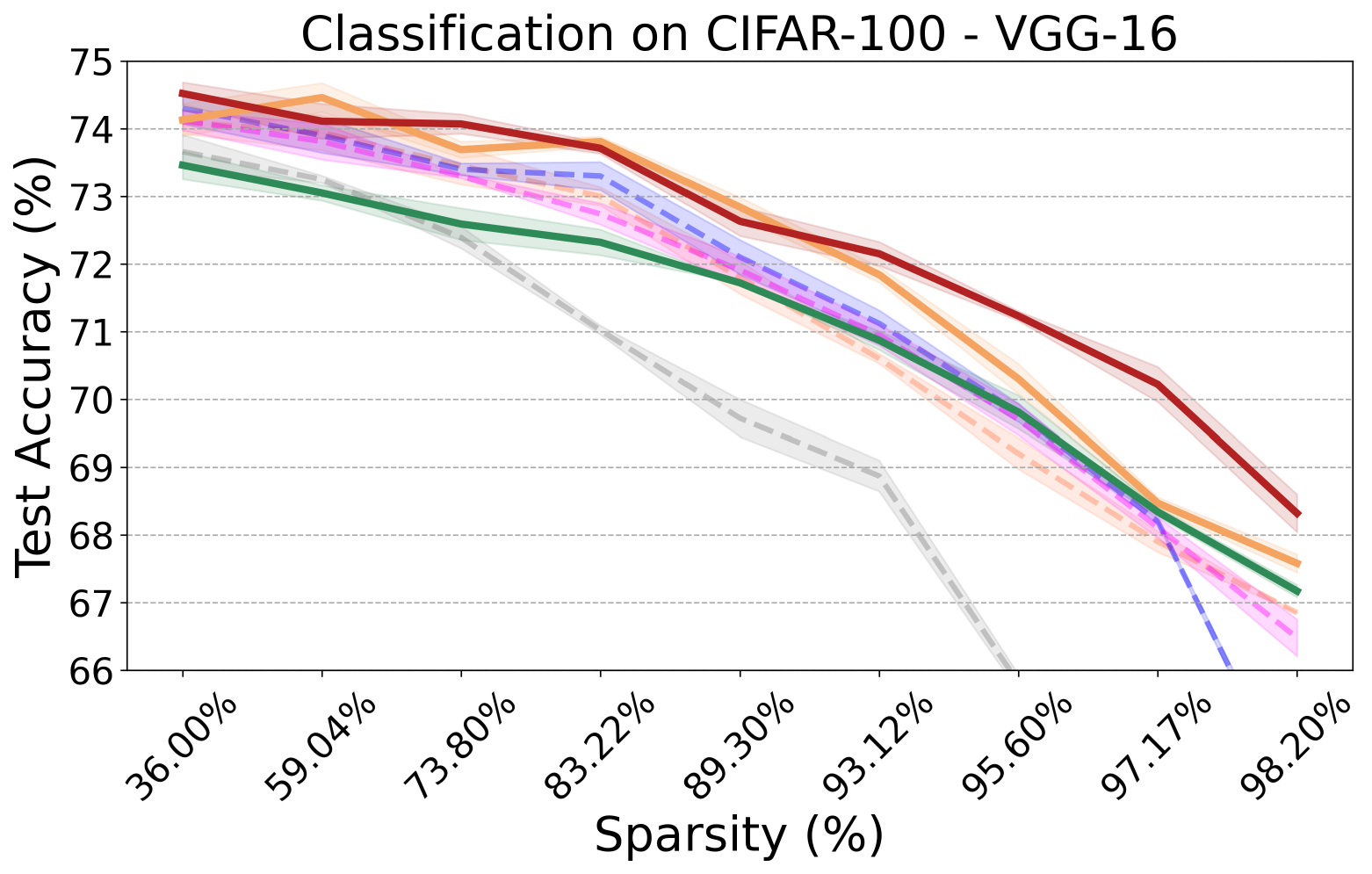}
\includegraphics[width=0.33\linewidth]{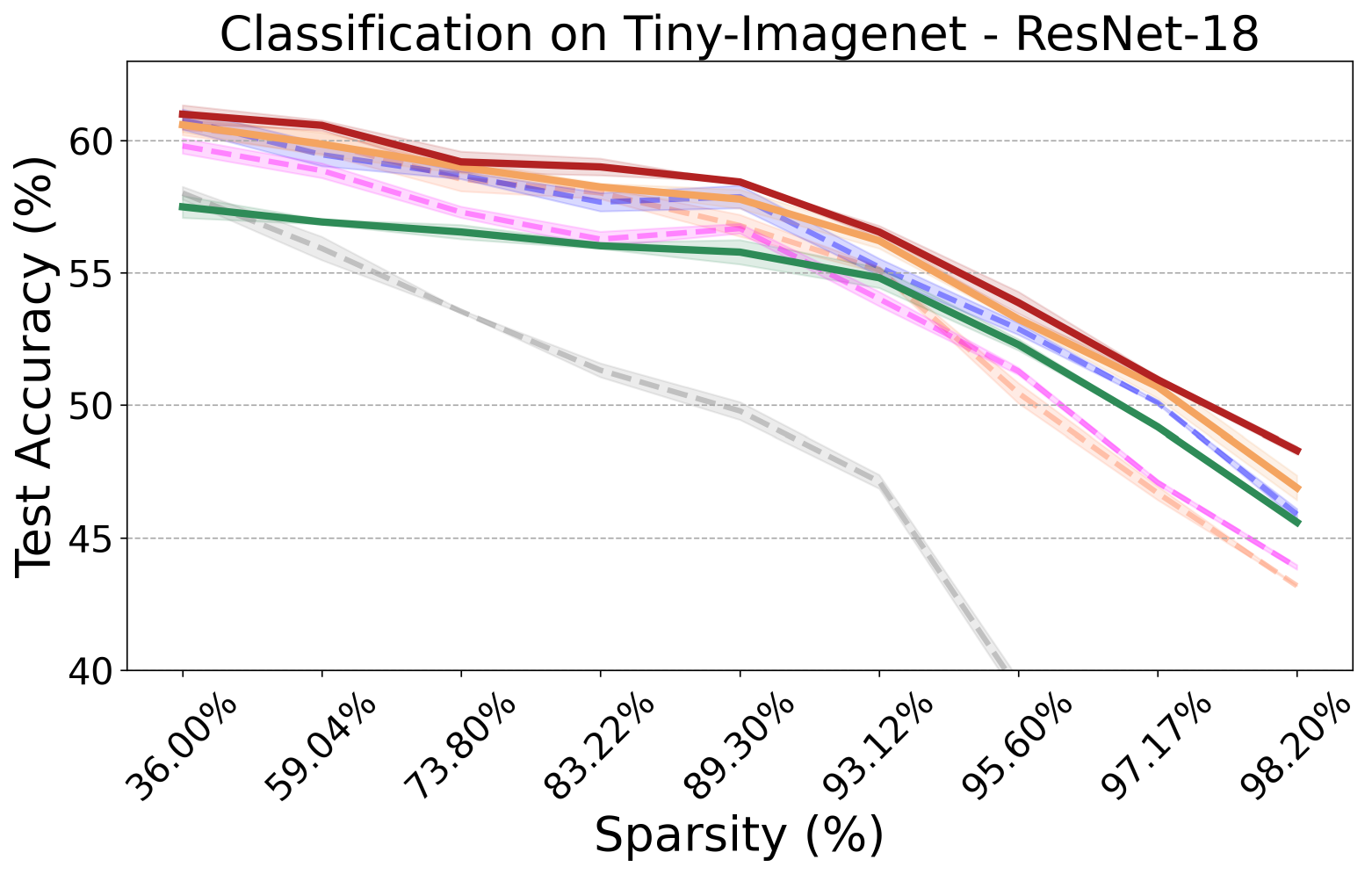}

\vspace{1mm}
\includegraphics[width=0.80\linewidth]{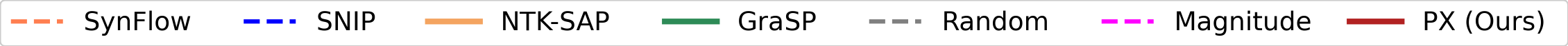}

\caption{Average classification accuracy at different sparsity levels on CIFAR-10 using ResNet-20, CIFAR-100 using VGG-16 and Tiny-ImageNet using ResNet-18, respectively. Each experiment is repeated three times. We report in shaded colors the standard deviation.}
\label{fig:classification_exp} \vspace{-5mm}
\end{figure*}

To provide an empirical evaluation of the strengths and effectiveness of our method, we compare PX with state-of-the-art foresight pruning algorithms. These include both data-driven methods like SNIP \cite{lee2018snip} and GraSP\cite{grasp2020wang}, as well as data-agnostic techniques such as SynFlow\cite{tanaka2020synflow} and NTK-SAP\cite{ntksap2023wang}. We also include two common baselines in PaI which are Random pruning and Magnitude-based pruning.

\smallskip
\noindent\textbf{CIFAR-10, CIFAR-100 \& Tiny-ImageNet.} In Fig. \ref{fig:classification_exp}. we report the classification results when using ResNet-20, VGG-16 and ResNet-18 on CIFAR-10, CIFAR-100 and Tiny-Imagenet, respectively.
For low sparsity levels, most techniques exhibit strong performance. On ResNet-20 (CIFAR-10), the performance gap across techniques is low, albeit being slightly higher for random pruning and GraSP. On VGG-16 (CIFAR-100) and ResNet-18 (Tiny-ImageNet), a comparable pattern emerges, but the disparity in performance widens noticeably. In all three cases, our approach consistently delivers the highest results.
At moderate sparsity levels, PX and NTK-SAP start to emerge as the top performers across all three experiments. Specifically, on VGG-16 (CIFAR-100), these two methods showcase competitive performance, with NTK-SAP being slightly ahead. The ranking of the other techniques remains consistent at these levels.
Finally, at extreme levels of sparsity, PX clearly outperforms all the competitors. In particular, on VGG-16 (CIFAR-100), there is a considerable disparity between PX and all other methods, including NTK-SAP. Notably, there's a substantial decline in NTK-SAP's performance at 98.20\% sparsity on ResNet-20 (CIFAR-10). Across all three experiments, GraSP, despite initially exhibiting lower performance, demonstrates commendable consistency in maintaining its results.

\smallskip
\noindent\textbf{ImageNet.} We conducted a comprehensive assessment of PX on larger-scale datasets, specifically ImageNet, employing ResNet-50 as the backbone model. In line with \cite{ntksap2023wang}, we examined two sparsity levels (89.26\% and 95.60\%). Our findings, detailed in Table \ref{tab:imagenetpai}., reveal that PX, NTK-SAP, and GraSP get top results, with PX exhibiting a slight advantage. Magnitude pruning surprisingly demonstrates greater competitiveness compared to SNIP, which performed well on smaller-scale datasets but ranks last in this evaluation.

\begin{table}[ht]
    \begin{center}
    \small
    \setcellgapes{1.5pt}
    \makegapedcells
    \resizebox{0.35\textwidth}{!}{
    \begin{tabular}{c c c}
        \bottomrule      
        \multicolumn{3}{c}{Classification on ImageNet - ResNet-50}\\ \hline
        Pruning Method & 89.26\% & 95.60\% \\
        \hline
        SynFlow \cite{tanaka2020synflow} & 66.48 \scriptsize{$\pm$ 0.12}  & 59.41 \scriptsize{$\pm$ 0.19} \\
        SNIP \cite{lee2018snip} & 60.50 \scriptsize{$\pm$ 0.34} & 45.82 \scriptsize{$\pm$ 0.35} \\
        NTK-SAP \cite{ntksap2023wang} & \underline{67.98} \scriptsize{$\pm$ 0.31}  & 59.84 \scriptsize{$\pm$ 0.30} \\
        GraSP \cite{grasp2020wang} & 67.21 \scriptsize{$\pm$ 0.52}  & \underline{60.01} \scriptsize{$\pm$ 0.16} \\
        Random & 64.97 \scriptsize{$\pm$ 0.27} & 56.79 \scriptsize{$\pm$ 0.44} \\
        Magnitude & 66.56 \scriptsize{$\pm$ 0.23}  & 47.80 \scriptsize{$\pm$ 0.21} \\
        
        \textbf{PX (Ours)} & \textbf{68.11} \scriptsize{$\pm$ 0.29} & \textbf{60.28} \scriptsize{$\pm$ 0.32} \\
        
        \bottomrule
    
    \end{tabular}}
    \end{center}
    \vspace{-5mm}
    \caption{Average classification accuracy at different sparsity ratios on the ImageNet dataset, using Kaiming normal initialized ResNet-50 as backbone. Each experiment is repeated three times. We report also the standard deviation. \textbf{Bold} indicates the best result. \underline{Underline} the second best.}
    \vspace{-5mm}
    \label{tab:imagenetpai}
\end{table}

\subsection{Starting From Pre-Trained Parameters}
\begin{figure*}
\centering
\includegraphics[width=0.33\linewidth]{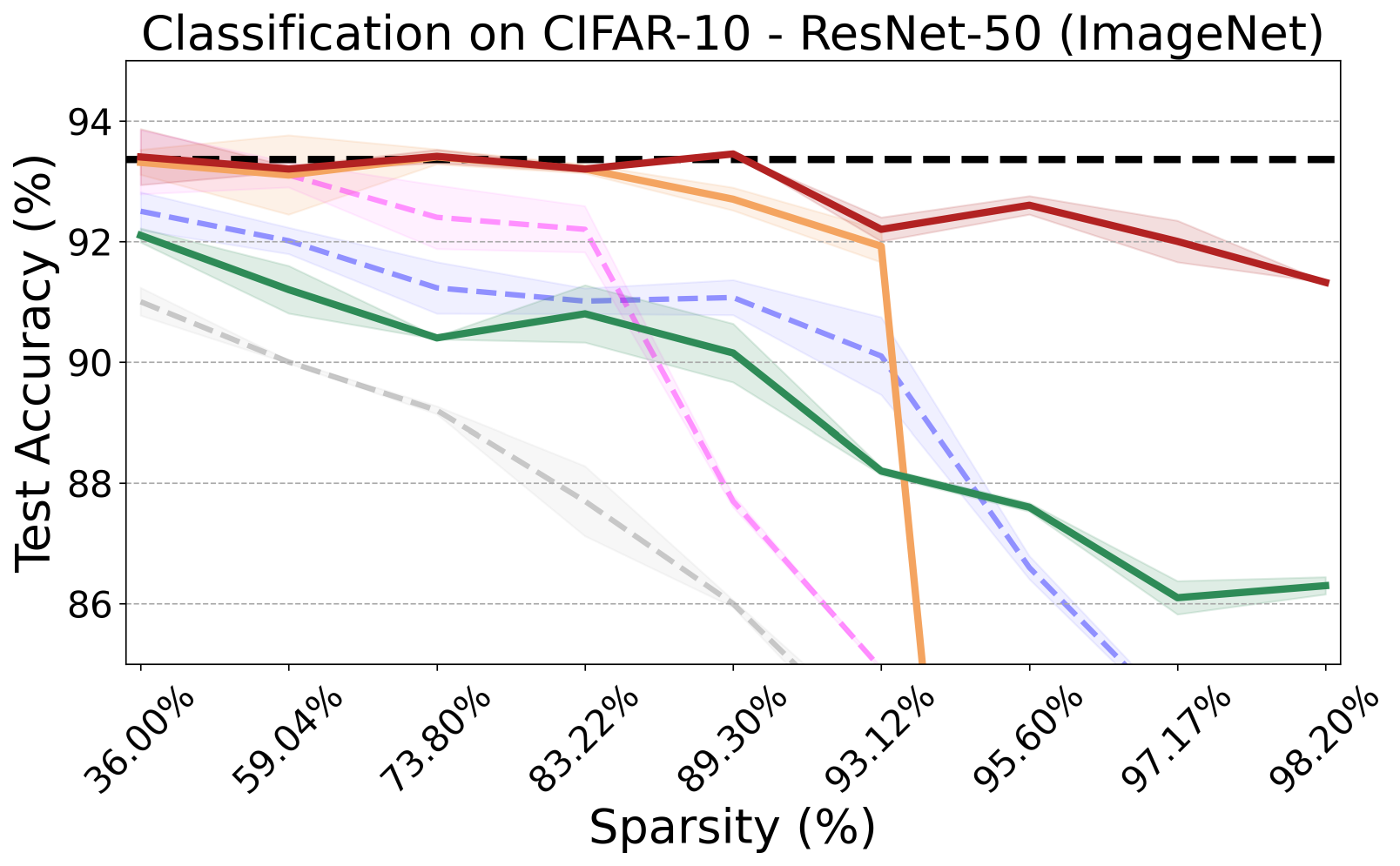}
\includegraphics[width=0.33\linewidth]{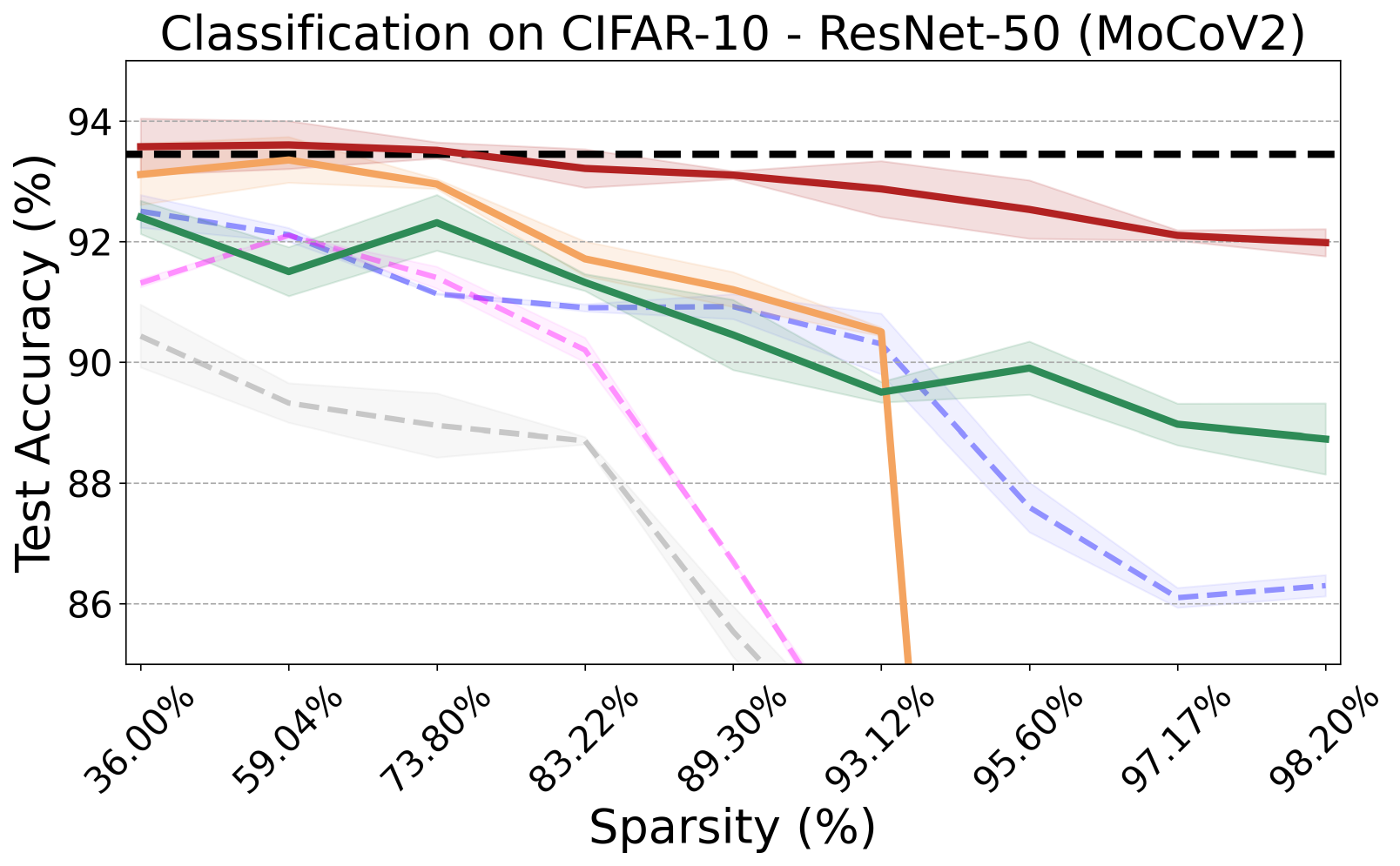}
\includegraphics[width=0.33\linewidth]{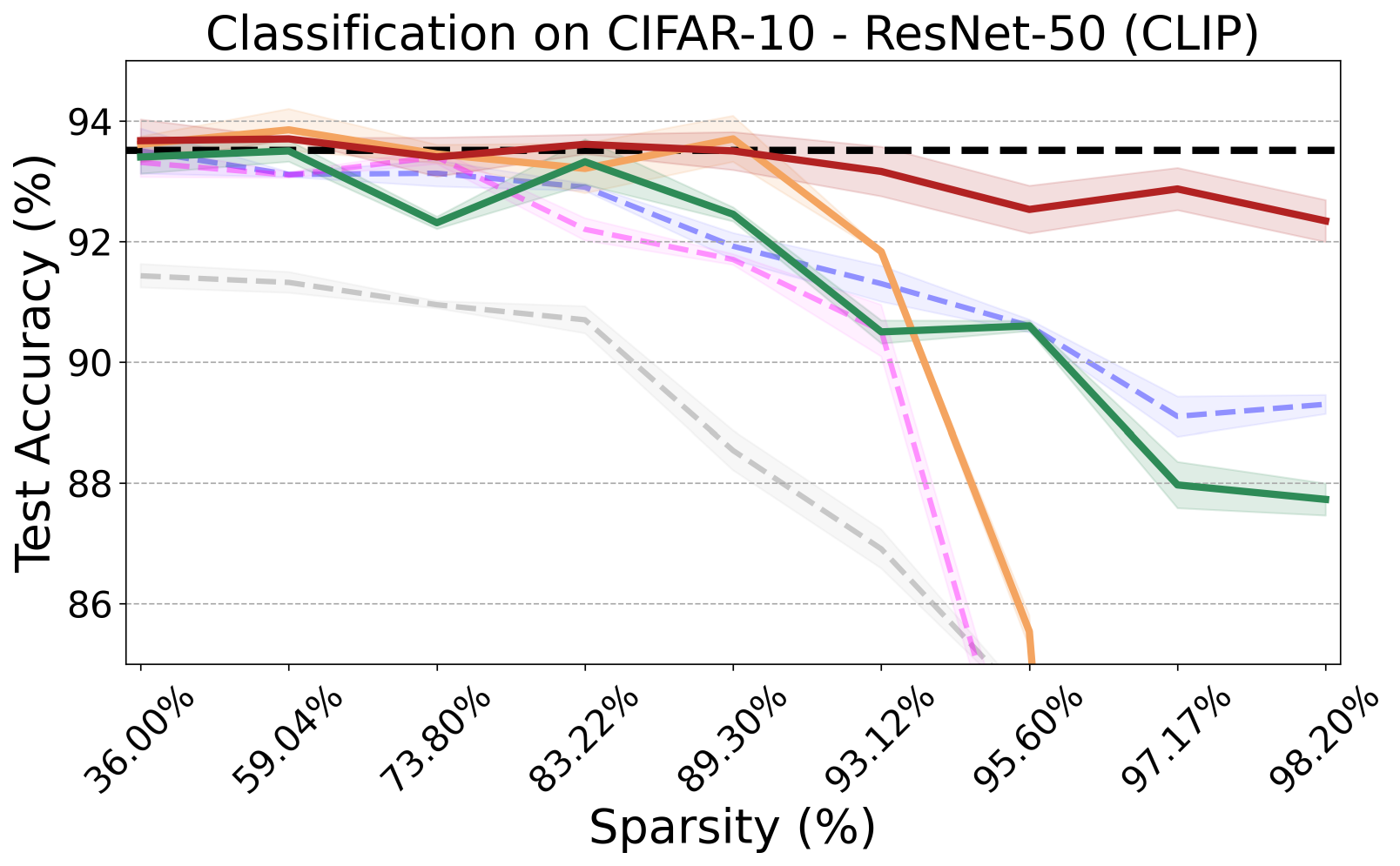}

\includegraphics[width=0.33\linewidth]{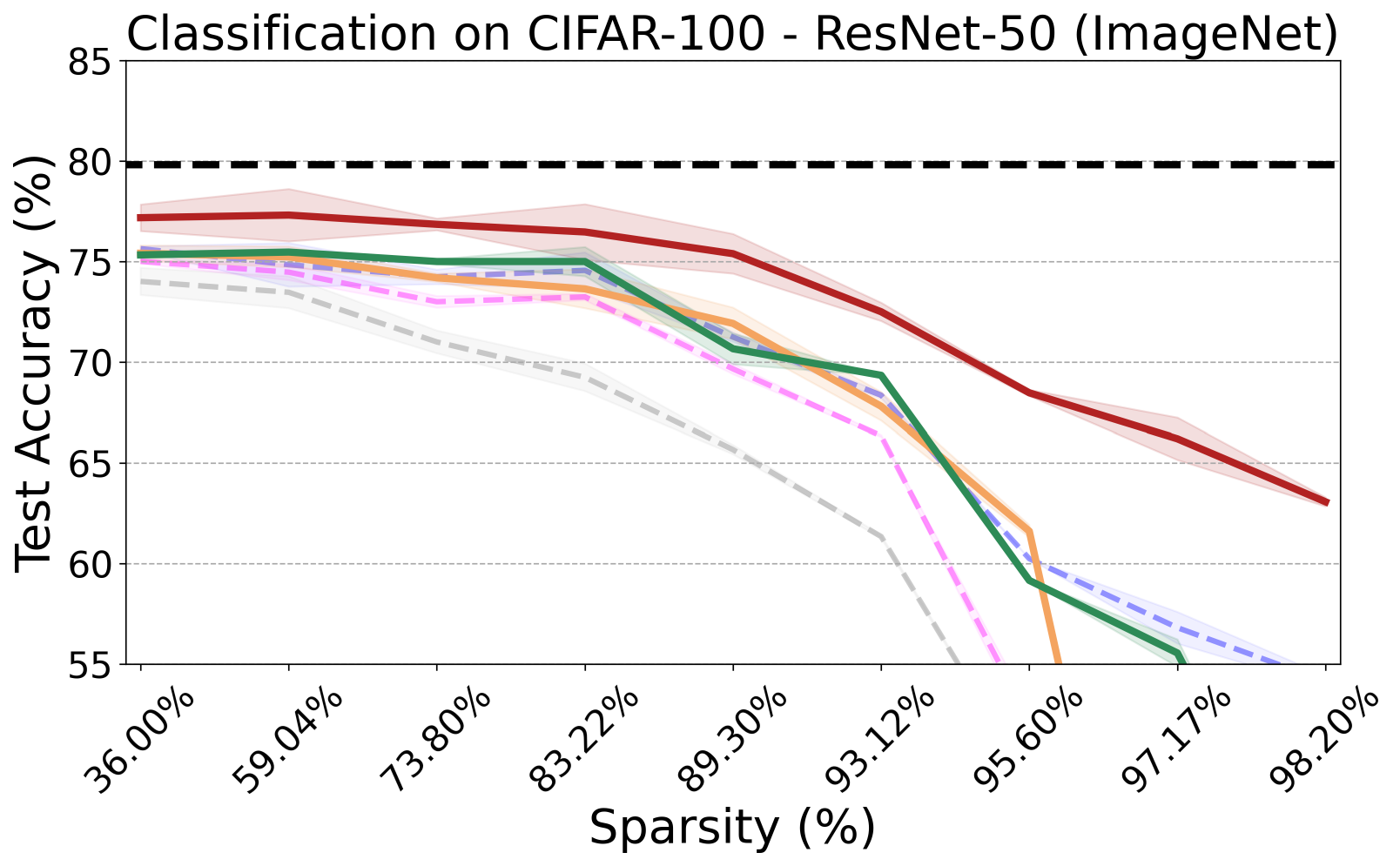}
\includegraphics[width=0.33\linewidth]{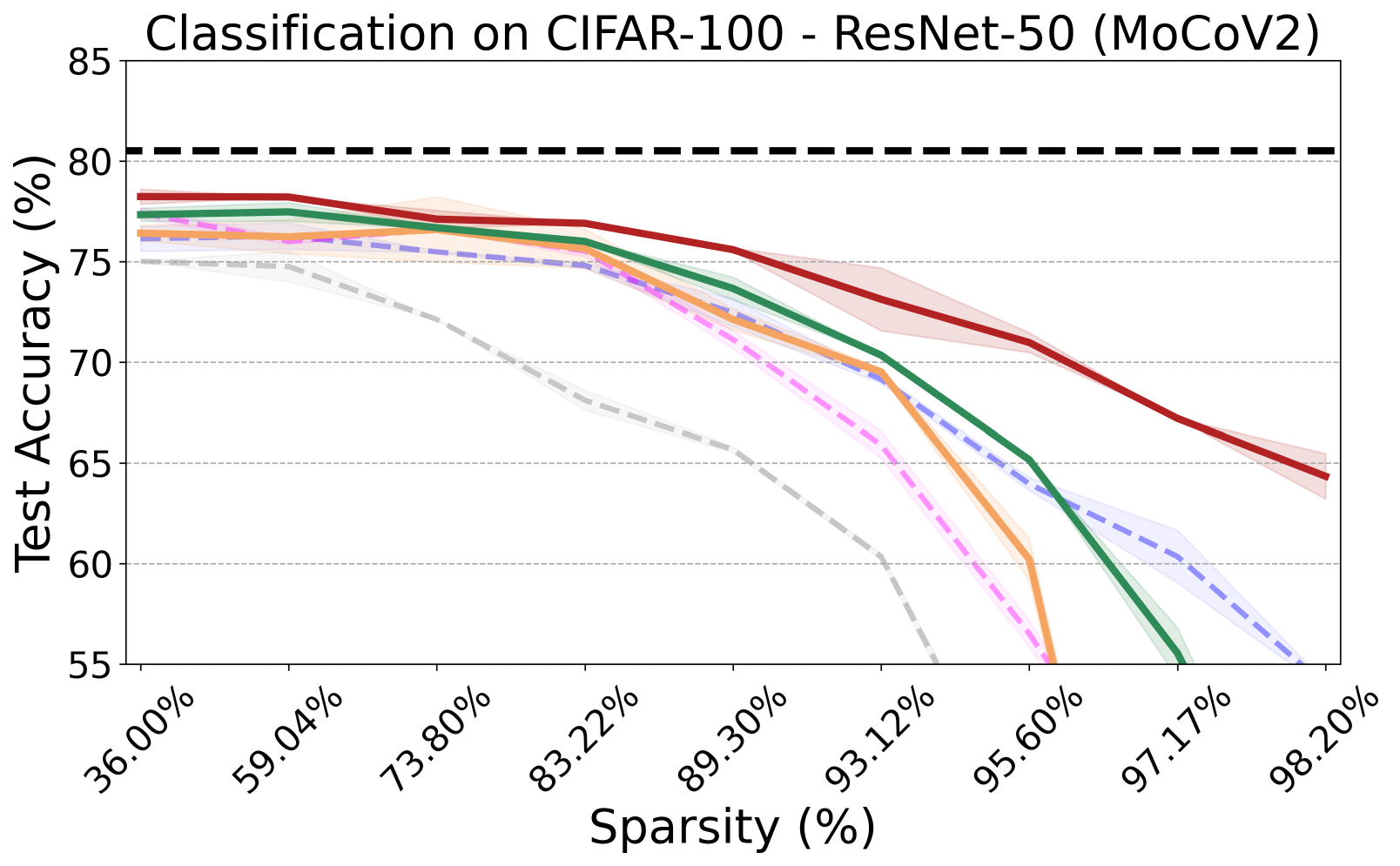}
\includegraphics[width=0.33\linewidth]{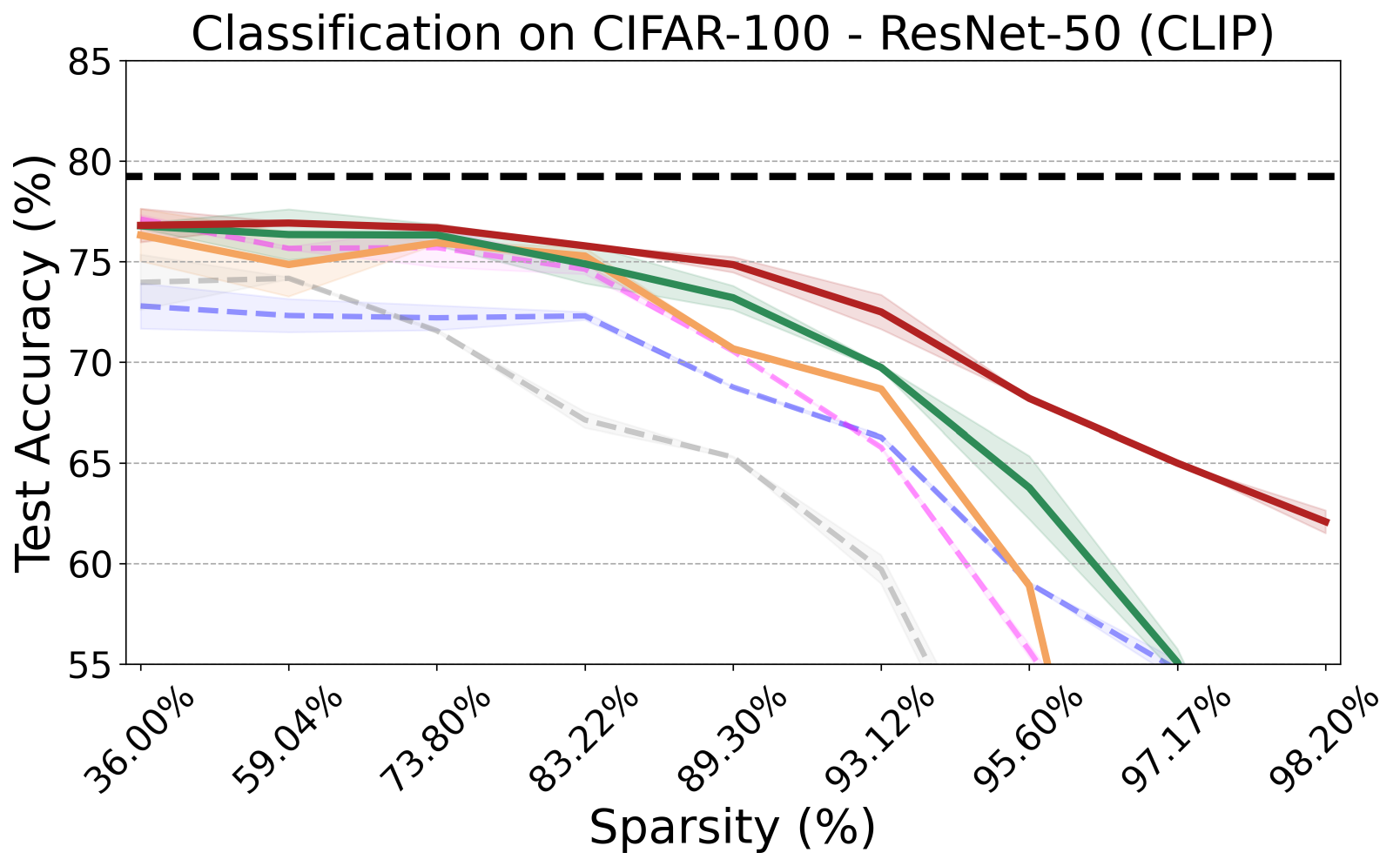}

\includegraphics[width=0.33\linewidth]{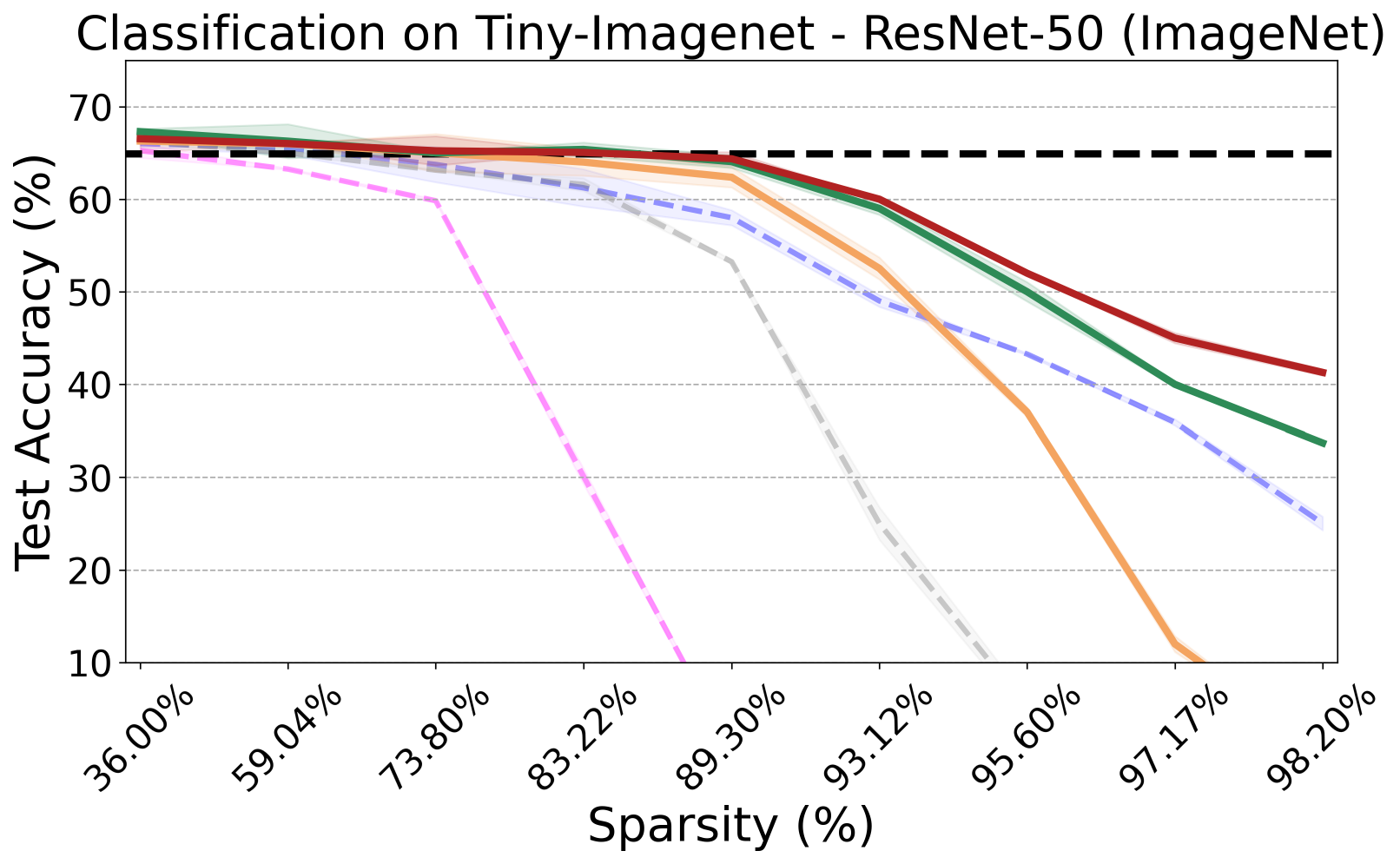}
\includegraphics[width=0.33\linewidth]{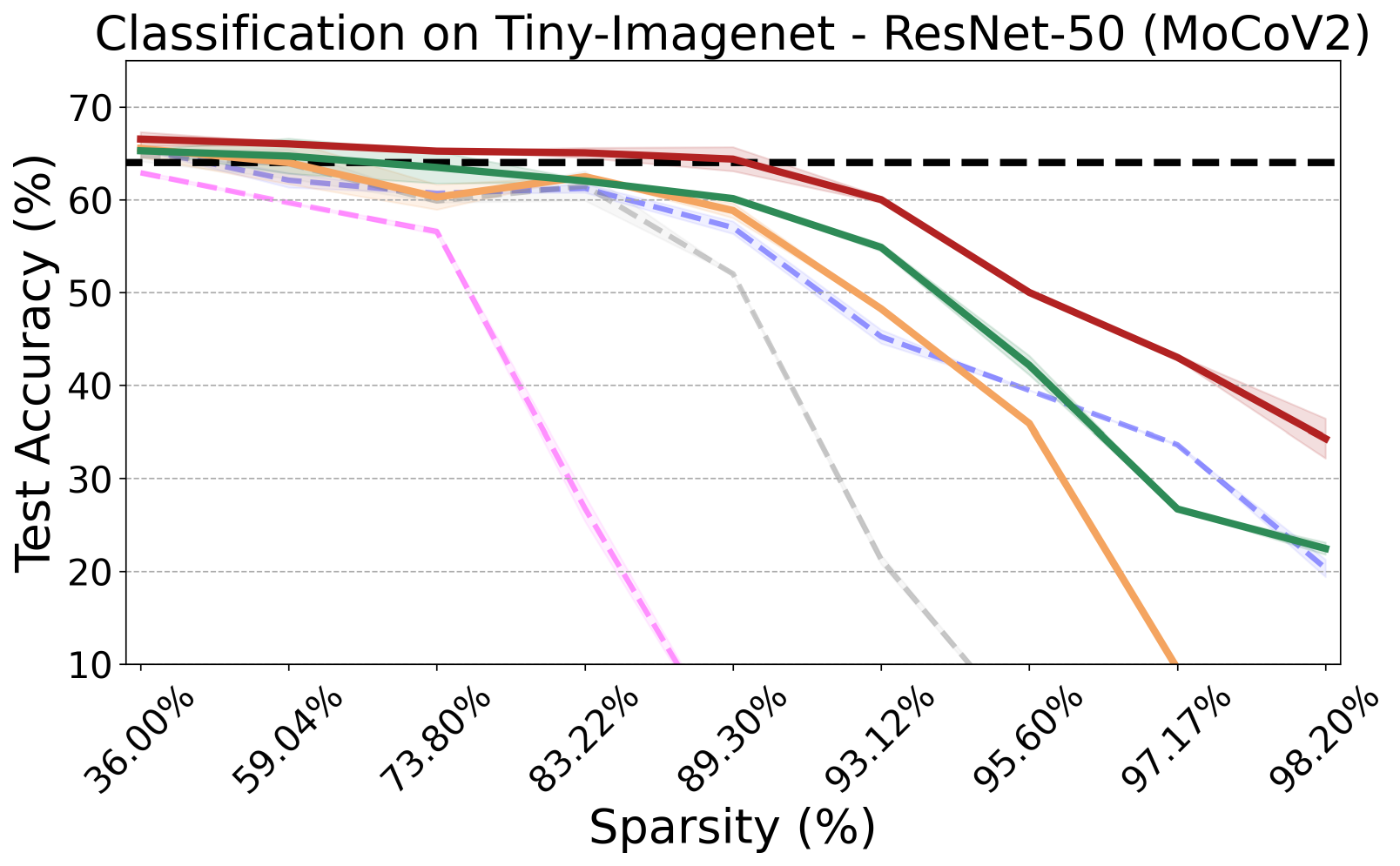}
\includegraphics[width=0.33\linewidth]{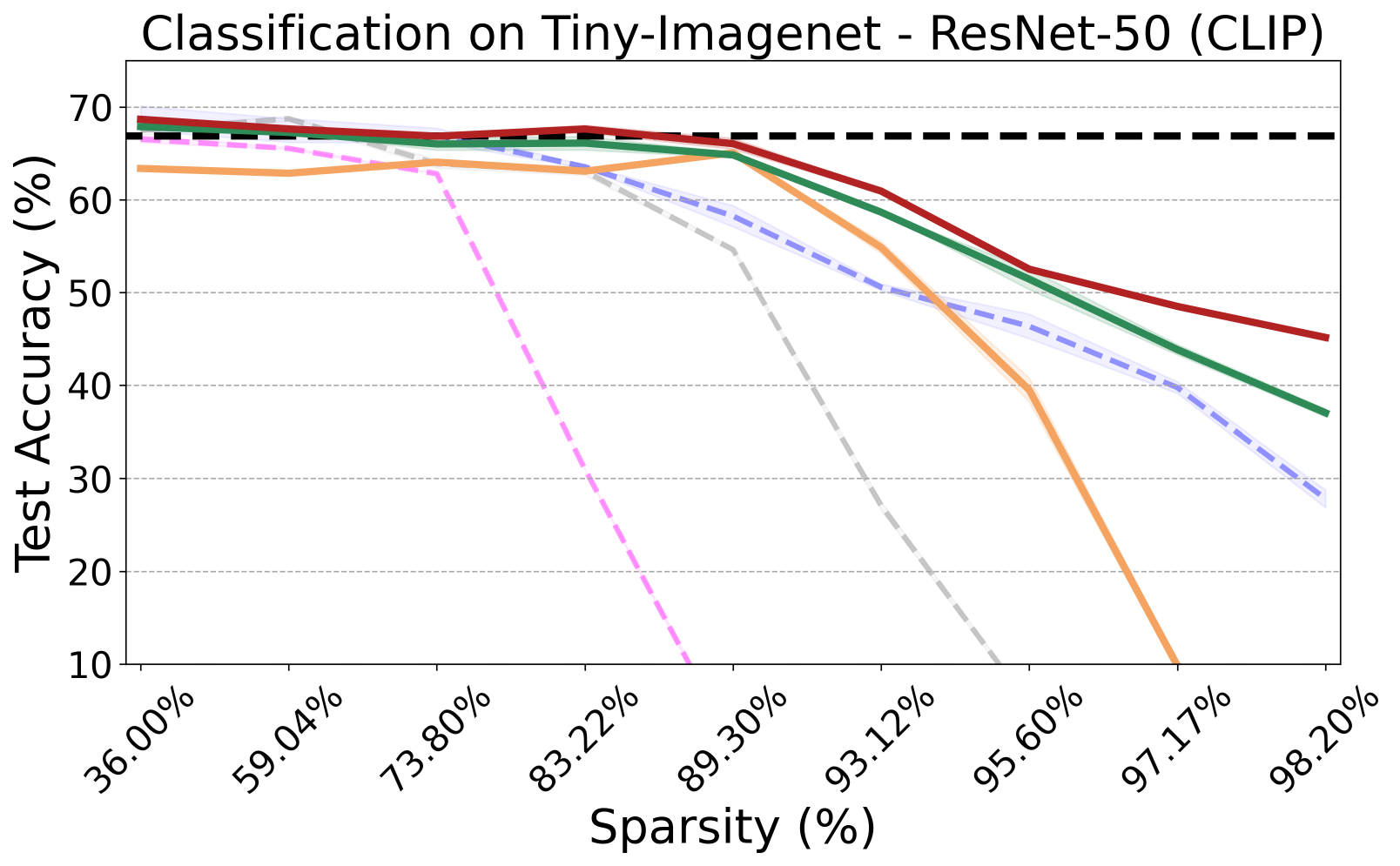}

\vspace{1mm}
\includegraphics[width=0.80\linewidth]{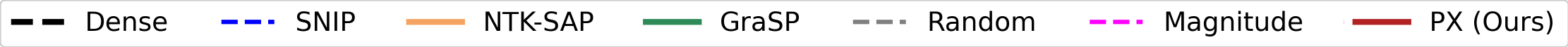}

\caption{Average classification accuracy at different sparsity levels on CIFAR-10, CIFAR-100 and Tiny-ImageNet using pre-trained ResNet-50 as architecture. The first column reports the results of starting from the supervised ImageNet pre-training. The second column reports the performance when starting from the MoCov2 pre-training on ImageNet. Finally, in the third column we report the results when starting from CLIP. Each experiment is repeated three times. We report in shaded colors the standard deviation.
}
\label{fig:pretrain_exp} \vspace{-5mm}
\end{figure*}

\begin{figure}\vspace{2mm}
\centering
\vspace{-1
mm}
\includegraphics[width=0.90\linewidth]{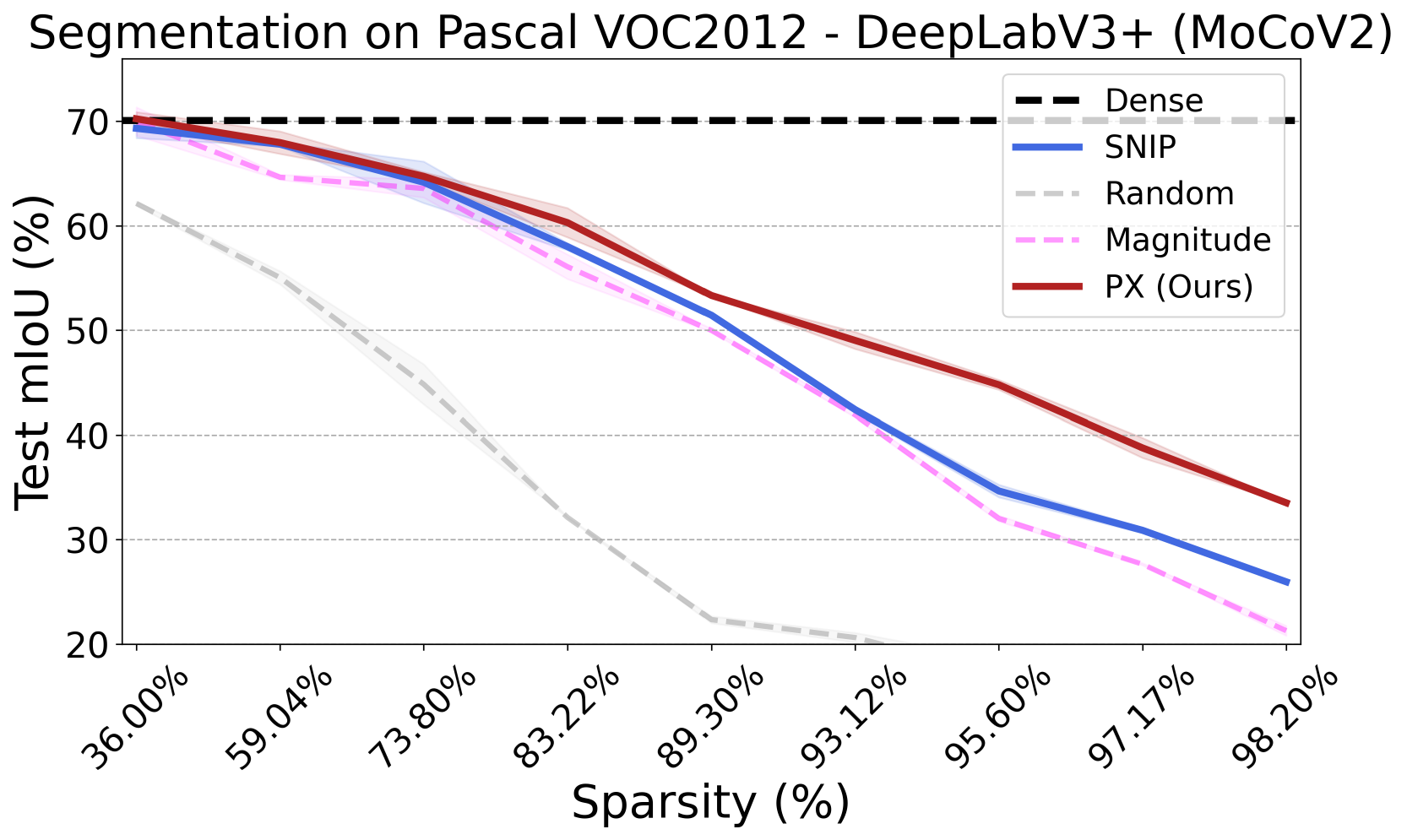}

\vspace{-3mm}

\caption{Average mean Intersection over Union (mIoU) at different sparsity levels on Pascal VOC2012 using DeepLabV3+ with pre-trained ResNet-50 as the backbone. Each experiment is repeated three times.
Standard deviations are in shaded colors.
}
\label{fig:segmentation_exp} \vspace{-6mm}
\end{figure}

Within this section, we examine the impact of initialization from pre-trained models: we aim at gaining insights into how we can leverage PaI algorithms to efficiently transfer knowledge to downstream tasks.

\smallskip
\noindent\textbf{Classification.} Fig. \ref{fig:pretrain_exp}. presents our findings on classification results across CIFAR-10, CIFAR-100, and Tiny-ImageNet using ResNet-50 initialized from ImageNet, MoCov2 on ImageNet, and CLIP pre-trainings. 
We do not report SynFlow here as it produces exploding gradients while estimating saliency scores.
This issue has been also observed in contexts like Neural Architecture Search \cite{cavagnero2023freerea}.

Until reaching extreme sparsity, most methods closely align with the performance of the dense baseline, underlining that employing PaI in this context serves as a viable, cost-free alternative to Iterative Magnitude Pruning (IMP). At extreme sparsity levels across all experiments, PX consistently outperforms other methods maintaining proximity to the dense baseline on simpler tasks (CIFAR-10) and showing a clear advantage over the competitors on more complex tasks (CIFAR-100 and TinyImageNet).

We remark that, while NTK-SAP stands as a state-of-the-art method, its performance drastically diminishes at extreme sparsity levels when initialized from pre-trained parameters. We attribute this decline to the interference of random mini-batches with the batch normalization statistics of the pre-trained model during its saliency estimation. This provides clear evidence of the limitations of data-free PaI methods.  

\smallskip
\noindent\textbf{Segmentation.} In Fig. \ref{fig:segmentation_exp}., we present the semantic segmentation results on the Pascal VOC2012 datasets, employing DeepLabV3+ on ResNet-50 initialized with MoCov2 on ImageNet (further results with other initialization in the supplementary).
Here we report only the results for our method, SNIP, Random, and Magnitude-based pruning. This selective reporting stems from issues encountered with other methodologies: SynFlow faced again challenges with exploding gradients, NTK-SAP resulted in layer collapse within the segmentation head due to the potential absence of positive saliency scores, a crucial factor in preventing such collapses. Similarly, GraSP, relying on a single round of pruning, encountered limitations in its applicability.

PX confirms its superiority to the other methods even for semantic segmentation. We also note that SNIP consistently demonstrates good performance and it appears as a remarkable result in comparison to the failure of NTK-SAP.
Overall, despite the encouraging results, it is apparent that significant effort needs to be directed toward PaI techniques specifically tailored for more complex vision tasks such as semantic segmentation. As of now, PaI methods only approximate the dense network results at trivial sparsity levels.
Surprisingly, Magnitude-based pruning also finds competitive subnetworks, comparable to other PaI methods, but only at trivial and moderate sparsity levels.

\subsection{Spectral Analysis of the Fixed-Weight-NTK}
\begin{figure}
\centering
\includegraphics[width=0.87\linewidth]{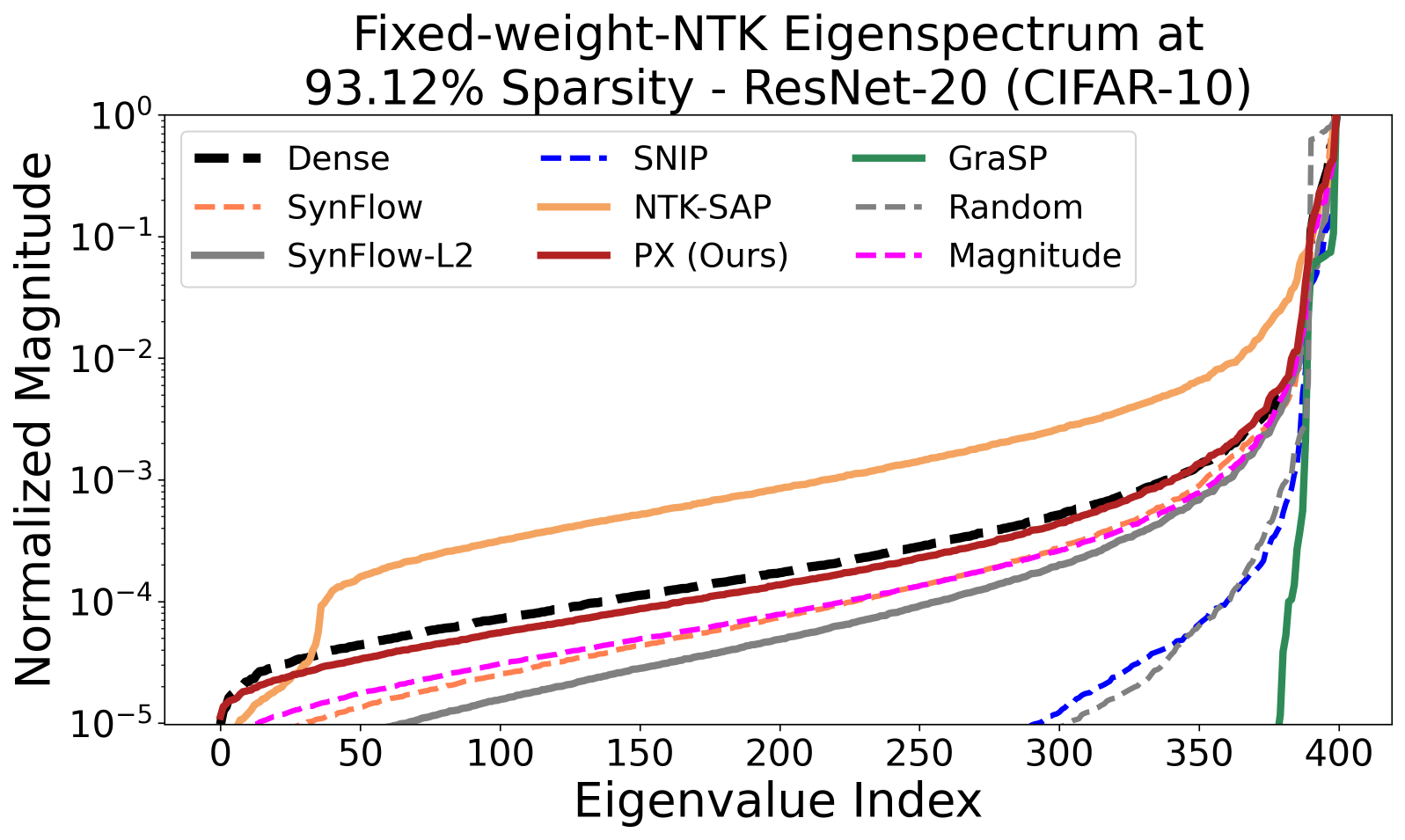} %0.9
\vspace{-3mm}
\caption{Fixed-Weight-NTK spectrum of ResNet-20 on the CIFAR-10 dataset at 93.12\% sparsity ratio.}
\label{fig:eigenspecturm} \vspace{-3mm}
\end{figure}

We designed PX to preserve the parameters that mostly influence the eigenspectrum of the NTK. To verify this behavior we visualize in Fig. \ref{fig:eigenspecturm}. the eigenvalues distributions of the Fixed-Weight-NTK\footnote{Note that in the limit of infinite width the Fixed-Weight-NTK \cite{ntksap2023wang} approaches the Analytic NTK \cite{NTK_Jacot2018}.}.
Our investigation focuses on pruned ResNet-20 subnetworks on CIFAR-10, at 93.12\% sparsity. Notably, even at this substantial level of sparsity, our approach closely mirrors the eigenspectrum of the original dense network confirming the expectations.  
To make our argument even more solid, we disregarded the data-dependent term in PX which implies falling back to SynFlow-L2. %(in gray). 
As can be observed, data play a central role when preserving the eigenspectrum of the NTK. This ablation study further reinforces our claims.

\subsection{Layer Width}
\begin{figure}
\centering
\includegraphics[width=0.87\linewidth]{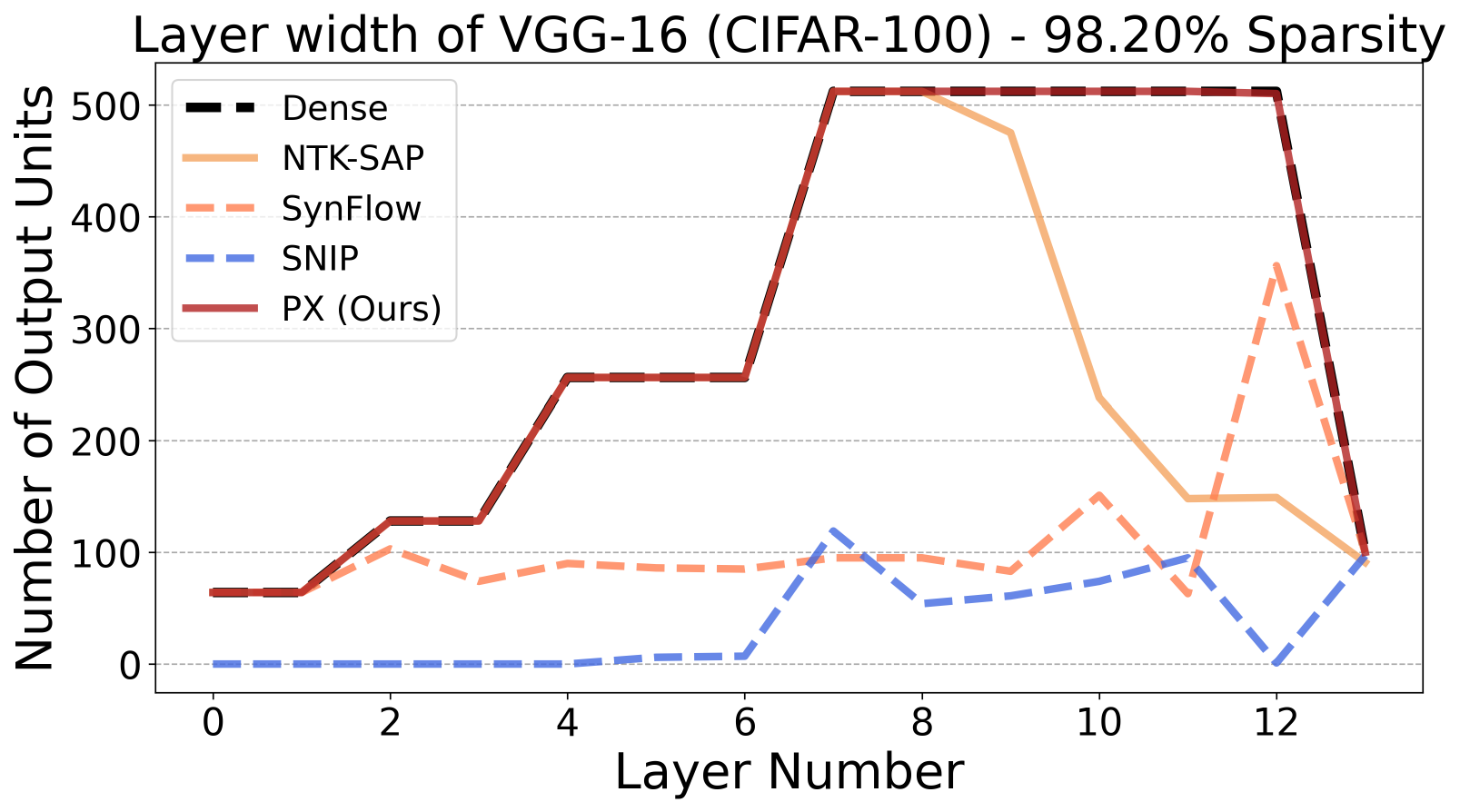}
\vspace{-3mm}
\caption{
Active output units at 98.20\% sparsity in VGG-16. For SNIP and PX data mini-batches are sampled from CIFAR-100. 
}
\label{fig:vgg_width} \vspace{-5mm}
\end{figure}

As discussed earlier, the integration of iterative pruning rounds and the layer-wise preservation of saliency are pivotal in preventing layer collapse using saliency functions \cite{tanaka2020synflow}. However, \cite{patil2021phew} observed that iterative PaI methods suffer a sudden reduction in the number of output units, leading to narrow layers and bottlenecks. Recent investigations \cite{golubeva2020wider} highlighted a correlation between network performance and higher output width when fixing a certain number of parameters in an architecture. Motivated by this, we investigate more in-depth the behavior of PX.

For fully connected layers every neuron constitutes an output unit.
For convolutional layers, we follow \cite{patil2021phew} and consider each kernel as an output unit. If all parameters within a convolutional kernel are pruned, the layer's output unit count is reduced by one. In Figure \ref{fig:vgg_width}, we conduct the analysis on the number of output units that each iterative PaI method preserves after pruning. Despite being iterative, PX is not affected by the issues mentioned in \cite{patil2021phew} and consistently preserves the output width of each layer.

\subsection{Execution Time Analysis}
\begin{table}[t!]
    \begin{center}
    \small
    \setcellgapes{1.5pt}
    \makegapedcells
    \resizebox{0.48\textwidth}{!}{
    \begin{tabular}{c|c@{~~}c|c@{~~}c|c@{~~}c|c@{~~}c|c@{~~}c||c@{~~}c}
        \bottomrule      
        \multicolumn{13}{c}{\textbf{ResNet-20 (CIFAR-10) - 98.20\% Sparsity}} \\ \hline
        & \multicolumn{2}{c|}{$T=1$} & \multicolumn{2}{c|}{$T=2$} & \multicolumn{2}{c|}{$T=10$} & \multicolumn{2}{c|}{$T=100$} & \multicolumn{2}{c||}{$T=200$} &  \multicolumn{2}{c}{$Epochs=960$} \\
        & $Acc$ & $Time$ & $Acc$ & $Time$ & $Acc$ & $Time$ & $Acc$ & $Time$ & $Acc$ & $Time$ & $Acc$ & $Time$ \\
        \hline\hline
        
        IMP  & - & - & - & - & - & - & - & - & - & - & 77.38 & 13708.80 \\ \hline
        
        \multicolumn{13}{c}{Single-shot PaI methods} \\ \hline
        
        Random  & 72.31 & 0.0003 & - & - & - & - & - & - & - & - & - & - \\
        Magnitude & 76.12 & 0.0006 & - & - & - & - & - & - & - & - & - & - \\ 
        SNIP \cite{lee2018snip} & 75.39 & 0.17 & - & - & - & - & - & - & - & - & - & - \\ 
        GraSP \cite{grasp2020wang} & 76.30 & 0.39 & - & - & - & - & - & - & - & - & - & - \\  \hline

        \multicolumn{13}{c}{Iterative PaI methods} \\ \hline
        
        SynFlow \cite{tanaka2020synflow} & 10.00 & 0.13 & 71.20 & 0.27 & 73.98 & 1.34 & 75.19 & 13.40 & 75.67 & 26.81 & - & - \\
        NTK-SAP \cite{ntksap2023wang} & 18.85 & 0.21 & 55.87 & 0.42 & 72.21 & 2.09 & 74.55 & 20.92 & 75.06 & 41.85 & - & - \\
        
        \textbf{PX (Ours)} & 76.15 & 0.14 & 76.47 & 0.28 & 77.04 & 1.41 & 77.08 & 14.13 & 77.39 & 28.27 & - & - \\
        
        \bottomrule
    
    \end{tabular}}
    \end{center}
    \vspace{-5mm}
    \caption{Comparison of the accuracy and execution time (in seconds) when varying the number of pruning rounds $T$, starting from ResNet-20 on CIFAR-10. The timings reported refer to the 95th percentile of 100 measurements.} \vspace{-4mm}
    \label{tab:tab_T_ablation}
\end{table}
\vspace{-1mm}
In Table \ref{tab:tab_T_ablation}, we show the effect of changing the number of pruning rounds $T$, presenting accuracy and total execution time of the pruning procedures in seconds. Here we include IMP to offer a broader context to our study. 
We ran our evaluation on NVIDIA Titan Xp GPU, Intel i7-9800X CPU, and using the \texttt{perf\_counter} clock from Python's \texttt{time} module.
IMP ($Epochs=960$) takes nearly 4 hours to outperform PX ($T=10$).
GraSP leads at $T=1$, but PX surpasses GraSP at $T=2$ without exceeding its time cost. 
Increasing $T$ beyond $100$ marginally improves results but does not alter conclusions, in line with \cite{tanaka2020synflow, ntksap2023wang}. 
More discussions about the computational cost of PX and its competitors are provided in the supplementary.

\section{Conclusion}
\vspace{-1mm}
Pruning at initialization offers the attractive possibility of reducing the number of parameters in a neural network, bypassing the need for training to identify the pruning mask. The NTK and its pathwise decomposition provide a powerful proxy for identifying parameters that are important for preserving training dynamics after pruning. While most methods consider only the data-independent component, we propose a new upper bound on the trace of the NTK which led to Path eXclusion (PX), that allows us to preserve its spectrum and consider the data-dependent component as well. We show experimentally that PX is not only robust to different architectures and tasks but can also be effectively used to search for subnetworks in large pre-trained models that retain almost intact transferability.

\smallskip
\noindent\textbf{Acknowledgements.} 
\small{L.I. acknowledges the grant received from the European Union Next-GenerationEU (Piano Nazionale di Ripresa E Resilienza (PNRR)) DM 351 on Trustworthy AI. T.T. acknowledges the EU project ELSA - European Lighthouse on Secure and Safe AI.
This study was carried out within the FAIR - Future Artificial Intelligence Research and received funding from the European Union Next-GenerationEU (PIANO NAZIONALE DI RIPRESA E RESILIENZA (PNRR) – MISSIONE 4 COMPONENTE 2, INVESTIMENTO 1.3 – D.D. 1555 11/10/2022, PE00000013). 
This manuscript reflects only the authors’ views and opinions, neither the European Union nor the European Commission can be considered responsible for them.
}

{\small
\bibliographystyle{ieee_fullname}
\bibliography{egbib}
}

\newpage
\appendix

\begin{table*}[btp]
    \begin{center}
    \small
    \setcellgapes{1.5pt}
    \makegapedcells
    \resizebox{0.98\textwidth}{!}{
    \begin{tabular}{c c c c c c c c c c}
        \bottomrule      
        \textbf{Dataset} & \textbf{Architecture} & \textbf{Optimizer} & \textbf{LR} & \textbf{LR Drop} & \textbf{Momentum} & \textbf{Weight Decay} & \textbf{Epochs} & \textbf{Batch Size} \\
        \hline\hline
        \multicolumn{9}{c}{\textbf{Classification - Kaiming Normal initialization \cite{he2015delving}}} \\ \hline

        CIFAR-10 \cite{krizhevsky2009learning} & ResNet-20 \cite{he2016deep} & SGD \cite{ruder2016overview} & 0.1 & x10 at epochs 80, 120 & 0.9 & 1e-4 & 160 & 128 \\

        CIFAR-100 \cite{krizhevsky2009learning} & VGG-16 \cite{simonyan2014very} & SGD (Nesterov) \cite{ruder2016overview} & 0.1 & x10 at epochs 60, 120 & 0.9 & 5e-4 & 160 & 128 \\

        Tiny-ImageNet \cite{deng2009imagenet} & ResNet-18 \cite{he2016deep} & SGD \cite{ruder2016overview} & 0.2 & x10 at epochs 100, 150 & 0.9 & 1e-4 & 200 & 256 \\

        ImageNet \cite{deng2009imagenet} & ResNet-50 \cite{he2016deep} & SGD \cite{ruder2016overview} & 0.1 & x10 at epochs 30, 60, 80 & 0.9 & 1e-4 & 90 & 448 \\

        \hline
        \multicolumn{9}{c}{\textbf{Classification - ImageNet \cite{deng2009imagenet}, MoCov2 on ImageNet \cite{chen2020improved}, CLIP \cite{radford2021learning} pre-trained models}} \\ \hline

        CIFAR-10 \cite{krizhevsky2009learning} & ResNet-50 \cite{he2016deep} & SGD \cite{ruder2016overview} & 0.1 & x10 at epochs 91, 136 & 0.9 & 1e-4 & 182 & 256 \\

        CIFAR-100 \cite{krizhevsky2009learning} & ResNet-50 \cite{he2016deep} & SGD \cite{ruder2016overview} & 0.1 & x10 at epochs 91, 136 & 0.9 & 1e-4 & 182 & 256 \\

        Tiny-ImageNet \cite{deng2009imagenet} & ResNet-50 \cite{he2016deep} & SGD \cite{ruder2016overview} & 0.1 & x10 at epochs 91, 136 & 0.9 & 1e-4 & 182 & 256 \\

        \hline
        \multicolumn{9}{c}{\textbf{Segmentation - ImageNet \cite{deng2009imagenet}, MoCov2 on ImageNet \cite{chen2020improved}, DINO on ImageNet \cite{caron2021emerging} pre-trained models}} \\ \hline
        
        Pascal VOC2012 \cite{everingham2015pascal} & DeepLabV3+ (ResNet-50) \cite{chen2018encoder} & SGD \cite{ruder2016overview} & 0.001 & x10 at epochs 50, 60 & 0.9 & 1e-4 & 80 & 4 \\
        
        \bottomrule
    
    \end{tabular}}
    \end{center}
    \vspace{-5mm}
    \caption{Training setups used in this work.} \vspace{-2mm}
    \label{tab:setups}
\end{table*}

\section{Implementation Details}
In Table \ref{tab:setups} you can find the training details used in this work. We evaluate each algorithm on \texttt{trivial} (36.00\%, 59.04\%, 73.80\%), \texttt{mild} (83.22\%, 89.30\%, 93.12\%) and \texttt{extreme} (95.60\%, 97.17\%, 98.20\%) sparsity ratios as \cite{ntksap2023wang}. In each experiment, we use 100 rounds for iterative PaI methods adopting an exponential schedule as \cite{tanaka2020synflow, frankle2021missing}. We train and test on the respective official splits of each dataset, repeating each experiment three times.

\smallskip
\noindent\textbf{Classification - Random initialization.} For the classification experiments starting from Kaiming Normal initialization \cite{he2015delving}, we follow \cite{ntksap2023wang, tanaka2020synflow}. The augmentations used when training on CIFAR-10 and CIFAR-100 \cite{krizhevsky2009learning} are Random Crop to 32$\times$32 with padding 4 followed by Random Horizontal Flipping with 0.5 probability. For the experiments on Tiny-ImageNet \cite{deng2009imagenet}, we augment the training images with Random Resized Crop to 64$\times$64 with scaling going from 0.1 to 1.0 using 0.8 x-ratio and 1.25 y-ratio. Then, we apply Random Horizontal Flipping with 0.5 probability. On ImageNet \cite{deng2009imagenet}, we apply Random Resized Crop to 224$\times$224 with scaling going from 0.2 to 1.0 using 3/4 x-ratio and 4/3 y-ratio. Then, we apply Random Grayscaling with 0.2 probability, Color Jitter with brightness, contrast, saturation and hue all set to 0.4. Finally, we apply Random Horizontal Flipping with 0.5 probability.

\smallskip
\noindent\textbf{Classification - Pre-trained models.} Regarding the classification experiments when starting from ImageNet \cite{deng2009imagenet}, MoCov2 on ImageNet \cite{chen2020improved} and CLIP \cite{radford2021learning} pre-trained models, we align with \cite{chen2021lottery}. Specifically, we use the same augmentations detailed in the previous paragraph but we adjust the cropping and rescaling transformations to ensure that the resultant image size is set at 224$\times$224 pixels, aligning with the dimensions of the images used in obtaining the pre-trained models.

\smallskip
\noindent\textbf{Segmentation.} For the semantic segmentation experiments we again align with \cite{chen2021lottery}. We employ the following augmentations during training: Random Scale with a range between 0.5 and 2.0, Random Crop to 513$\times$513, followed by Random Horizontal Flipping with 0.5 probability.

\smallskip
\noindent\textbf{Pre-trained models \& Architectures.} Regarding the pre-trained models used in our experiments, we employed the official ImageNet pre-trained model from the PyTorch torchVision package \cite{torchvision2016}. The MoCov2 ImageNet model we used is the official one from Facebook research\footnote{\url{https://github.com/facebookresearch/moco}}. The CLIP pre-trained model is the official one from OpenAI\footnote{\url{https://github.com/openai/CLIP}}. Finally, we base our experiments on DINO \cite{caron2021emerging} from its officially released pre-trained model\footnote{\url{https://github.com/facebookresearch/dino}}.

Our code is based on the framework for Pruning-at-Initialization provided by \cite{tanaka2020synflow}. Moreover, we used their implementations for the architectures used in our classification experiments. For the segmentation experiments, we align with \cite{chen2021lottery} and use the same implementation of DeepLabV3+\footnote{\url{https://github.com/VainF/DeepLabV3Plus-Pytorch}}.

\smallskip
\noindent\textbf{Choice of the pruning set.} To perform the foresight pruning procedure using data-driven methods we employ a pruning data split composed of ten examples per class, in line with the work of \cite{lee2018snip, ntksap2023wang}. For the data-free strategies, we use an equal amount of mini-batches.
\begin{figure*}[tb]
\centering
\includegraphics[width=0.33\linewidth]{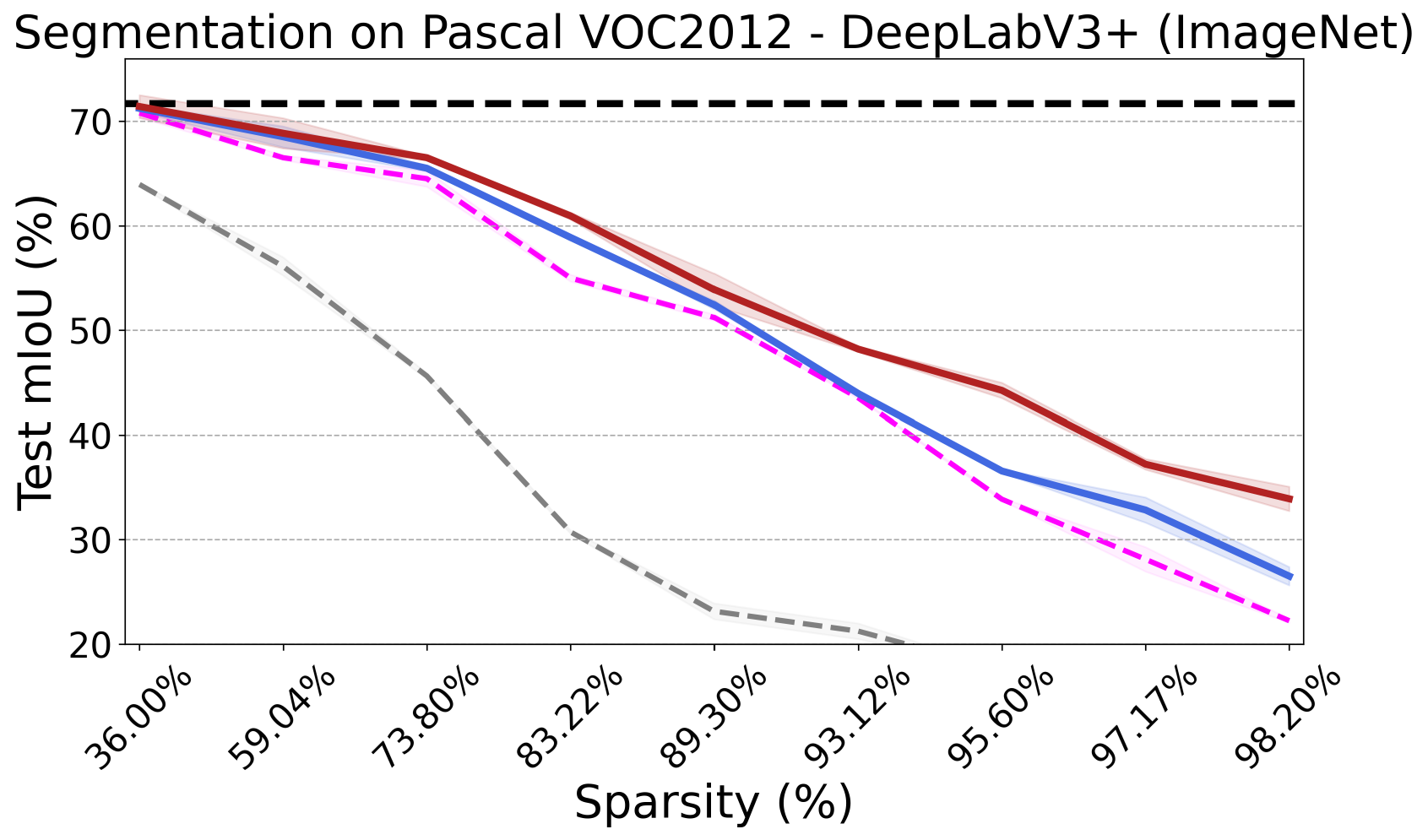}
\includegraphics[width=0.33\linewidth]{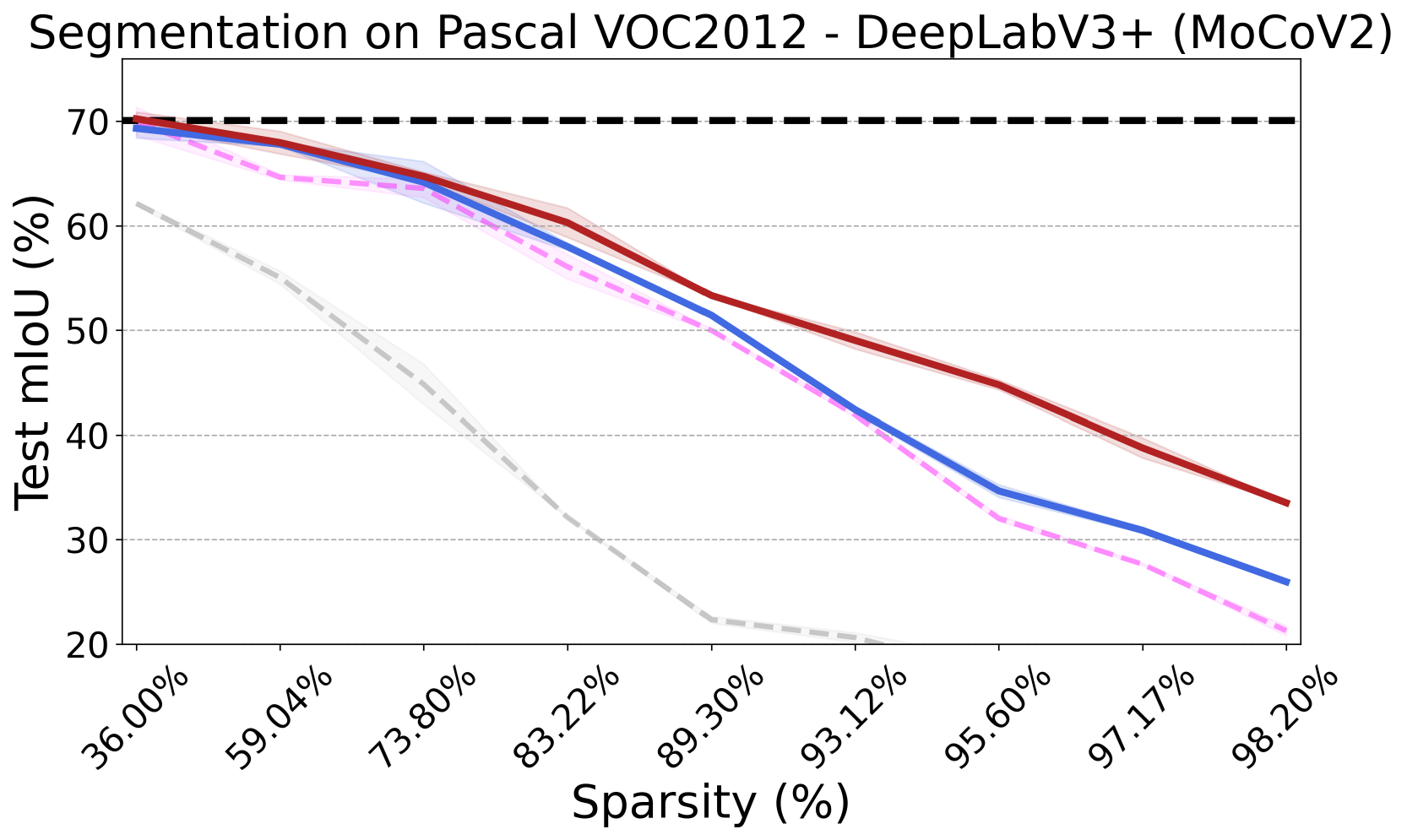}
\includegraphics[width=0.33\linewidth]{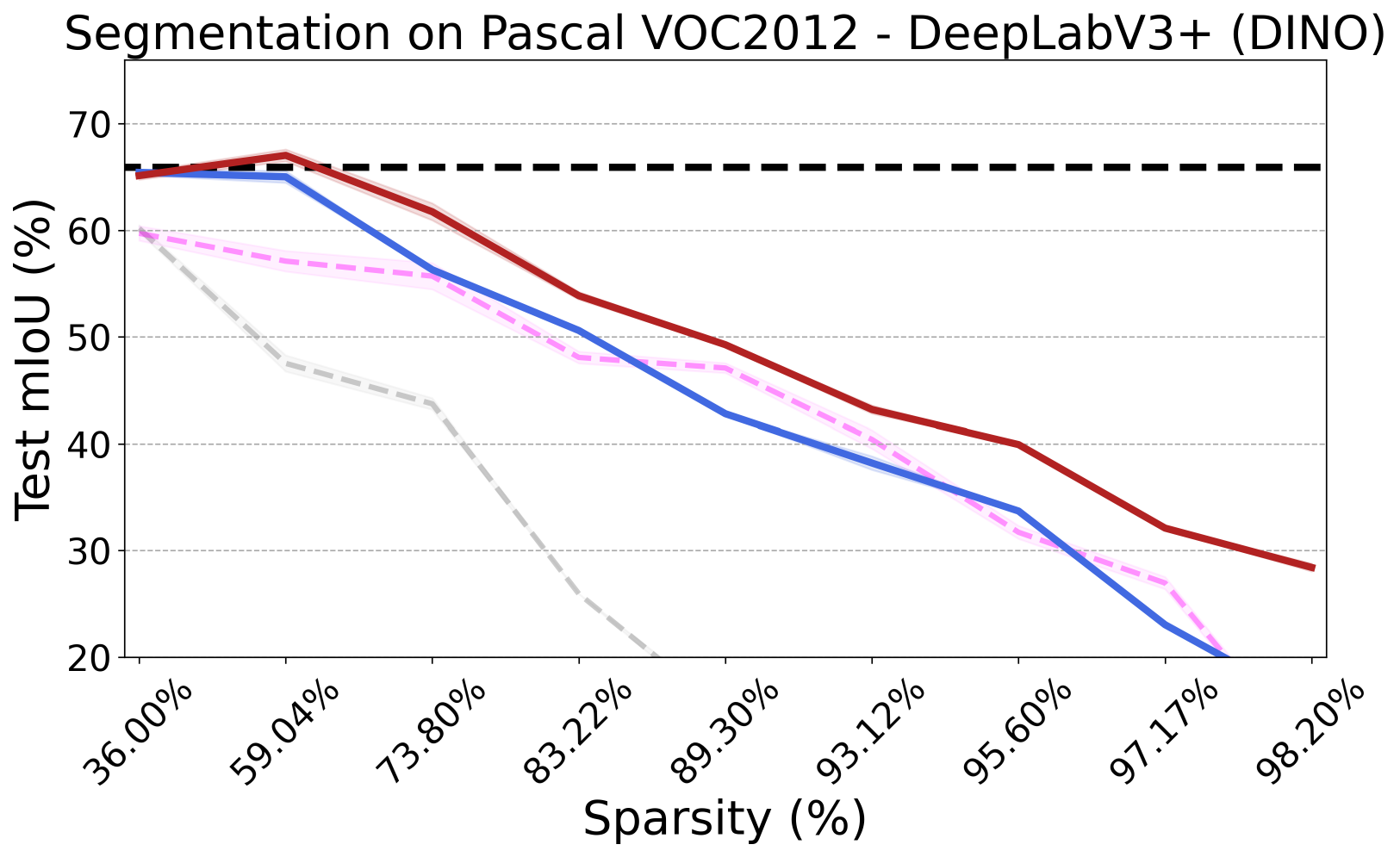}

\vspace{1mm}
\includegraphics[width=0.57\linewidth]{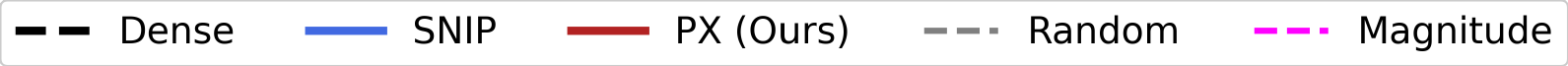}

\caption{Average mean Intersection over Union (mIoU) at different sparsity levels on Pascal VOC2012 using DeepLabV3+ with pre-trained ResNet-50 as the backbone. Each experiment is repeated three times.
Standard deviations are in shaded colors.
}
\label{fig:add_segmentation_exp} \vspace{-4mm}
\end{figure*}

\begin{algorithm}[tb]
\scriptsize
\hrule
\smallskip
\KwData{Network $\bm{f}$ parametrized by $\bm{\theta}$, pruning dataset $\mathcal{D}$ made of $B$ mini-batches, number of pruning rounds $T$, final sparsity level $k$}
\KwResult{Parameter mask $\bm{M}$ used to sparsify $\bm{f}$ before training}
\smallskip

\ccom{\ul{.detach()}: the operation is detached from the computational graph.}\\
\ccom{$\bm{z}$ is the output of a network, $\bm{a}$ is the vector of activations.}\\
\smallskip
  
$\bm{g}, \bm{h}$ = $\bm{f}$ \ccom{create two copies of $\bm{f}$}\\
$\bm{M} = \mathbbm{1}$ \ccom{init. parameter mask to all 1s}\\

\For{$t$ \cres{in} $1, 2, ..., T$}{
    \ccom{perform the $t$-th pruning round}\\
    $p$ = $k^{(t / T)}$ \ccom{compute the \% of weights to remove at round $t$}\\
    $\bm{s}$ = $\mathbb{0}$ \ccom{init. saliency scores to all 0s}\\

    \For{$i$ \cres{in} $1,2,...,B$}{
        $\bm{x}$ = $\mathcal{D}_i$ \ccom{mini-batch of data at index $i$}\\
        
        \_, $\bm{a}$ = $\bm{f}(\bm{x}, \bm{\theta} \odot \bm{M}, \_)$.detach() \ccom{record activations $\bm{a}$}\\

        $\bm{z}_{\bm{g}}$, \_ = $\bm{g}(\bm{x}^2, \mathbbm{1} \odot \bm{M}, \bm{a})$.detach() \ccom{force activations to $\bm{a}$}\\

        $\bm{z}_{\bm{h}}$, \_ = $\bm{h}(\mathbbm{1}, \bm{\theta}^2 \odot \bm{M}, \mathbbm{1})$ \ccom{forward an input of all 1s}\\

        $R$ = ($\bm{z}_{\bm{g}} \odot \bm{z}_{\bm{h}}$).sum() \ccom{compute score function}\\
        $\bm{s}$ += $R\text{.backward()} \odot \bm{\theta}^2$ \ccom{update param.-wise saliency scores}\\ 
    }
    \ccom{update $\bm{M}$ to keep only top-$p$ parameters} \\
    $\bm{\tilde{s}}$ = sort\_descending($\bm{s}$) \ccom{sorted saliency scores in descending order}\\

    $p$ = $\text{length(}\bm{\theta}\text{)} \cdot p$ \ccom{compute top-$p$ threshold index}\\

    \For{$j$ \cres{in} $1,2,...,\text{length(}\bm{\theta}\text{)}$}{
        \If{$\bm{s}_j - \bm{\tilde{s}}_p < 0$}{
            $\bm{M}_j = 0$ \ccom{set the mask of the $j$-th param. to zero}\\
        }
    }
}
\Return{$\bm{M}$}
\hrule\vspace{1mm}
\caption{\small Pruning via Path eXclusion (PX)}
\label{algo:pseudocode}
\end{algorithm}

\section{Additional Discussions}
In this section we provide the derivations and intuitions about the mathematics used in the main submission, to make it clear how the path-wise perspective can be studied via forward and backward passes on any architecture. Note that in all of our derivations and formulas, we skip bias terms as we embed them in the weight matrix by adding an additional input set to 1 to each neuron.

\smallskip
\noindent\textbf{Frobenius norm of the Path Activation matrix.} In Eq. (7) of the main submission we applied the definition of the Frobenius norm on the Path Activation matrix
\begin{equation*}
\begin{split}
    \|\bm{J}_{\bm{v}}^{\bm{f}}(\bm{X})\|_F^2 &= \sum_{n=1}^N \sum_{k=1}^{K} \sum_{p=1}^{P} \left(\sum_{s=1}^d \mathbb{I}[p \in \mathcal{P}_{s \rightarrow k}] a_p(\bm{x}_n, \bm{\theta}) \bm{x}_{n_s}\right)^2 \\
\end{split}
\end{equation*}
We observe that by fixing a path $p$, the inner sum from $s=1$ to $d$ will have only one non-zero term (given by the indicator function). Specifically, the one for which $s$ is the starter input node: the other input nodes cannot take part in path $p$ as the first connection will define a different path. This means that the equality reported in Eq. (7) of the main submission holds, provided that we take the correct input node $s | s \in p$.

\begin{figure*}[tb]
\centering
\includegraphics[width=0.33\linewidth]{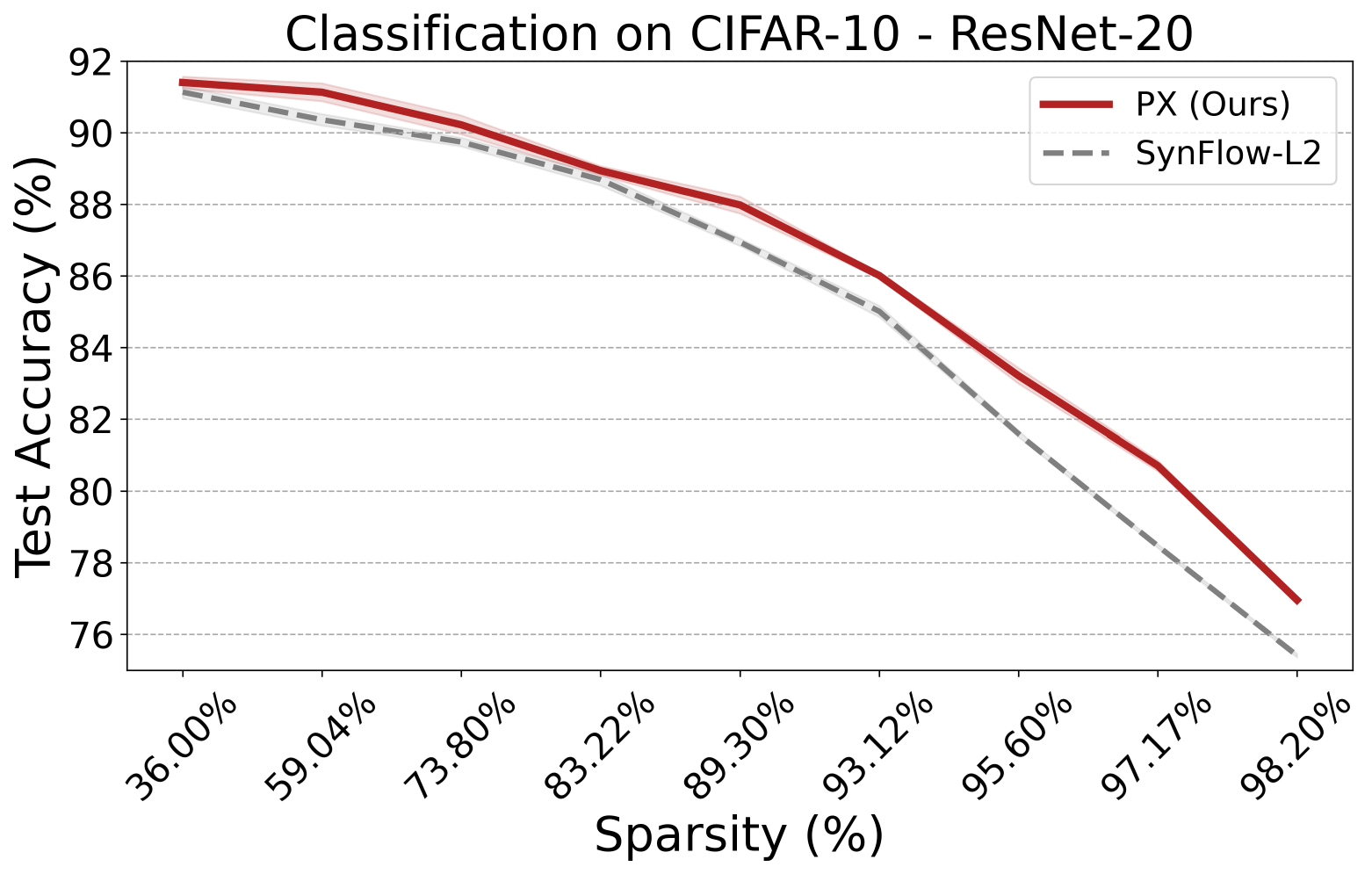}
\includegraphics[width=0.33\linewidth]{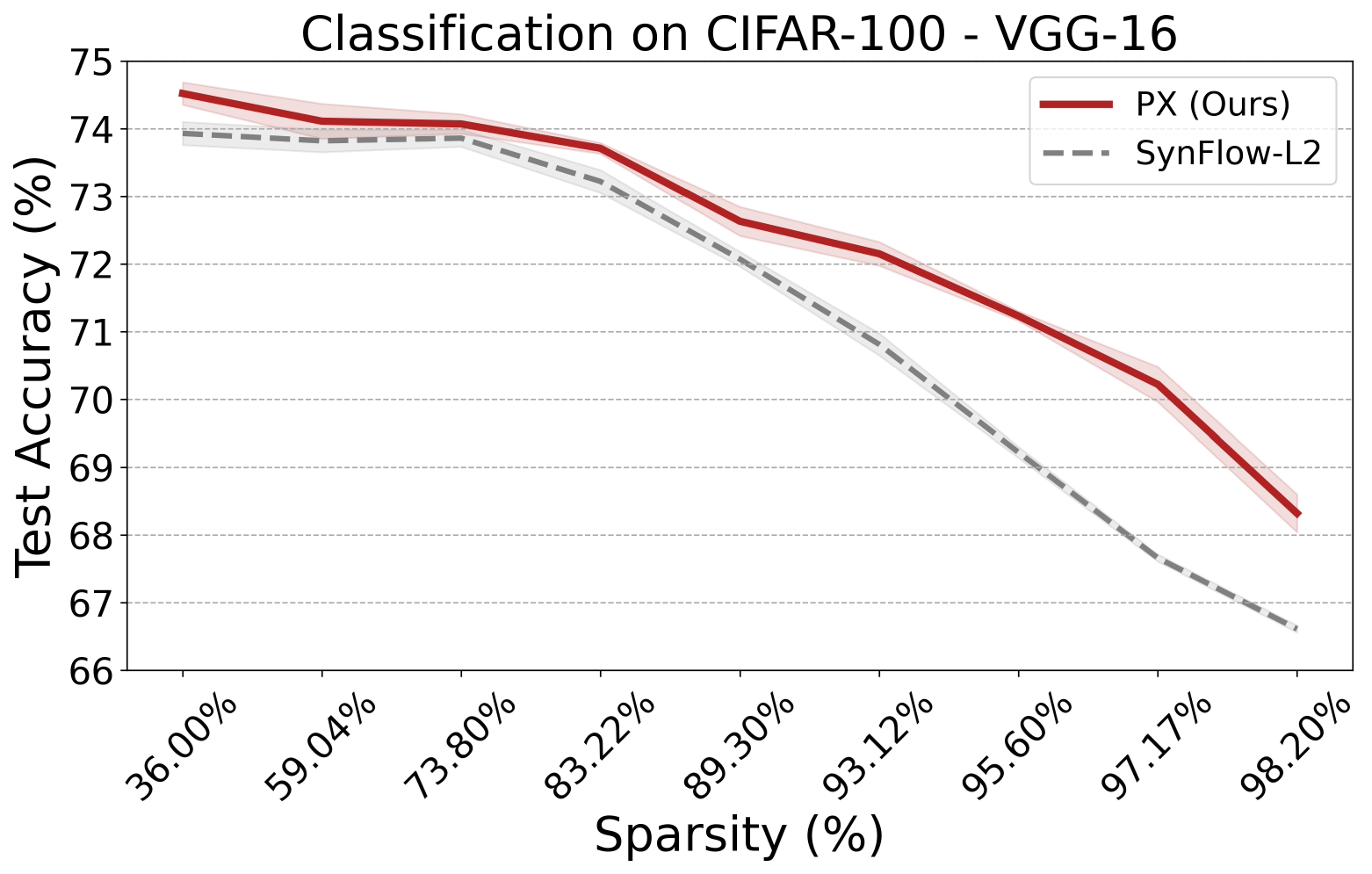}
\includegraphics[width=0.33\linewidth]{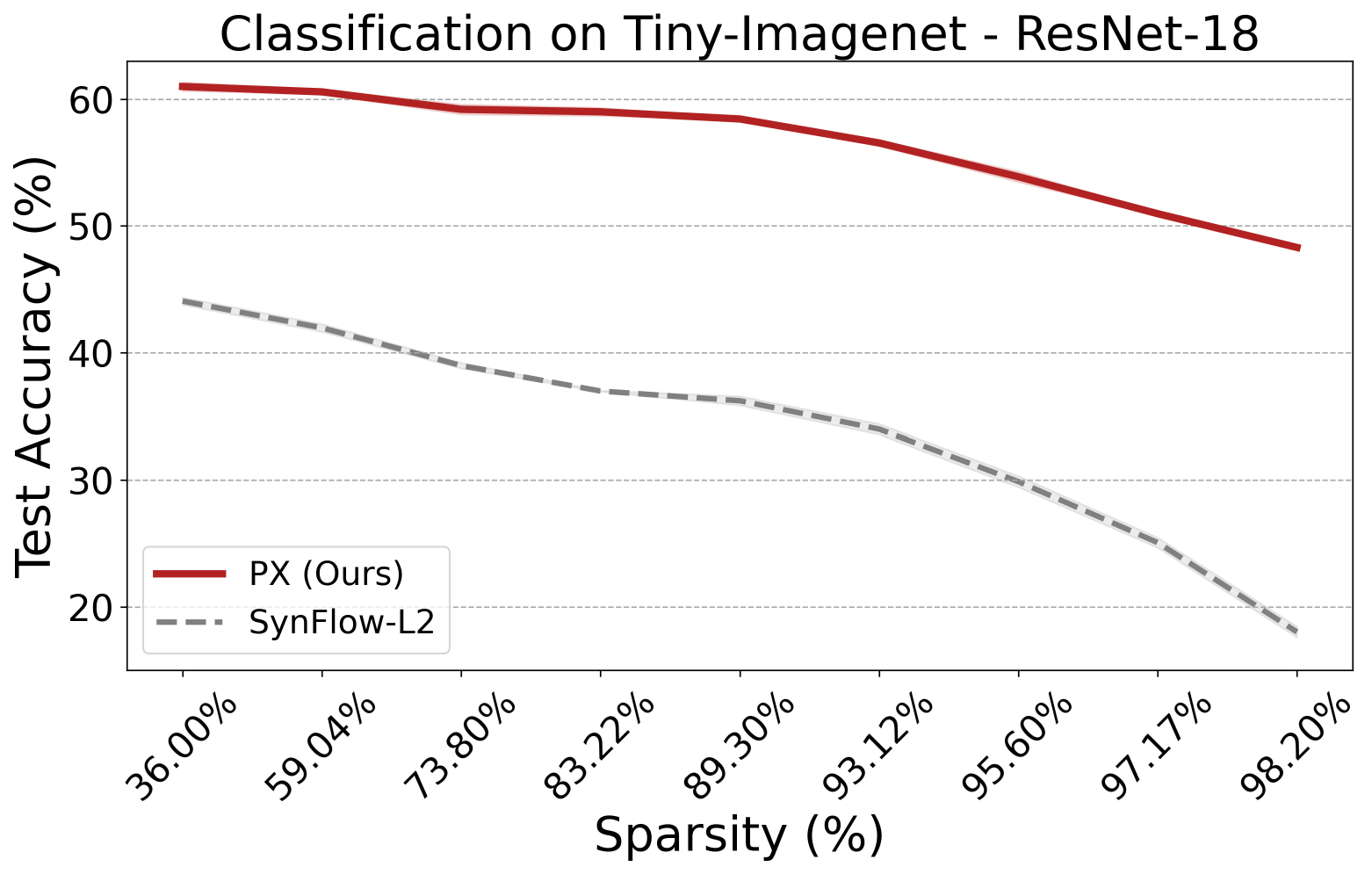}
\vspace{-1mm}
\caption{Average classification accuracy at different sparsity levels on CIFAR-10 using ResNet-20, CIFAR-100 using VGG-16 and Tiny-ImageNet using ResNet-18, respectively. Each experiment is repeated three times. We report in shaded colors the standard deviation.}
\label{fig:ablation_random} \vspace{-2mm}
\end{figure*}

\begin{figure*}[tb]
\centering
\includegraphics[width=0.33\linewidth]{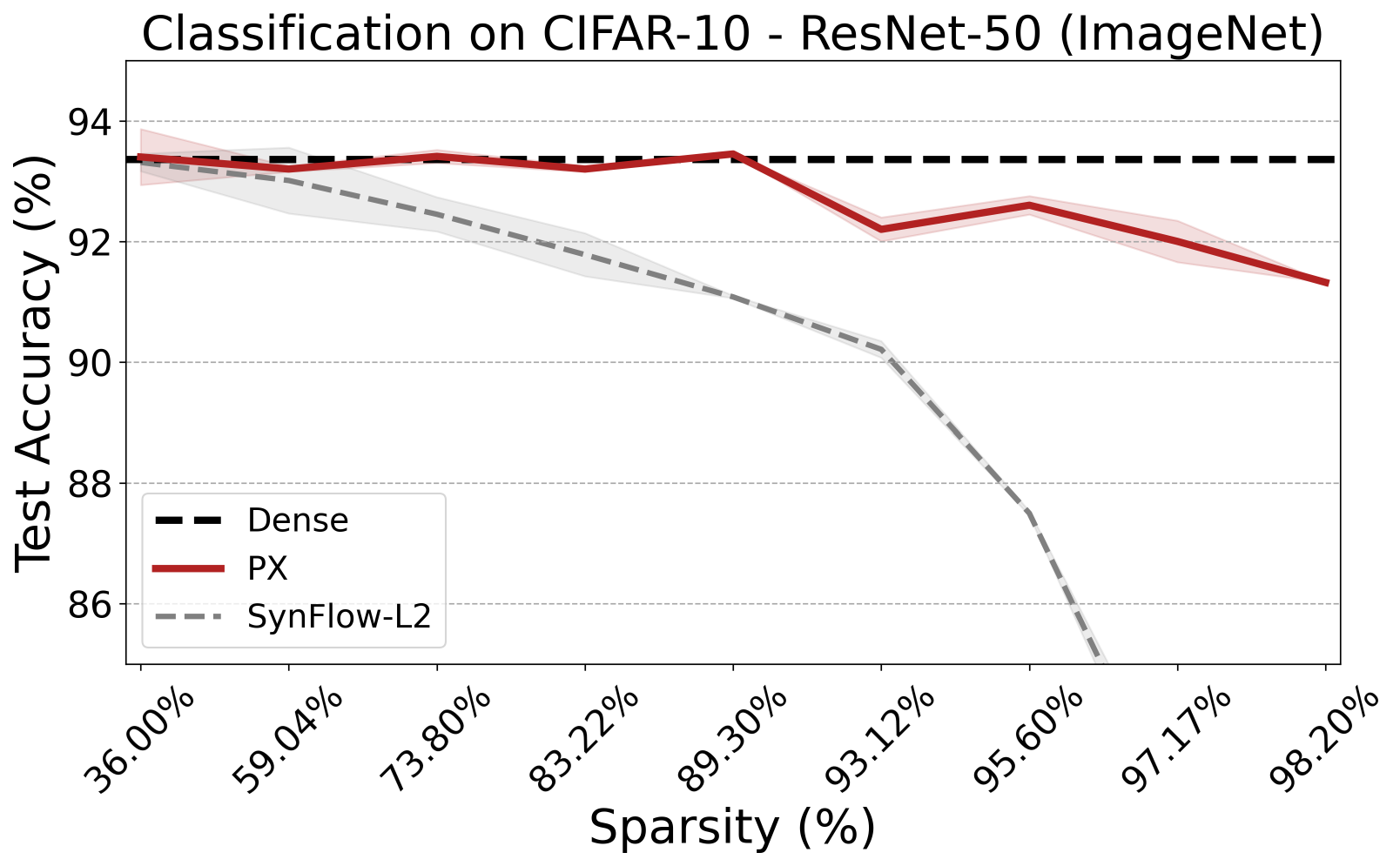}
\includegraphics[width=0.33\linewidth]{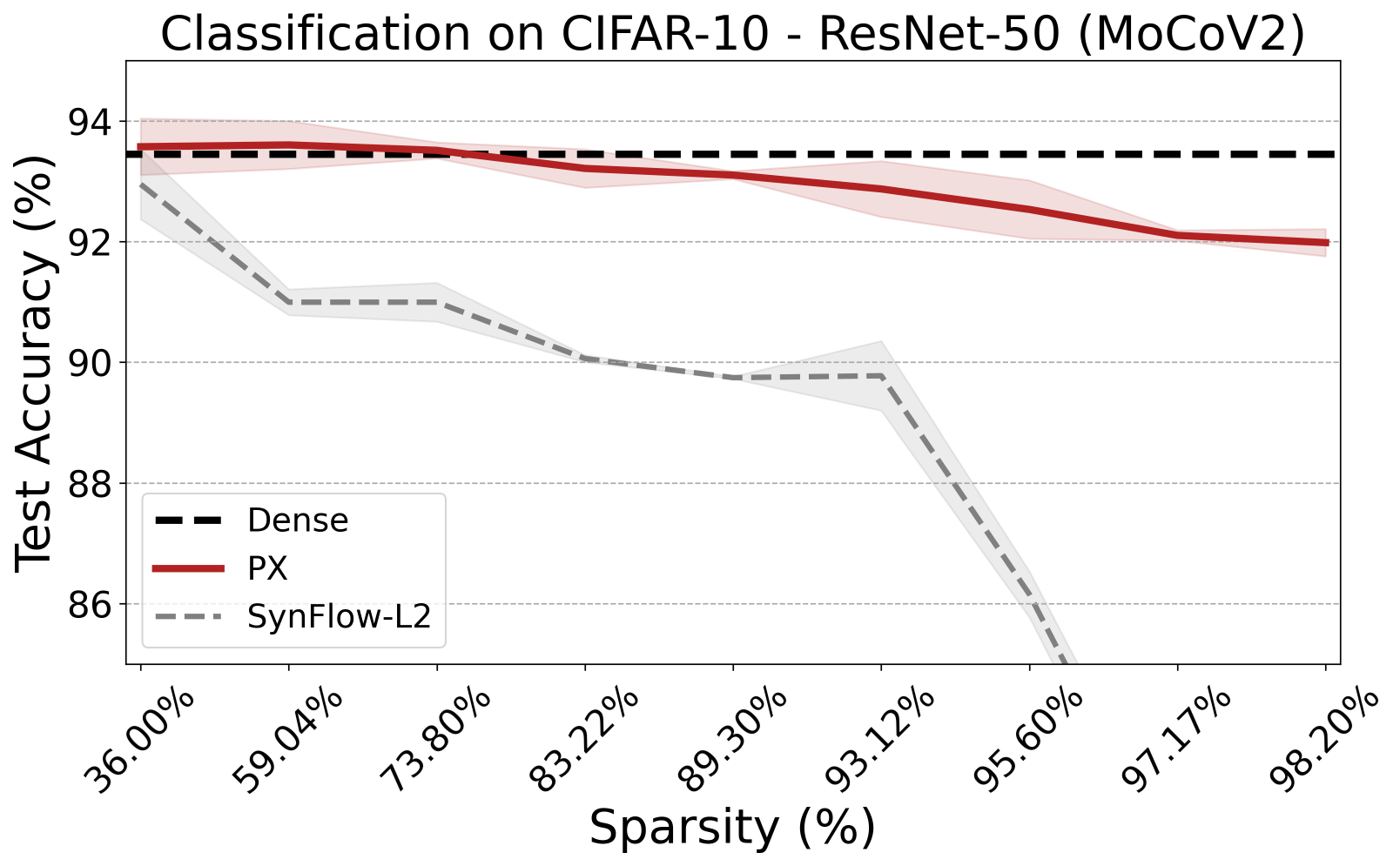}
\includegraphics[width=0.33\linewidth]{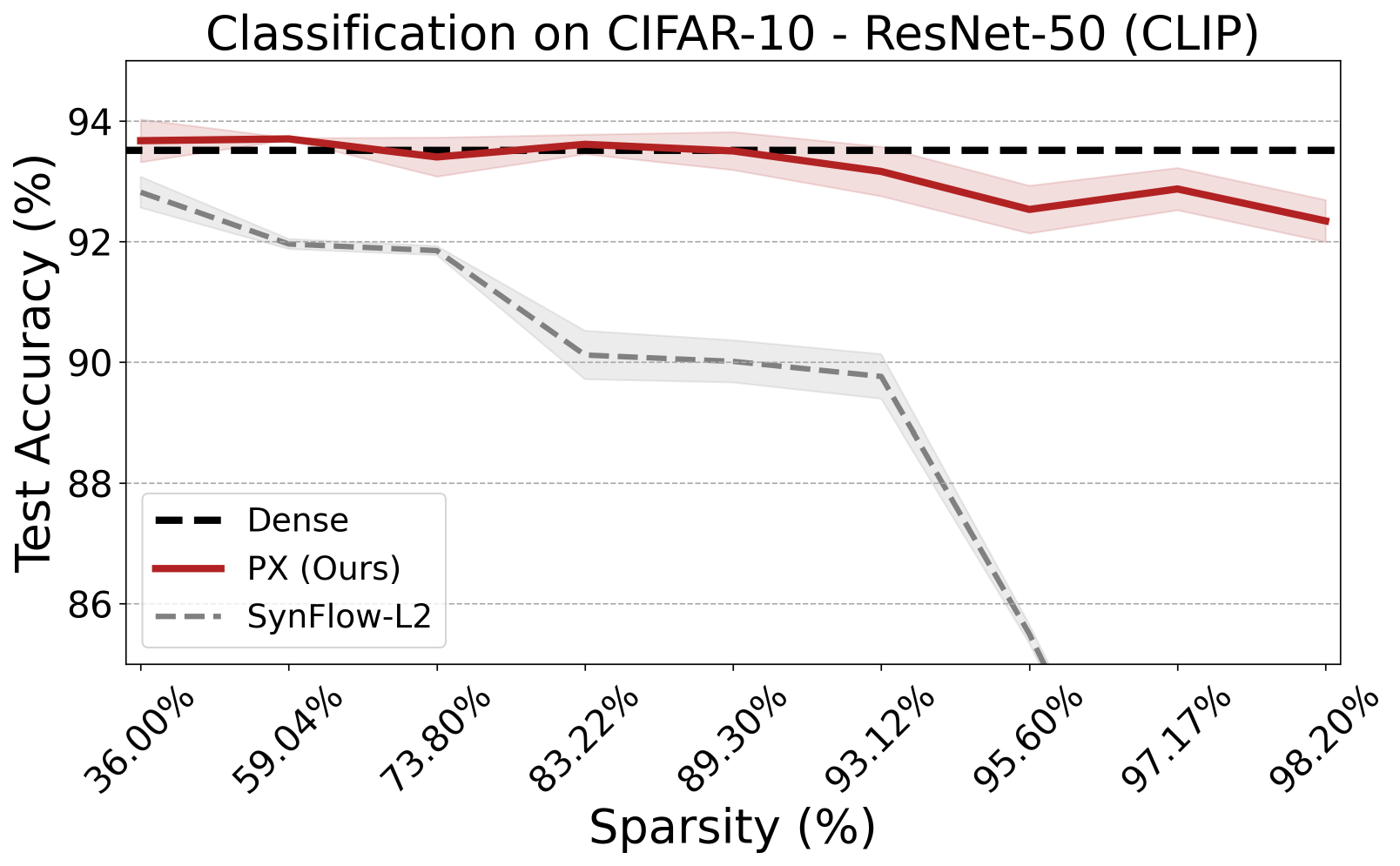}

\includegraphics[width=0.33\linewidth]{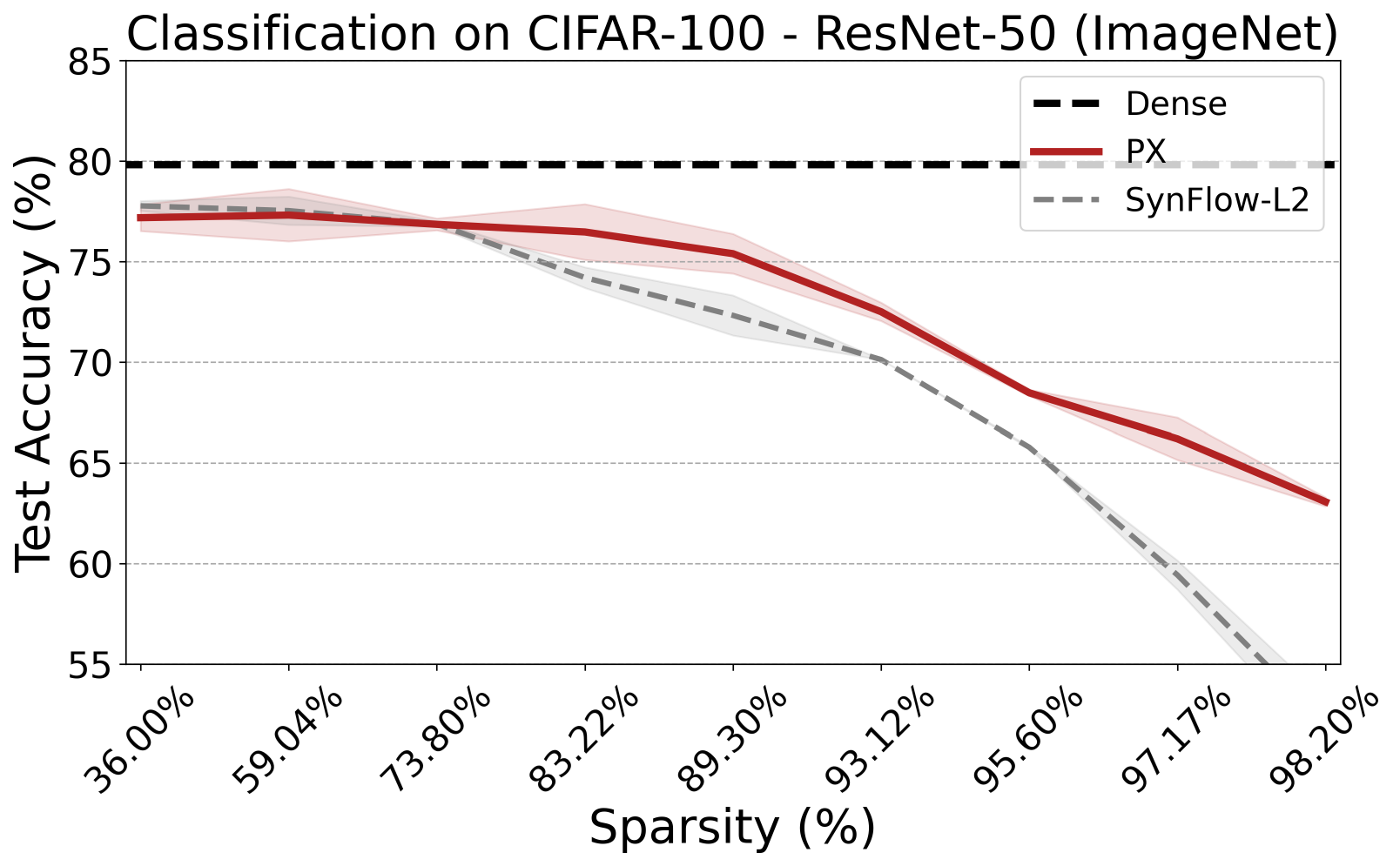}
\includegraphics[width=0.33\linewidth]{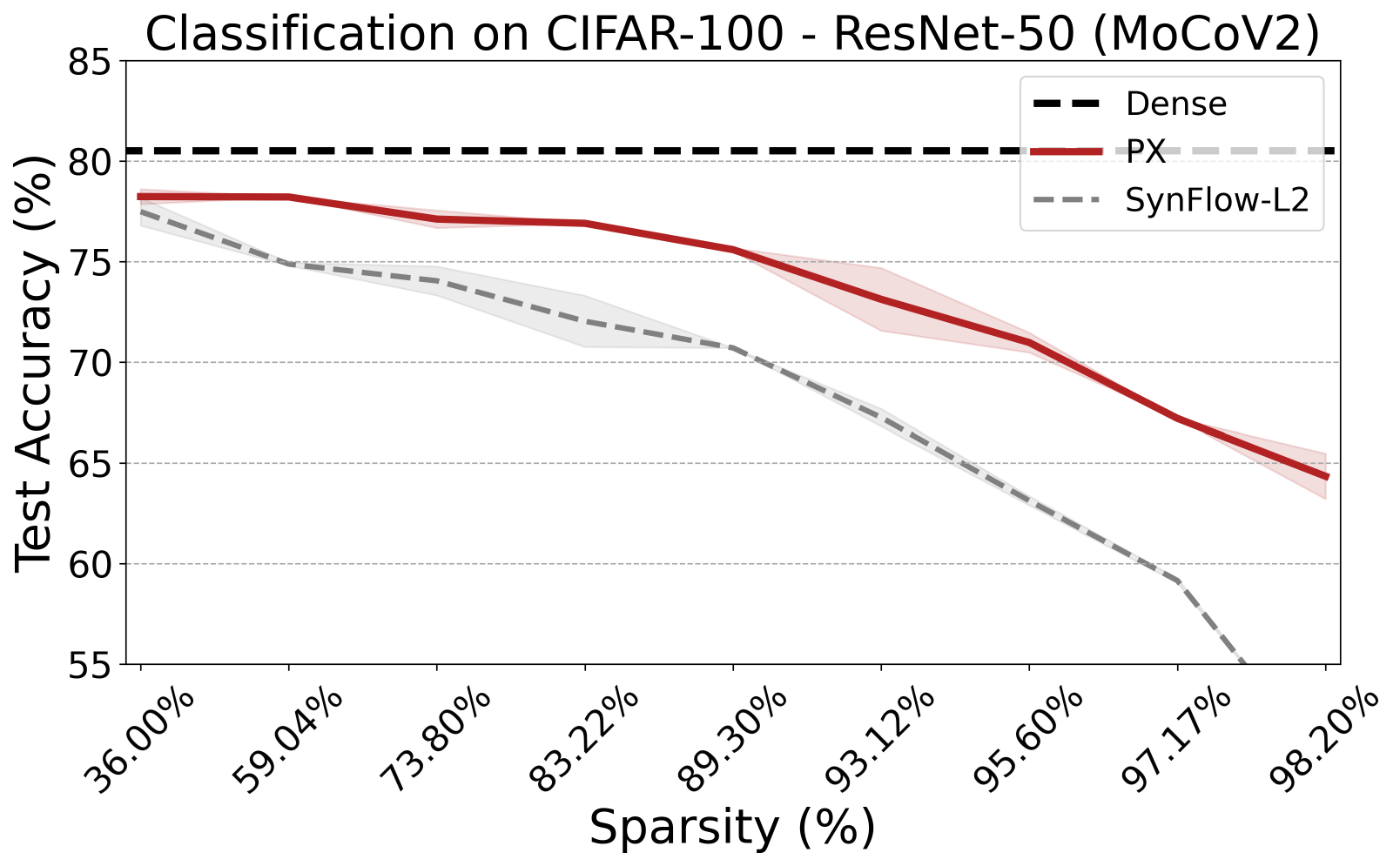}
\includegraphics[width=0.33\linewidth]{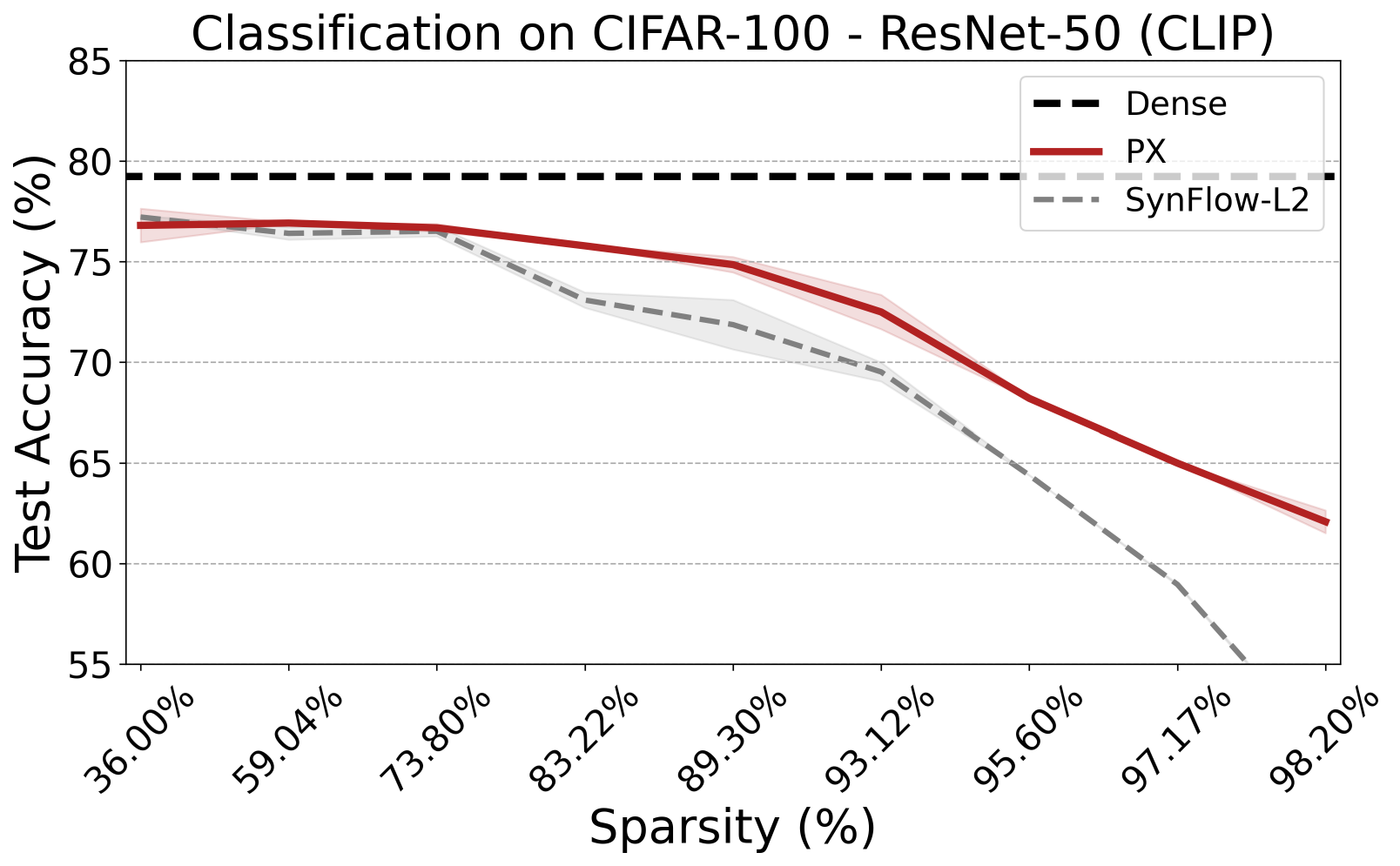}

\includegraphics[width=0.33\linewidth]{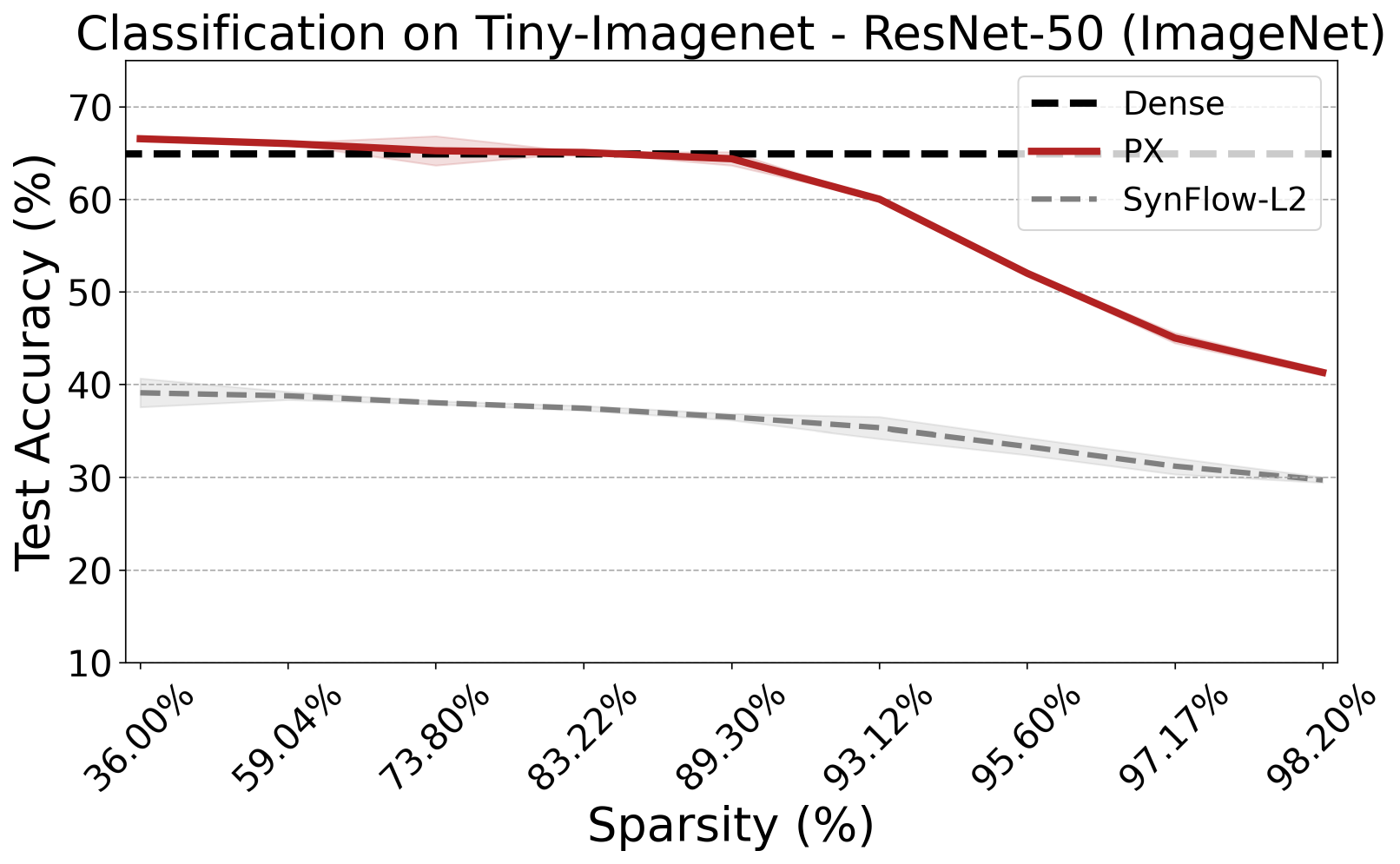}
\includegraphics[width=0.33\linewidth]{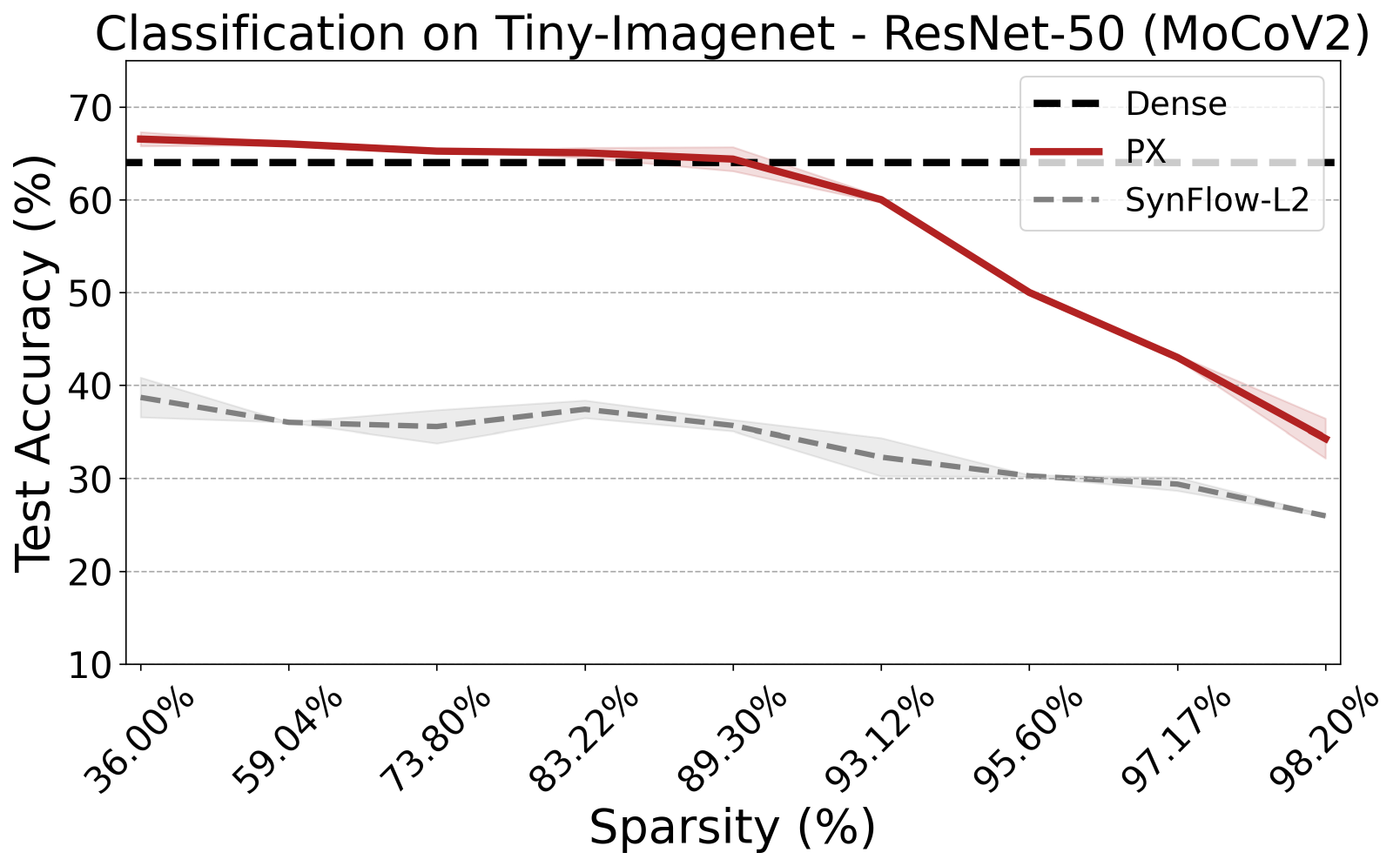}
\includegraphics[width=0.33\linewidth]{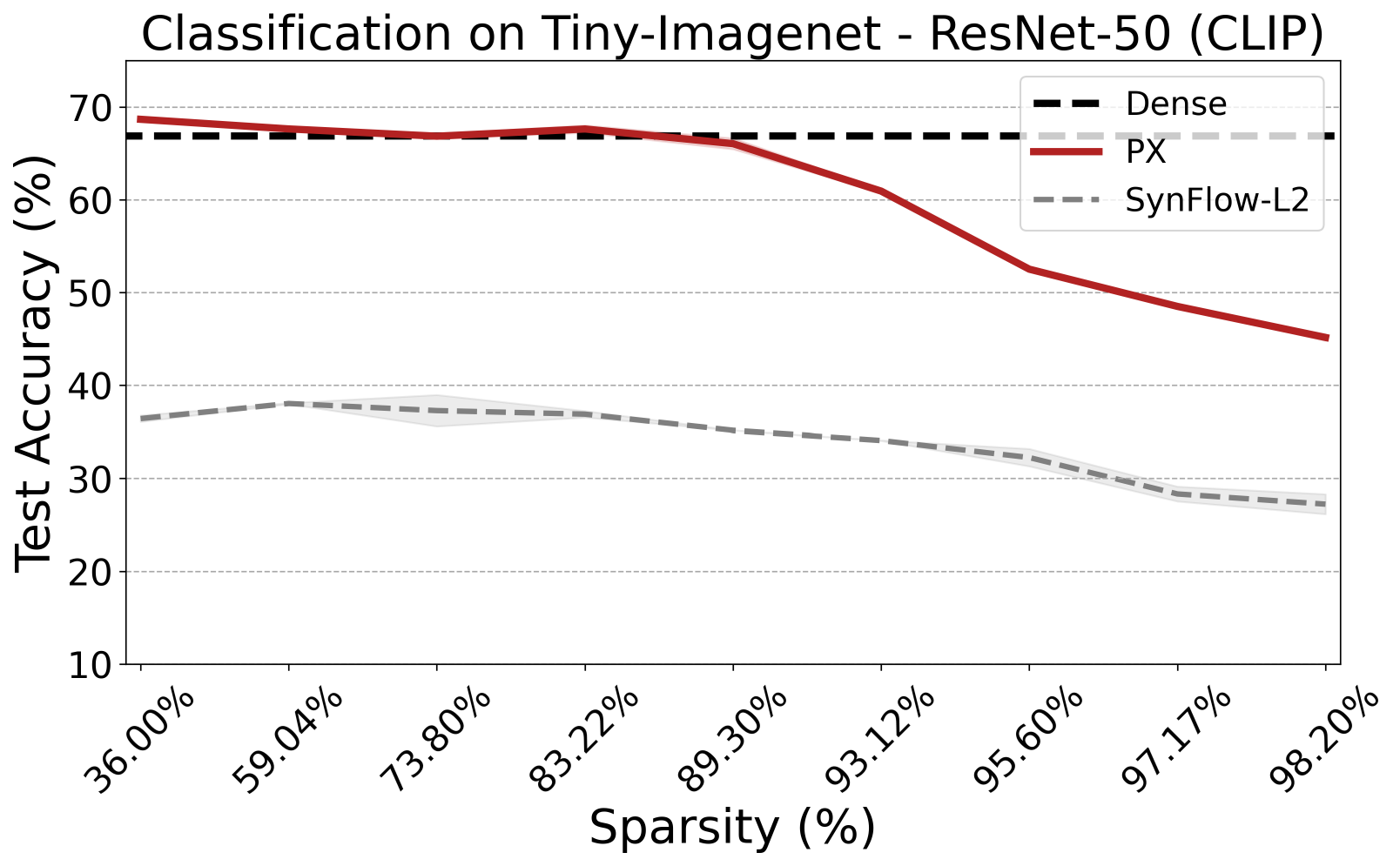}
\vspace{-3mm}
\caption{Average classification accuracy at different sparsity levels on CIFAR-10, CIFAR-100 and Tiny-ImageNet using pre-trained ResNet-50 as architecture. The first column reports the results of starting from the supervised ImageNet pre-training. The second column reports the performance when starting from the MoCov2 pre-training on ImageNet. Finally, in the third column we report the results when starting from CLIP. Each experiment is repeated three times. We report in shaded colors the standard deviation.}
\label{fig:ablation_pretrain} \vspace{-2mm}
\end{figure*}

\begin{figure*}[tb]
\centering
\includegraphics[width=0.33\linewidth]{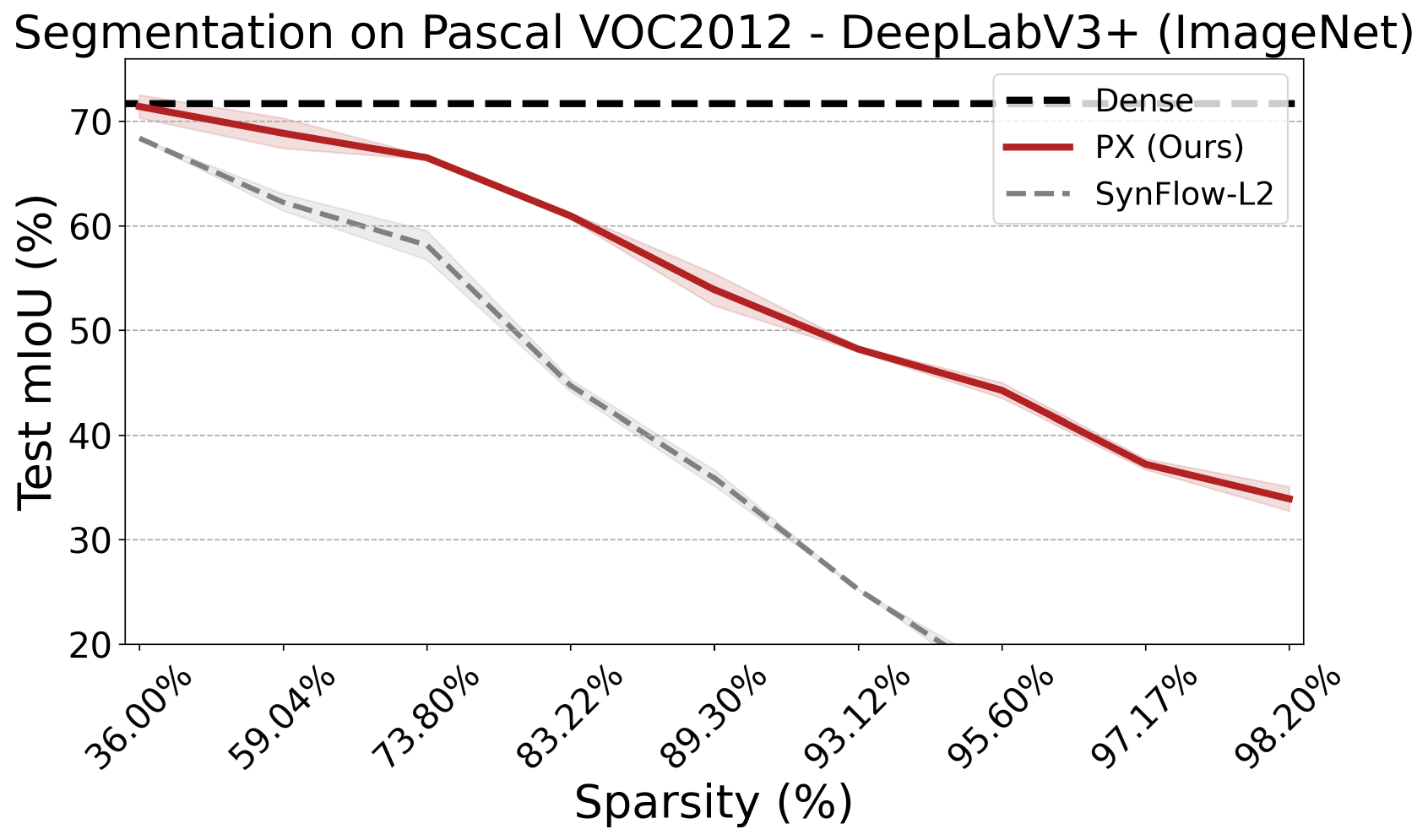}
\includegraphics[width=0.33\linewidth]{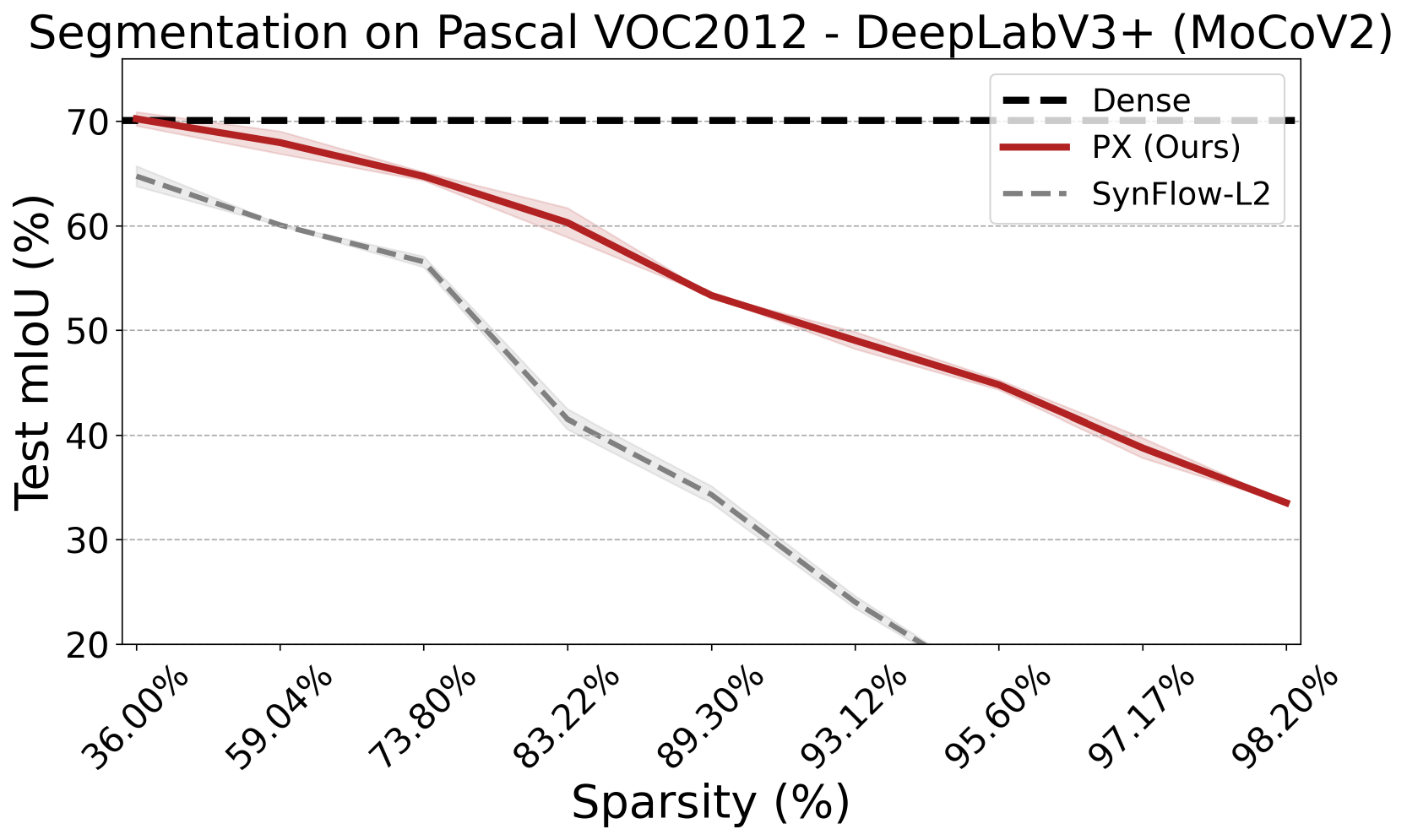}
\includegraphics[width=0.33\linewidth]{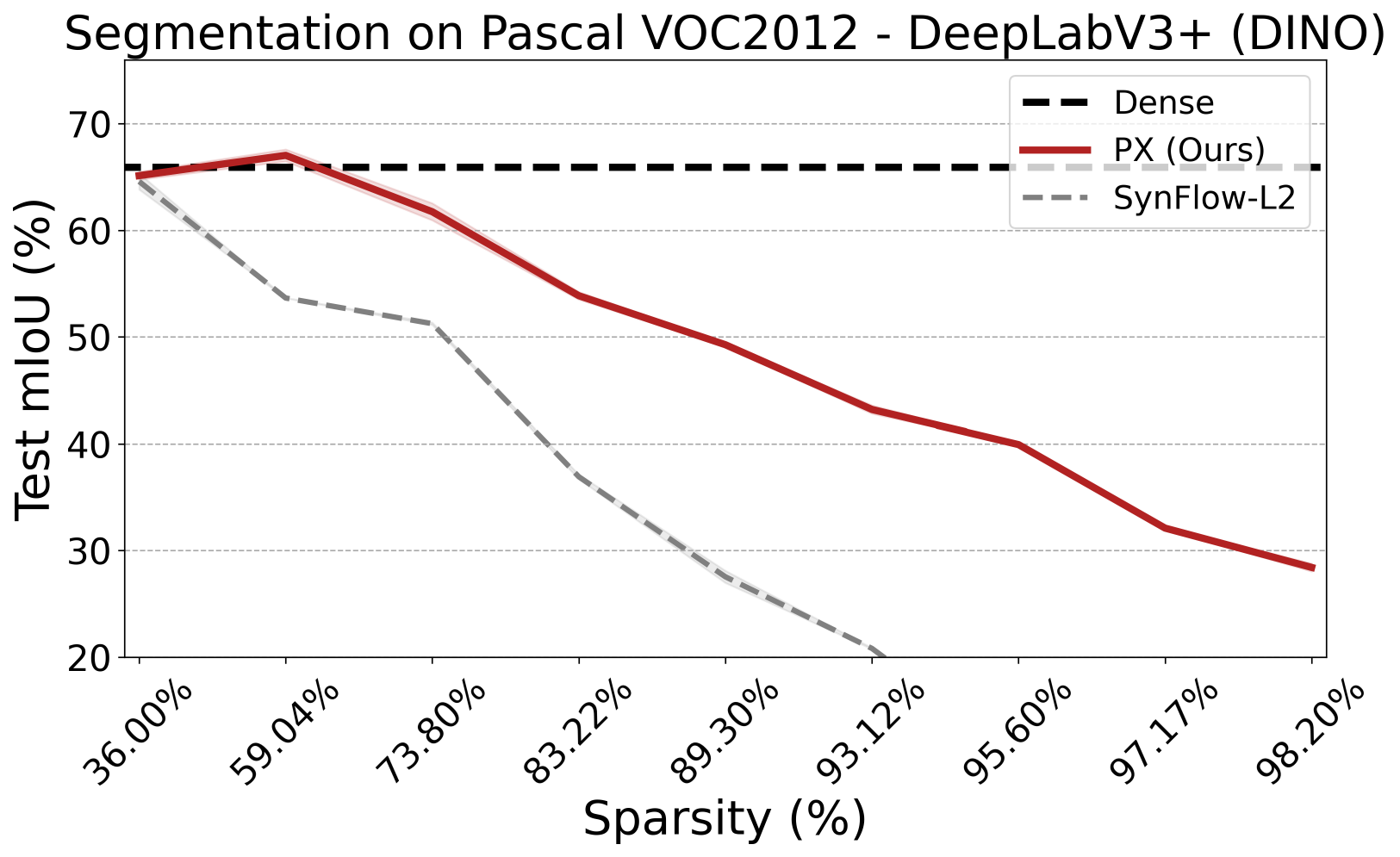}

\caption{Average mean Intersection over Union (mIoU) at different sparsity levels on Pascal VOC2012 using DeepLabV3+ with pre-trained ResNet-50 as the backbone. Each experiment is repeated three times.
Standard deviations are in shaded colors.
}
\label{fig:ablation_segm} \vspace{-2mm}
\end{figure*}

\begin{figure*}[tb]
\centering
\includegraphics[width=0.33\linewidth]{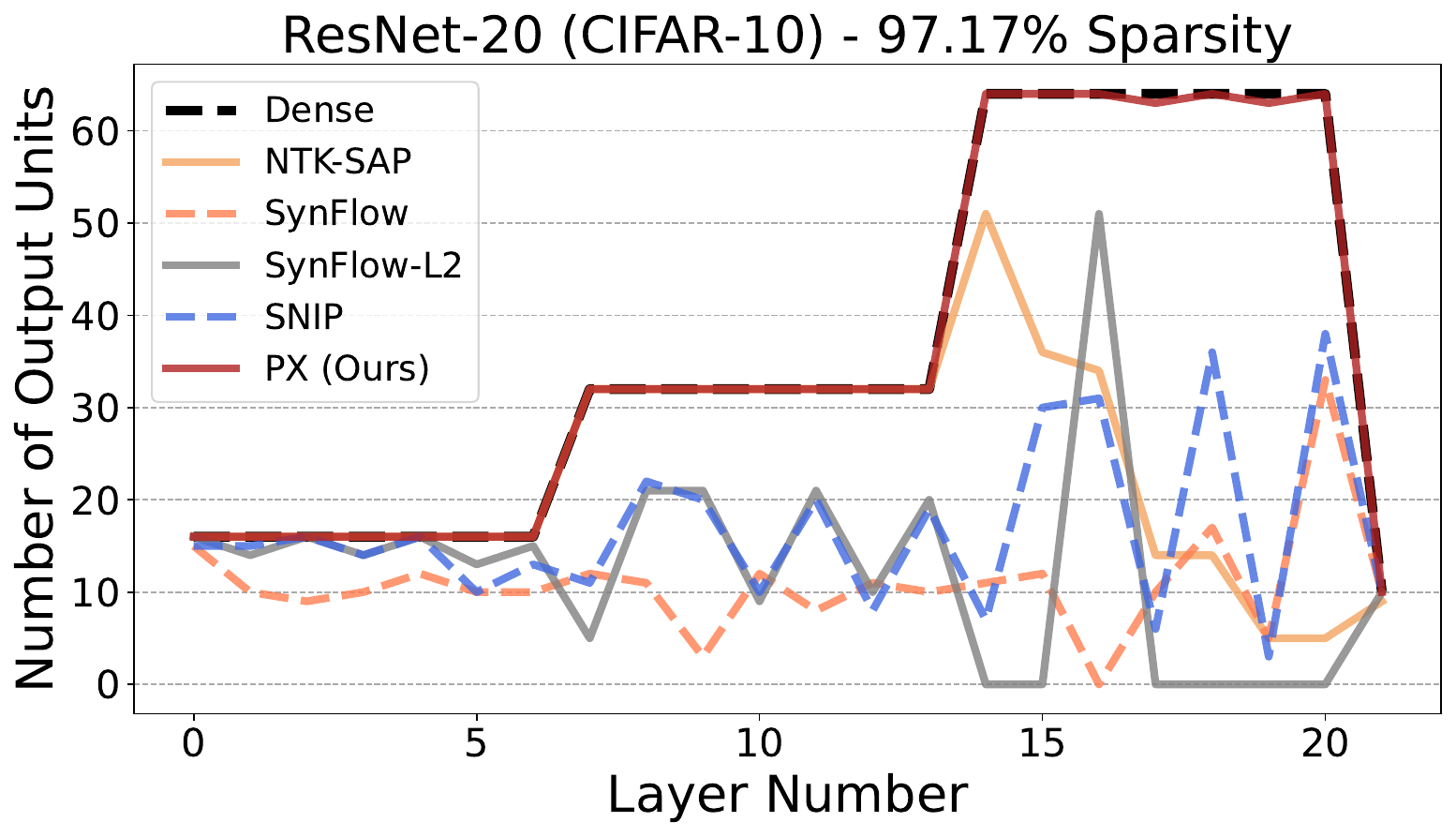}
\includegraphics[width=0.33\linewidth]{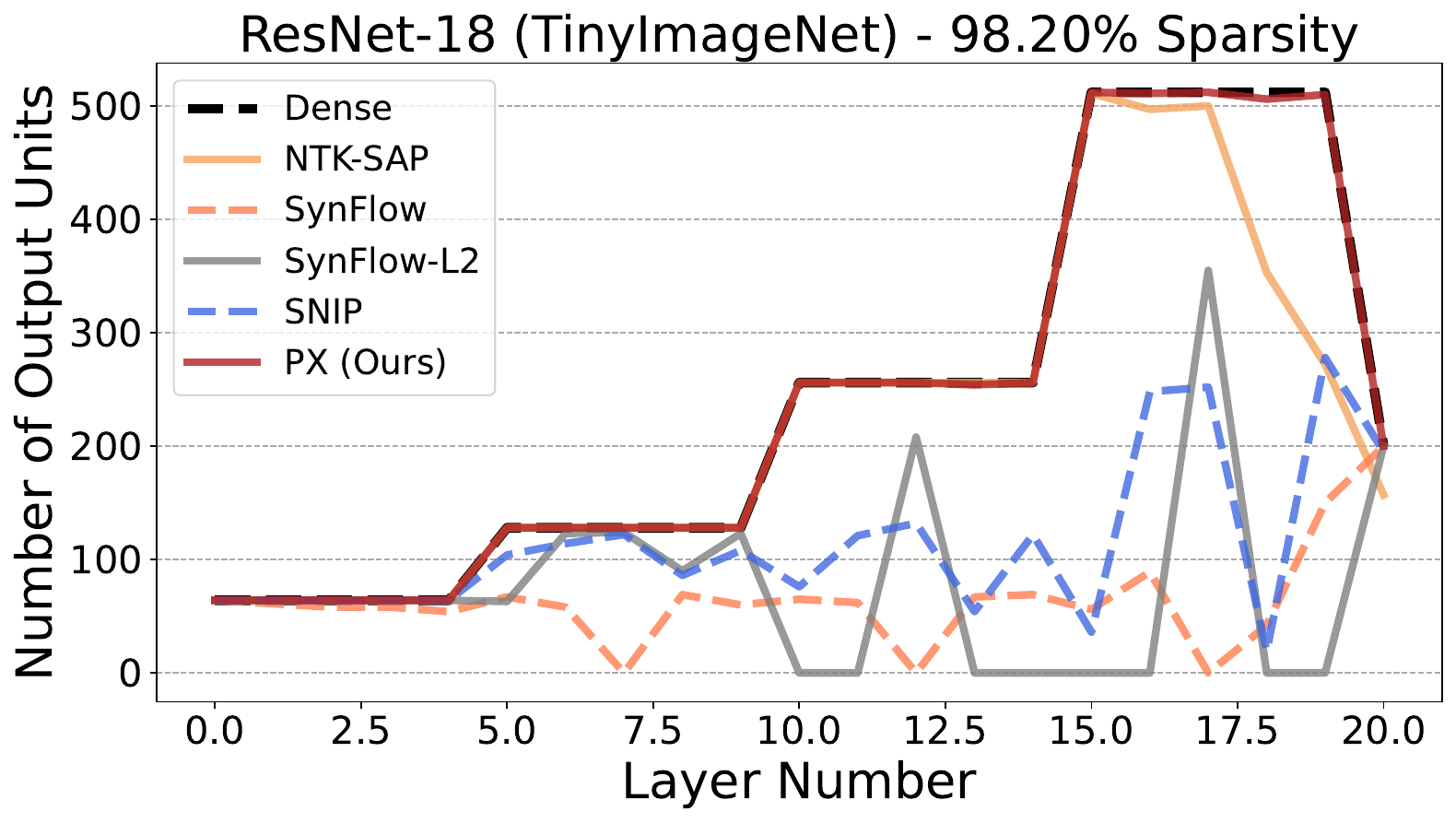}
\includegraphics[width=0.33\linewidth]{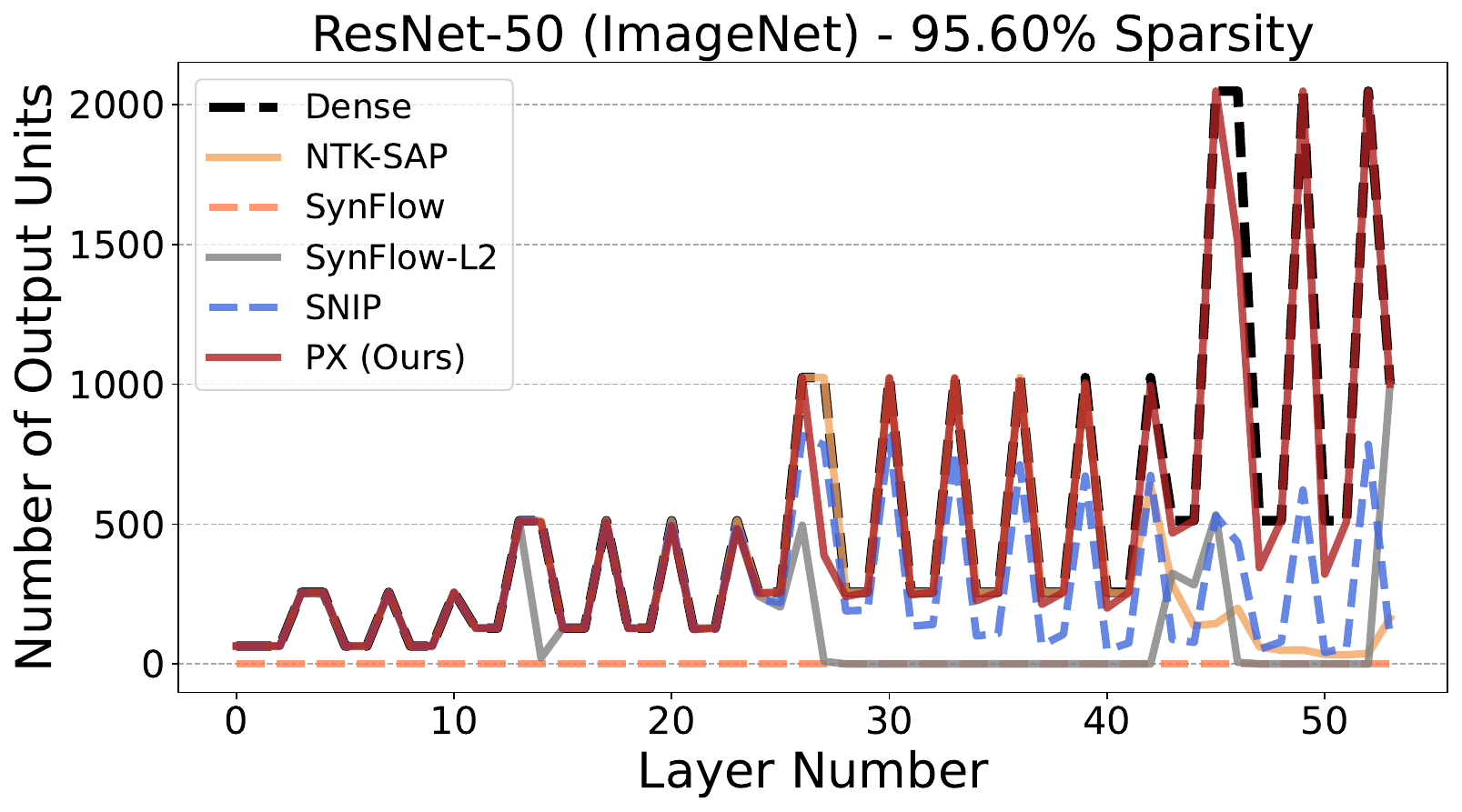}

\caption{Active output units at different sparsities in ResNet-20,  ResNet-18 and ResNet-50. For SNIP and PX data mini-batches are sampled from CIFAR-10, Tiny-ImageNet and ImageNet, respectively.
}
\label{fig:add_widths} \vspace{-3mm}
\end{figure*}

\begin{figure*}[th!]
\centering
\includegraphics[width=0.33\linewidth]{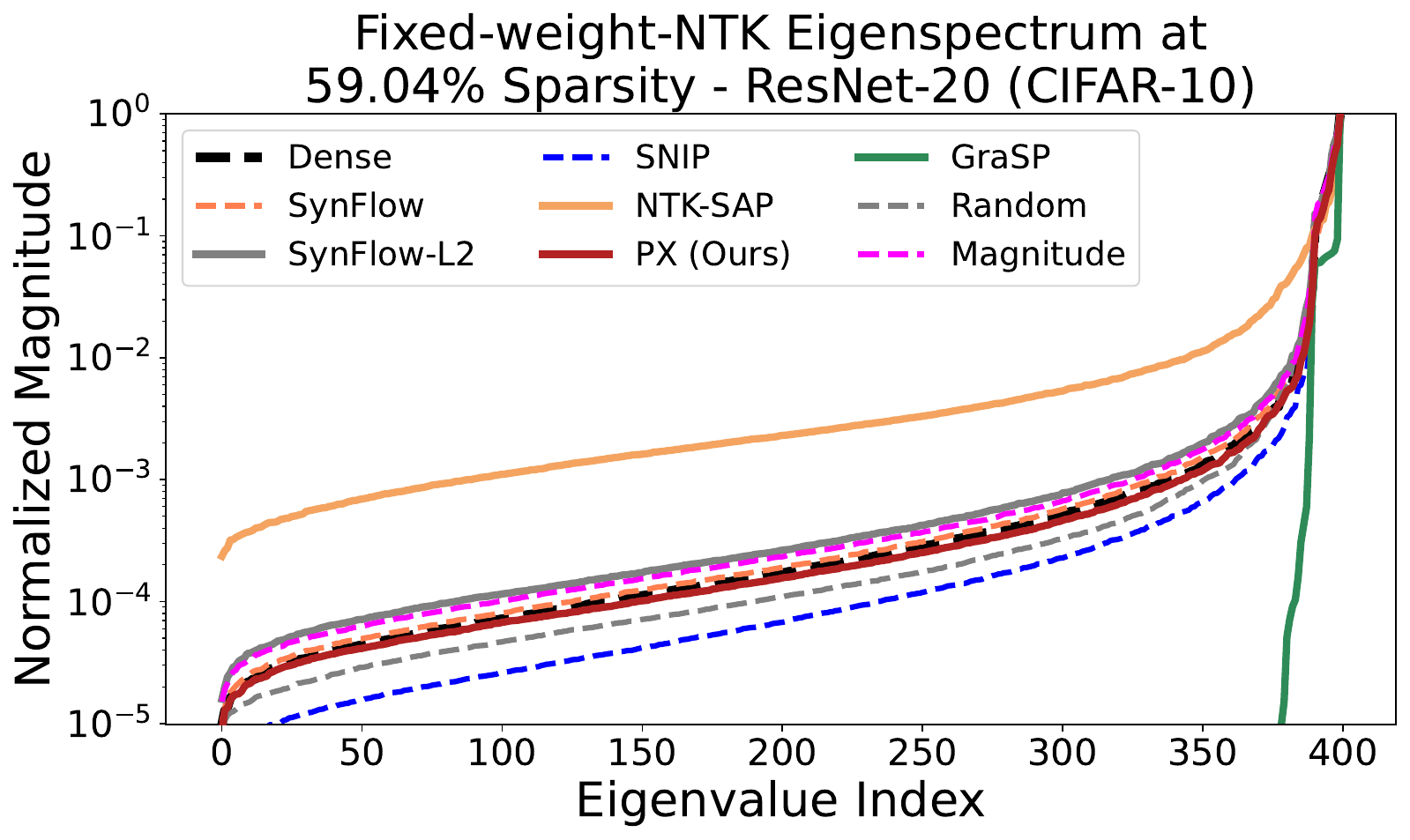}
\includegraphics[width=0.33\linewidth]{Spectrum_93.12+L2.pdf}
\includegraphics[width=0.33\linewidth]{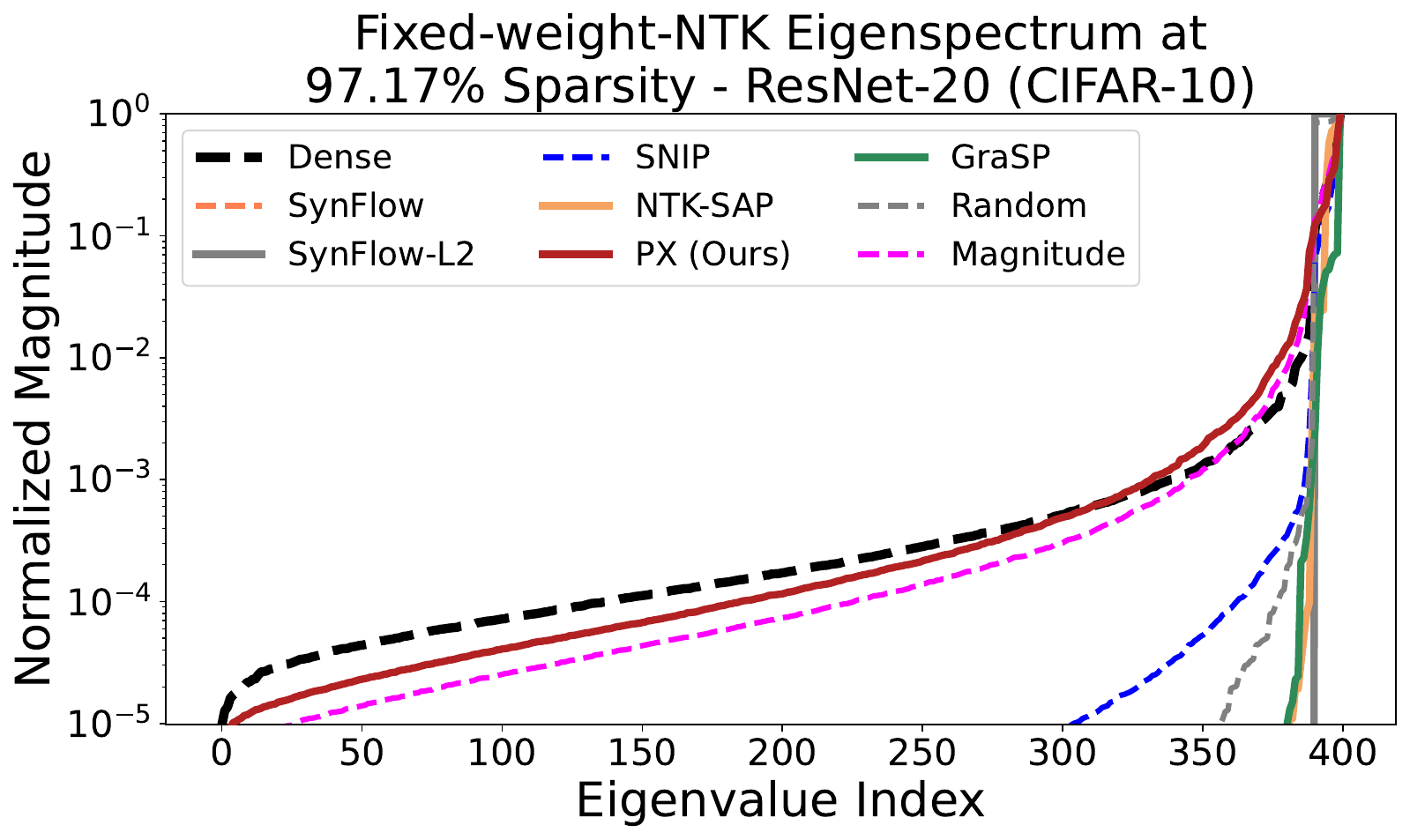}

\caption{Fixed-Weight-NTK spectrum of ResNet-20 on the CIFAR-10 dataset at different sparsity ratios.
}
\label{fig:add_spectrum} \vspace{-3mm}
\end{figure*}

\begin{table}[t!]
    \begin{center}
    \small
    \setcellgapes{1.5pt}
    \makegapedcells
    \resizebox{0.44\textwidth}{!}{
    \begin{tabular}{c c c c}
        \bottomrule      
        Pruning  & Computational & Epochs to Reach & Accuracy at  \\
        Method & Complexity & 98.20\% Sparsity & 98.20\% Sparsity \\        
        \hline\hline
        
        IMP \cite{frankle2018lottery} & $\mathcal{O}(1)$ & 960  & 77.38  \\ \hline
        
        \multicolumn{4}{c}{Single-shot PaI methods} \\ \hline
        
        Random & $\mathcal{O}(1)$ & 0  & 72.31 \\
        Magnitude & $\mathcal{O}(1)$ & 0  & 76.12 \\
        SNIP \cite{lee2018snip} & $B \cdot ([FP] + [BP])$ & 0  & 75.39 \\
        GraSP \cite{grasp2020wang} & $B \cdot (2[FP] + 2[BP])$ & 0  & 76.30 \\ \hline

        \multicolumn{4}{c}{Iterative PaI methods} \\ \hline
        
        SynFlow \cite{tanaka2020synflow} & $T \cdot ([FP] + [BP])$ & 0  & 75.19 \\
        NTK-SAP \cite{ntksap2023wang} & $T \cdot B \cdot (3[FP] + [BP])$ & 0  & 74.55 \\
        
        \textbf{PX (Ours)} & $T \cdot B \cdot (3[FP] + [BP])$ & 0  & 77.08 \\
        
        \bottomrule
    
    \end{tabular}}
    \end{center}
    \vspace{-5mm}
    \caption{Comparison of the computational complexity of each pruning procedure. Last column is the accuracy at 98.20\% sparsity when starting from ResNet-20 on CIFAR-10.} \vspace{-4mm}
    \label{tab:complexity}
\end{table}

\smallskip
\noindent\textbf{From layer-wise to path-wise.} Here we provide the key idea on how to pass from the layer-wise perspective to the path-wise perspective when considering activations $\bm{a} \in \mathbb{R}^m$ and parameters $\bm{\theta} \in \mathbb{R}^m$. 
First of all, the relationship between $\bm{a}$ and $a_p$ is given by the definition of the latter. According to what we reported in Section 3.3 of the main submission, $a_p(\bm{x}, \bm{\theta}) = \prod_{\{i|\theta_i \in p\}} \mathbb{I}[z_i > 0]$ where $z_i$ is the activation of the neuron connected to the previous layer through parameter $\theta_i$. Thus, $a_p$ is a binary value representing the activation status of path $p$. This is equivalent to assigning a binary mask to each paramenter $\bm{\theta}_j$ representing the activation status of the neuron that it connects to (\ie $\bm{a}_j(\bm{x},\bm{\theta}) = \mathbb{I}[z_j > 0]$). This resulting binary-valued mask is $\bm{a}$. 
The expression of the $k$-th output of a neural network, can be written as \cite{gebhart2021unified}
\begin{equation*}
\begin{split}
    \bm{f}^k(\bm{x},\bm{\theta}) &= \sum_{s=1}^d \sum_{p \in \mathcal{P}_{s \rightarrow k}} \bm{v}_p(\bm{\theta}) a_p(\bm{x}, \bm{\theta}) \bm{x}_s \\
    &= \sum_{s=1}^d \sum_{p \in \mathcal{P}_{s \rightarrow k}} \prod_{j|j \in p} \bm{\theta}_j \bm{a}_j(\bm{x}, \bm{\theta}) \bm{x}_s \\
    &= \left[ \prod_{l=1}^L \bm{\theta}^{[l]} \bm{a}^{[l]}(\bm{x}, \bm{\theta}) \bm{x} \right]_k
\end{split}
\end{equation*}
where $L$ is the total number of layers and $l$ identifies the parameters and the activations of the $l$-th layer. We can recover the last equality observing that the second row of the equation is the definition of $M$ (where $M$ is the length of path $p$) consecutive matrix multiplications between the inputs and parameters of each layer, followed by an element-wise multiplication with the activations.

\smallskip
\noindent\textbf{Computational cost analysis.} This analysis, detailed in Table \ref{tab:complexity}, focuses on understanding the computational complexity of our method in contrast to that of our competitors. Notably, we include IMP in this assessment to offer a broader context to our study. In the second column of the table, we present the computational cost of invoking each pruning procedure, measured in numbers of macro-operations performed to obtain the final pruning scores. Where we report $\mathcal{O}(1)$ complexity, it means that the scores can be obtained immediately by simply looking at some intrinsic property of the network, such as the magnitude of the weights, which does not require any additional processing. Here, $T$ represents the required number of pruning iterations, and $B$ indicates the number of mini-batches processed by each algorithm during this procedure. Additionally, we denote the costs of a single forward pass with $[FP]$ and of a single backward pass with $[BP]$. 
Columns three and four illustrate the training epochs necessary for the pruning algorithm to attain approximately equivalent accuracy at 98.20\% sparsity when starting from ResNet-20 on CIFAR-10. This metric trivially stands at zero for PaI methods, given that the procedure is executed prior to training. However, IMP demands a minimum of 6 iterative rounds of pruning and subsequent re-training (each full training cycle spans 160 epochs) to surpass the accuracy reached by our method.
Within this analysis, it becomes apparent that PX's computational complexity mirrors that of NTK-SAP \cite{ntksap2023wang}, aligning generally with iterative PaI methods, which are explicitly constructed to iterate across $T$ rounds preceding training.

\smallskip
\noindent\textbf{Procedure clarifications and pseudocode.} In Algorithm \ref{algo:pseudocode}, we provide the pseudocode to further clarify the role of $T$ and $B$ in the implementation of PX. The functions $\bm{g}, \bm{h}$ are copies of $\bm{f}$ defined for clarity, but the memory usage does not double as PX only stores $\bm{\theta}$ and the derivatives \wrt $\bm{\theta}^2$ are computed in a single pass (lines 13-15).
PX is an iterative PaI method but differs from the standard framework in lines 3, 9-15. We remark that iteratively refining the pruning mask $\bm{M}$ in $T$ rounds while yielding positive saliency scores guarantees to avoid layer collapse \cite{tanaka2020synflow}.

\begin{table}[h!]
    \begin{center}
    \small
    \setcellgapes{1.5pt}
    \makegapedcells
    \resizebox{0.44\textwidth}{!}{
    \begin{tabular}{c | c | c | c | c | c }
        \bottomrule      
        \multicolumn{6}{c}{\textbf{ResNet-20 (CIFAR-10) - 98.20\% Sparsity - ($\bm{T=100}$)}} \\
        \hline\hline
        {} & $B=1$ & $B=1$ & $B=1$ & $B=2$ & $B=4$ \\
        {} & $|D|=1$ & $|D|=5$ & $|D|=10$ & $|D|=50$ & $|D|=100$ \\
        \hline
        
        \multicolumn{6}{c}{Data-driven Single-shot PaI methods} \\ \hline

        SNIP \cite{lee2018snip} & 73.01 & 74.11 & 75.39 & 75.17 & 75.49 \\
        GraSP \cite{grasp2020wang} & 74.74 & 76.11 & 76.30 & 76.41 & 76.36 \\ \hline

        \multicolumn{6}{c}{Data-driven Iterative PaI methods} \\ \hline
        
        \textbf{PX (Ours)} & 76.96 & 77.06 & 77.08 & 77.16 & 77.13 \\
        
        \bottomrule
    
    \end{tabular}}
    \end{center}
    \vspace{-5mm}
    \caption{Ablation on the amount of data used to estimate the saliency scores for the data-driven pruning methods. We report the classification accuracy at 98.20\% sparsity when starting from ResNet-20 on CIFAR-10, while varying the amount of examples per class $|D|$.} 
    \vspace{-4mm}
    \label{tab:tab_B_ablation}
\end{table}

\section{Additional Experiments}

\smallskip
\noindent\textbf{Segmentation experiments.} In Fig. \ref{fig:add_segmentation_exp} we report the full results of the semantic segmentation experiments on the Pascal VOC2012 \cite{everingham2015pascal} dataset. In each experiment the architecture used is DeepLabV3+ \cite{chen2018encoder} on a ResNet-50 \cite{he2016deep} backbone, starting from ImageNet \cite{deng2009imagenet}, MoCov2 on ImageNet \cite{chen2020improved} and DINO \cite{caron2021emerging} pre-trained models. 

The general trend reported in the main paper is confirmed also in this setting, where our method is able to retain the accuracy of the dense baseline at trivial sparsities.

\smallskip
\noindent\textbf{SynFlow-L2 ablation study.} 
As mentioned in the main paper, our method can be interpreted as an extension of SynFlow-L2 \cite{gebhart2021unified}. The core difference is that PX reweights the network's outputs on the basis of the information provided by the data: that information indicates how much each weight contributes to the upper bound on the trace of the NTK reported in Eq. (6) of the main submission.
Thus, by comparing PX with SynFlow-L2 we conduct an ablation study to provide further evidence regarding the soundness of the hypotheses underpinning our algorithm, proving the importance of the data-dependent component. In Fig. \ref{fig:ablation_random}, \ref{fig:ablation_pretrain} and \ref{fig:ablation_segm} we observe that our method is always able to improve over SynFlow-L2. Furthermore, the latter exhibits a drastic decrease in performance when the cardinality of the network's output increases. 
As already noticed in \cite{patil2021phew}, this is attributed to the combined effect of reducing the layer width while keeping a high path count in the architecture.

\smallskip
\noindent\textbf{Layer widths.} In Fig. \ref{fig:add_widths}, we present additional plots on the layer width to confirm the trend reported in the main submission regarding the number of output units preserved by PX at each layer. We observe again that PX is able to preserve the output width despite the very high sparsity ratios under exam and the different model sizes.

\smallskip
\noindent\textbf{Spectral analyses.} Fig. \ref{fig:add_spectrum} provides further evidence regarding the preservation of the full eigenspectrum thanks to the approximation of our upper bound detailed in Eq. (6) of the main submission.

\smallskip
\noindent\textbf{Data amount analysis.} We indicate with $|D|$ the number of examples per class. The relationship between $|D|$ and the number of batches $B$ is given by $B = \lceil |D| \times |C|/b \rceil$, where $|C|$ is the number of classes, and $b$ is the batch size. In Table \ref{tab:tab_B_ablation}, we extend Table \ref{tab:complexity} by showing the effect of changing $B$ and $|D|$ on the data dependent methods. PX is less sensitive than the competitors, and increasing $|D|$ does not yield a significant performance gain in line with the findings of \cite{lee2018snip}.

\end{document}